\documentclass[10pt,twocolumn,letterpaper]{article}

\usepackage[pagenumbers]{cvpr} 

\usepackage{graphicx}
\usepackage{amsmath}
\usepackage{amssymb}
\usepackage{booktabs}
\usepackage{enumitem}
\usepackage{makecell}
\usepackage{multicol}
\usepackage{multirow}
\usepackage{soul}
\usepackage{float}
\usepackage{media9}

\usepackage{algorithm}
\usepackage{algorithmic}

\usepackage[dvipsnames]{xcolor}

\definecolor{cvprblue}{rgb}{0.21,0.49,0.74}
\usepackage[pagebackref,breaklinks,colorlinks,citecolor=cvprblue]{hyperref}

\usepackage[capitalize]{cleveref}
\crefname{section}{Sec.}{Secs.}
\Crefname{section}{Section}{Sections}
\Crefname{table}{Table}{Tables}
\crefname{table}{Tab.}{Tabs.}

\makeatletter 
\def\@captype{figure} 
\makeatother

\def\paperID{***} 
\def\confName{CVPR}
\def\confYear{2024}

\title{RealisDance: Equip controllable character animation with realistic hands}

\author{Jingkai Zhou~~~Benzhi Wang~~~Weihua Chen~~~Jingqi Bai~~~Dongyang Li~~~Aixi Zhang~~~Hao Xu\\~~~Mingyang Yang~~~Fan Wang\\ {Alibaba Group}}

\begin{document}
\maketitle

\begin{abstract}
Controllable character animation is an emerging task that generates character videos controlled by pose sequences from given character images. Although character consistency has made significant progress via reference UNet, another crucial factor, pose control, has not been well studied by existing methods yet, resulting in several issues: 1) The generation may fail when the input pose sequence is corrupted. 2) The hands generated using the DWPose sequence are blurry and unrealistic. 3) The generated video will be shaky if the pose sequence is not smooth enough. In this paper, we present RealisDance to handle all the above issues. RealisDance adaptively leverages three types of poses, avoiding failed generation caused by corrupted pose sequences. Among these pose types, HaMeR provides accurate 3D and depth information of hands, enabling RealisDance to generate realistic hands even for complex gestures. Besides using temporal attention in the main UNet, RealisDance also inserts temporal attention into the pose guidance network, smoothing the video from the pose condition aspect. Moreover, we introduce pose shuffle augmentation during training to further improve generation robustness and video smoothness. Qualitative experiments demonstrate the superiority of RealisDance over other existing methods, especially in hand quality. Codes are available at \href{https://github.com/damo-cv/RealisDance}{this link}.
\end{abstract}    
\section{Introduction}

Controllable character image animation has attracted widespread attention~\cite{pidm, disco, dream_pose, aa, magic_animate, magic_pose, champ}. It takes character images and pose sequences as input, and aims to generate videos in which the character's clothing and ID are consistent with the given ones and the character moves according to the pose sequence. Recently, with the introduction of the reference UNet~\cite{aa, magic_animate, magic_pose}, great progress has been made in character consistency. Beyond character consistency, we observe that the generation quality is also highly correlated with pose control. However, existing methods rarely explore pose control, thus suffering from unstable generation, poor hand quality, and video shaking.

\begin{figure}[t]
    \centering
    \addtolength{\tabcolsep}{-5.5pt}
    \begin{tabular}{ccc}
        Ref Image & Gen Frame 1 & Gen Frame 2\\
        \includegraphics[width=0.33\linewidth]{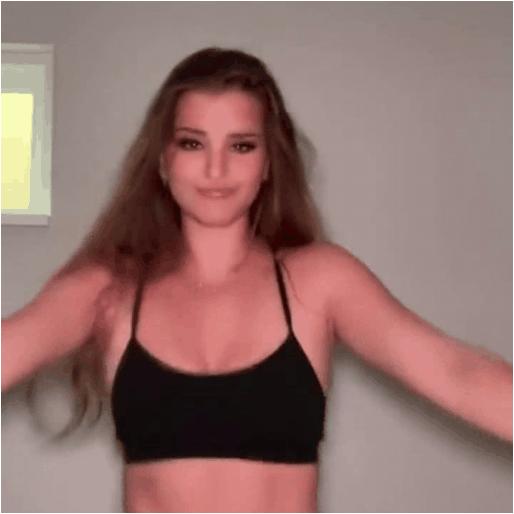} &
        \includegraphics[width=0.33\linewidth]{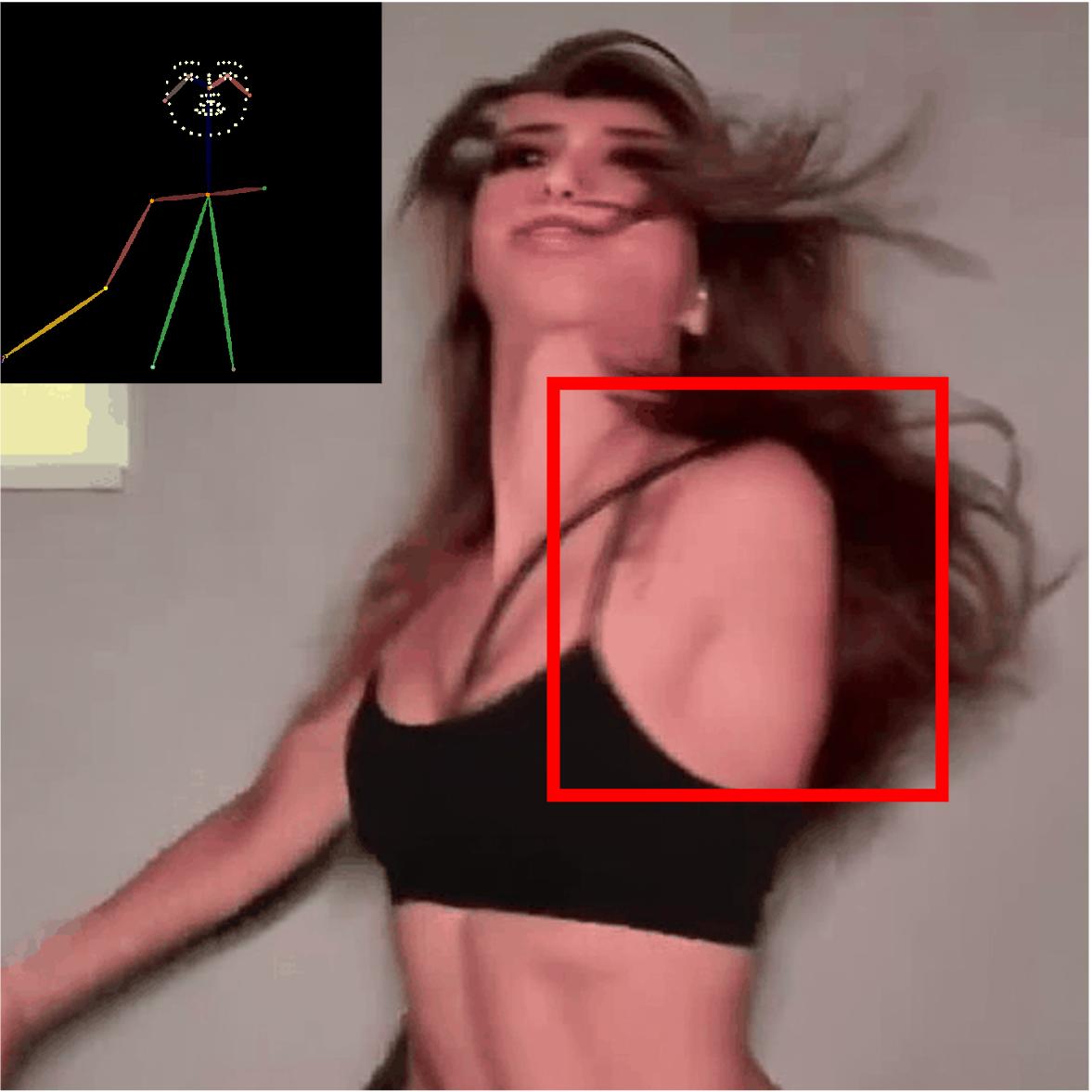} & 
        \includegraphics[width=0.33\linewidth]{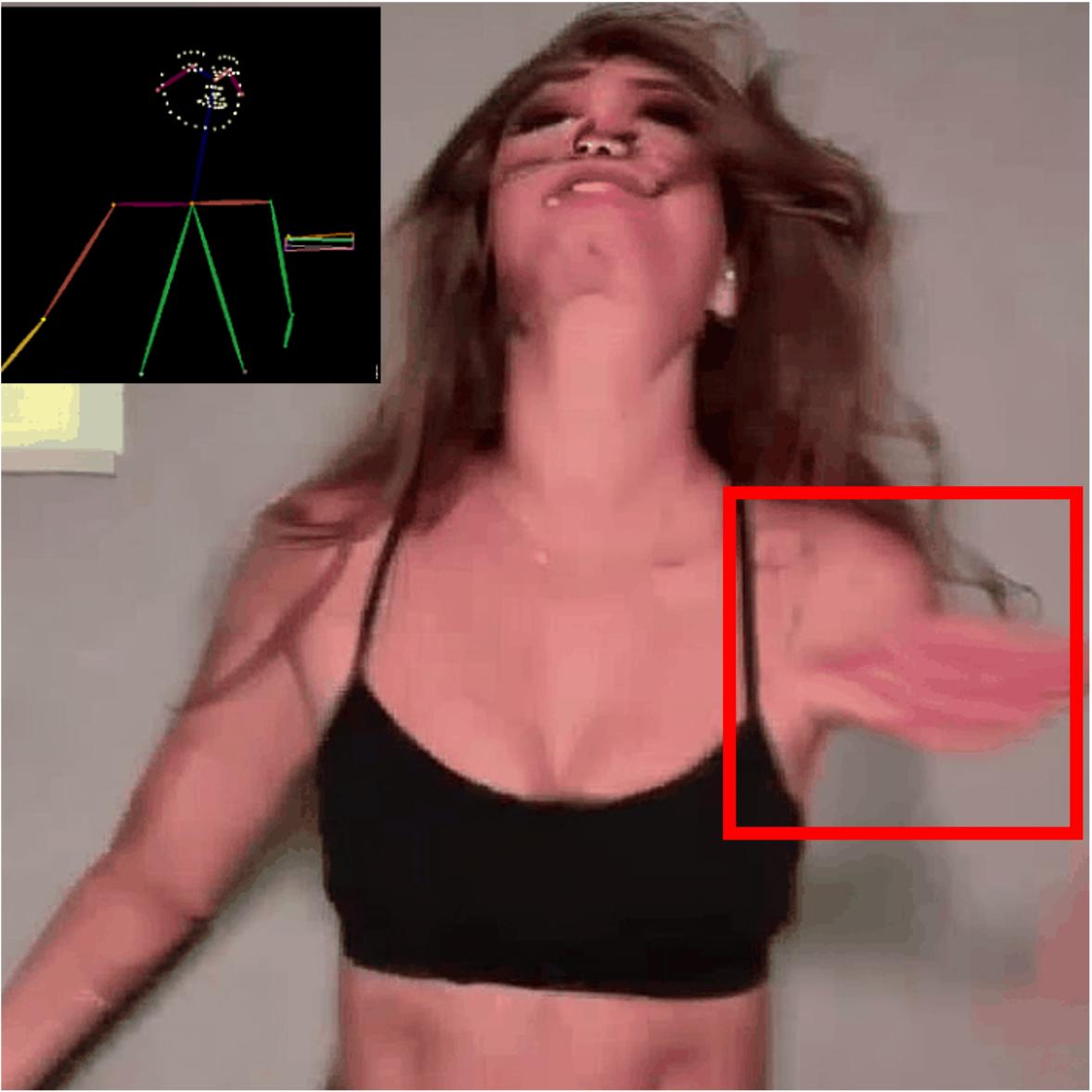} \\
        \includegraphics[width=0.33\linewidth]{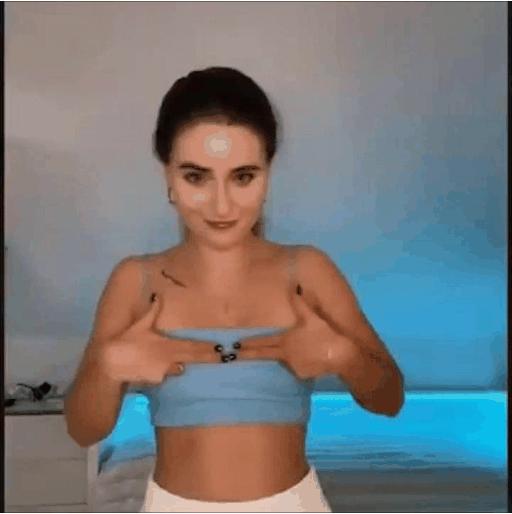} &
        \includegraphics[width=0.33\linewidth]{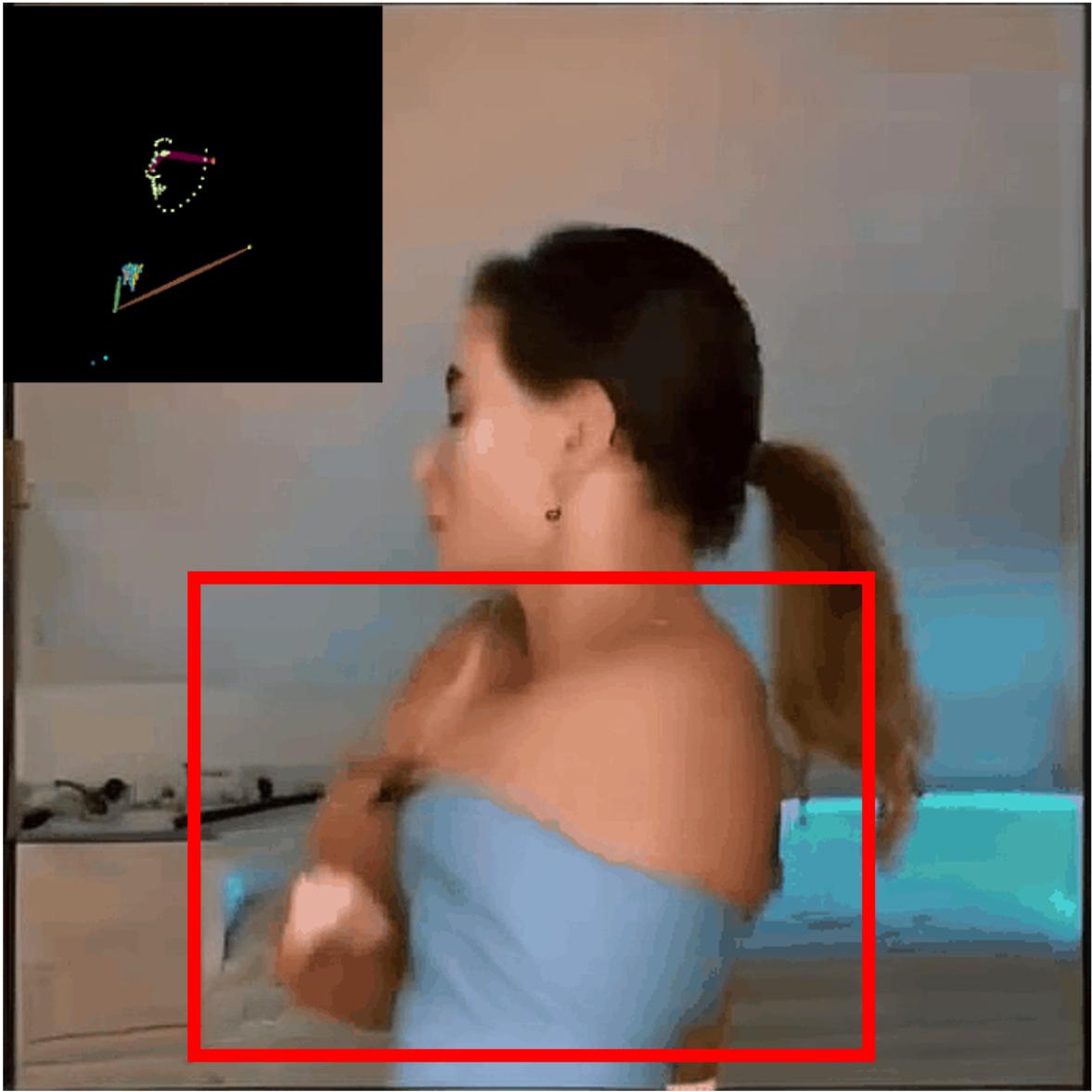} & 
        \includegraphics[width=0.33\linewidth]{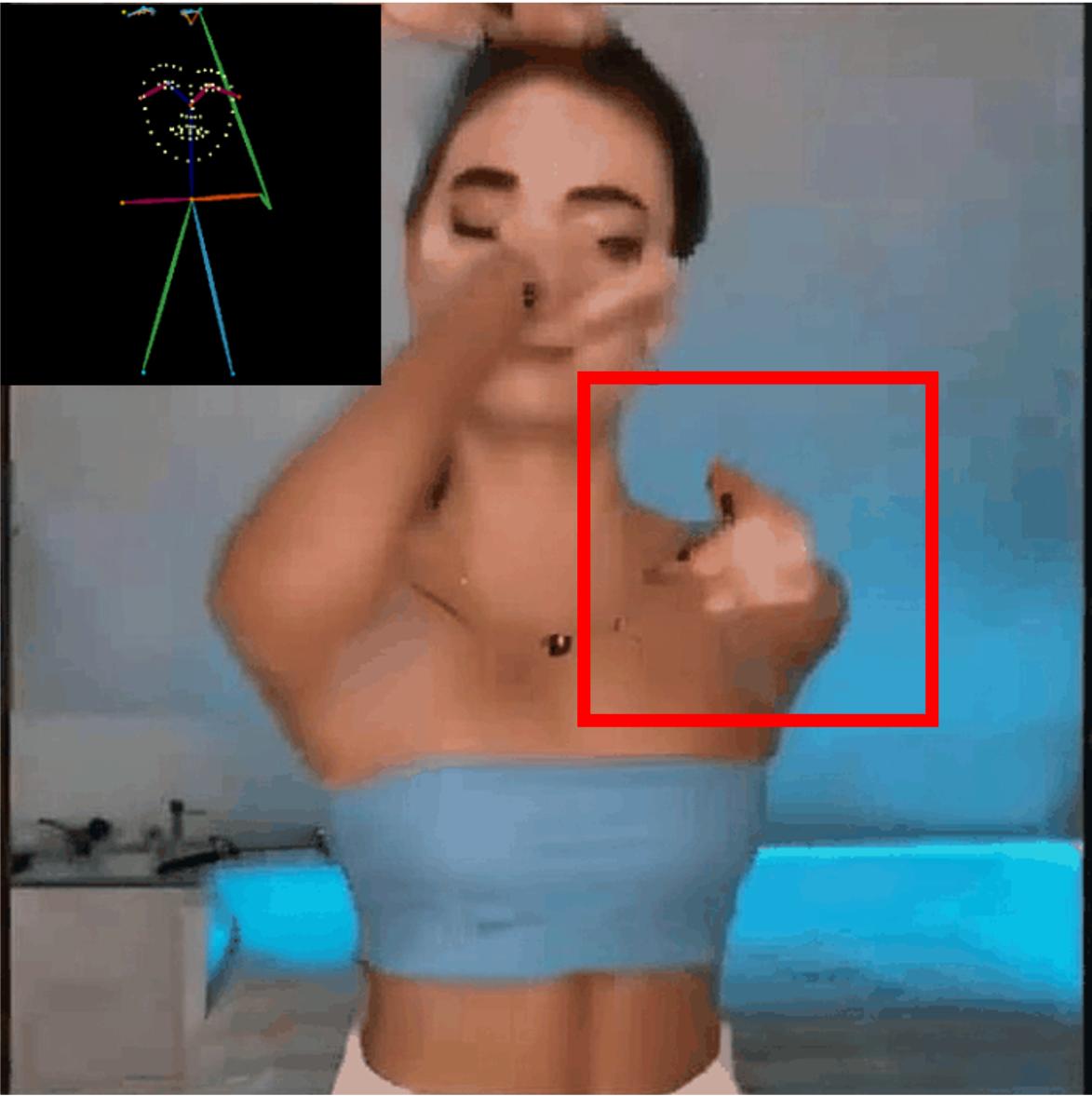} \\
        \includegraphics[width=0.33\linewidth]{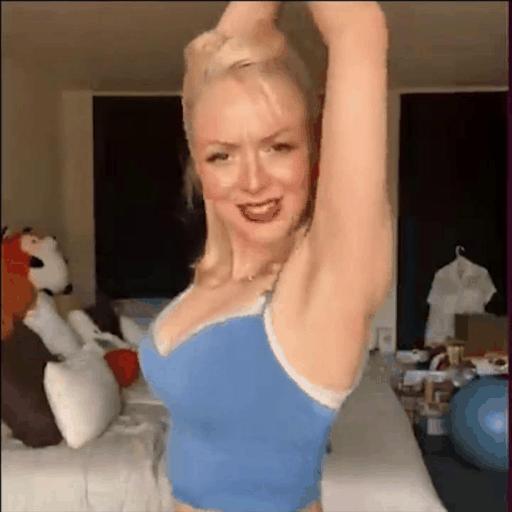} &
        \includegraphics[width=0.33\linewidth]{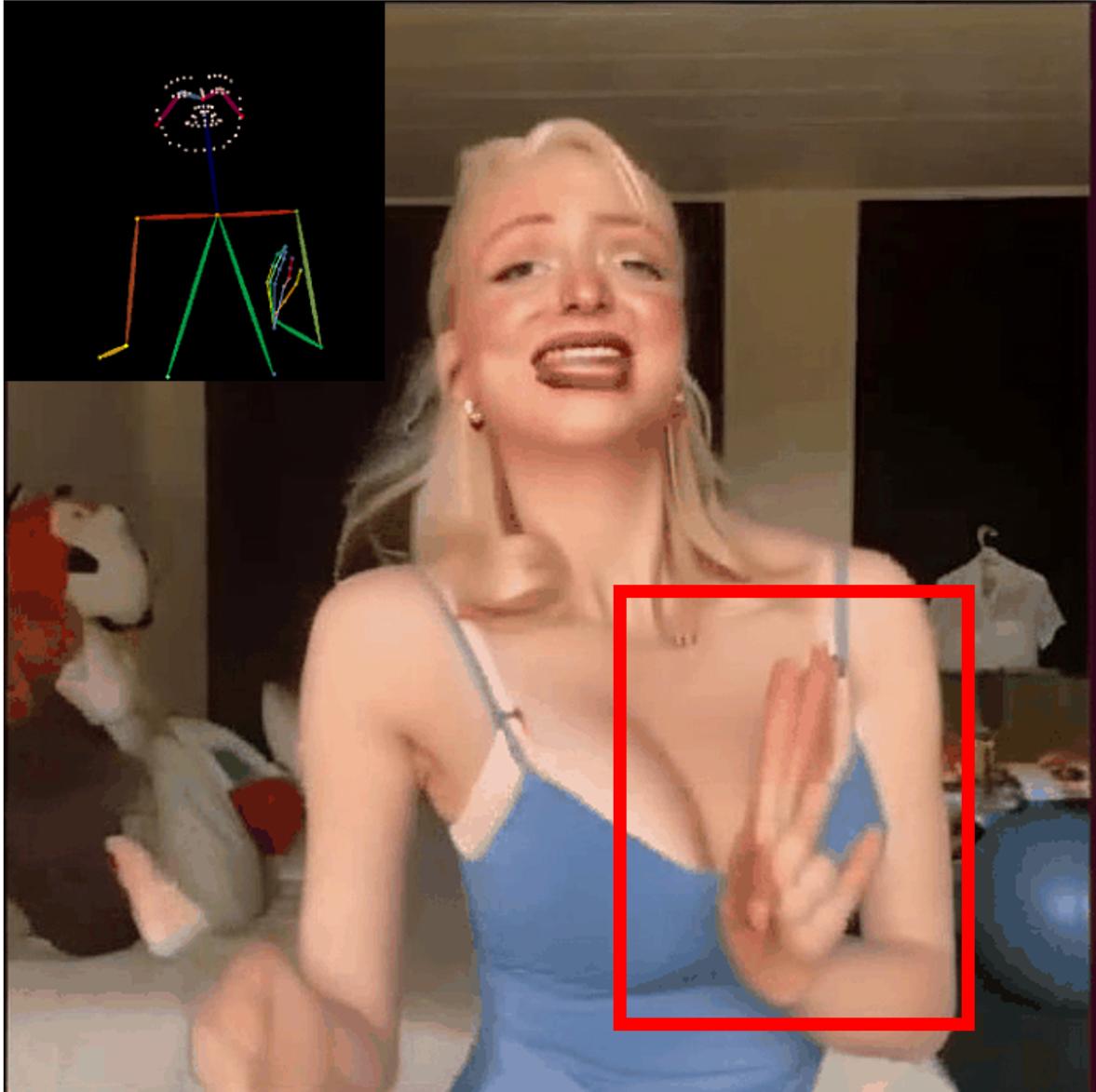} & 
        \includegraphics[width=0.33\linewidth]{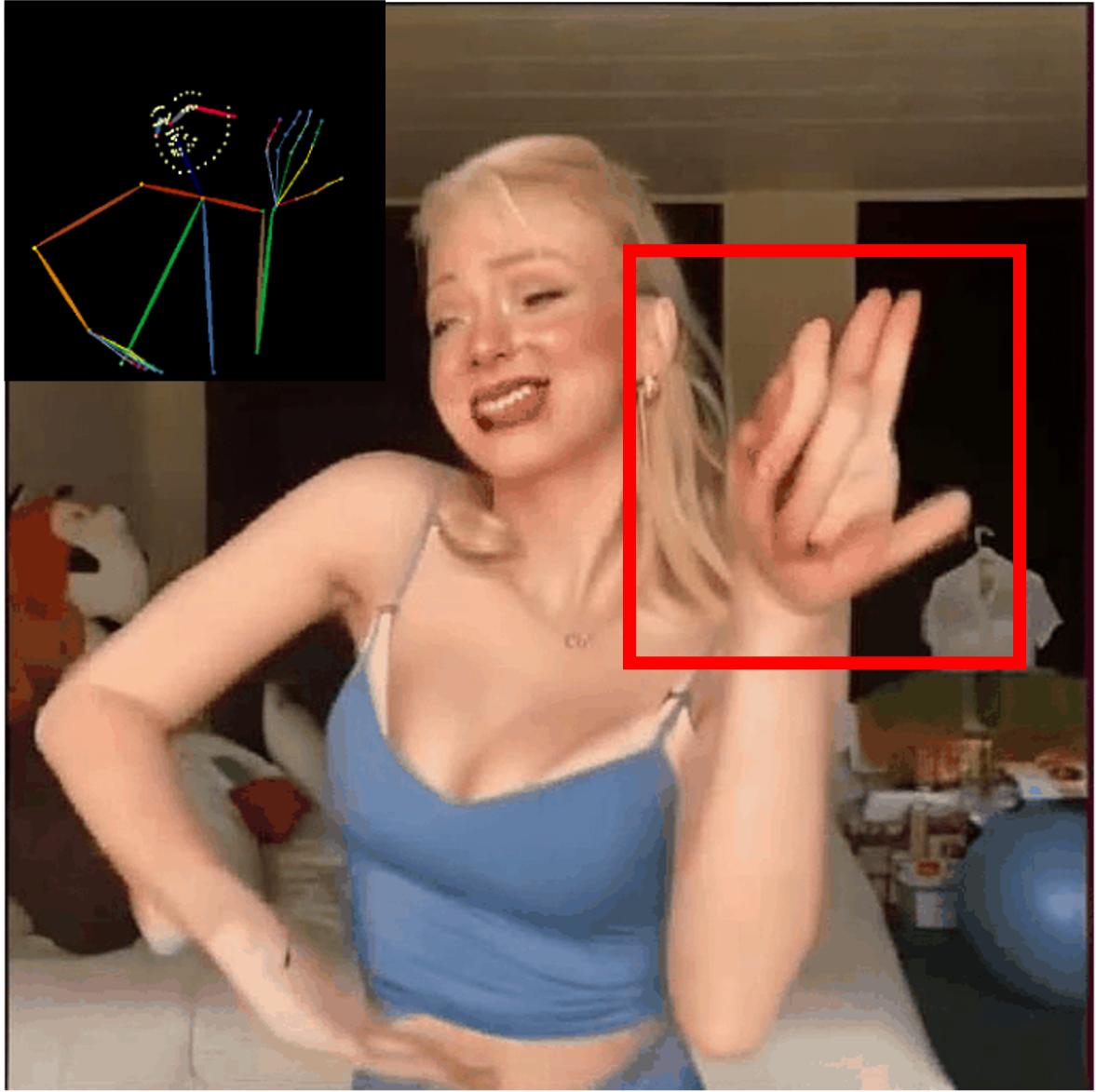} \\
        \includegraphics[width=0.33\linewidth]{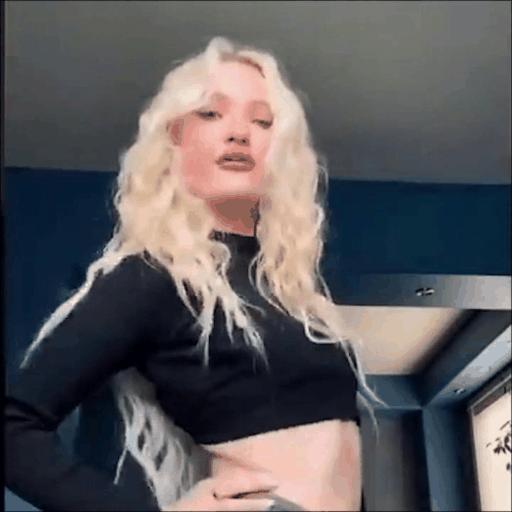} &
        \includegraphics[width=0.33\linewidth]{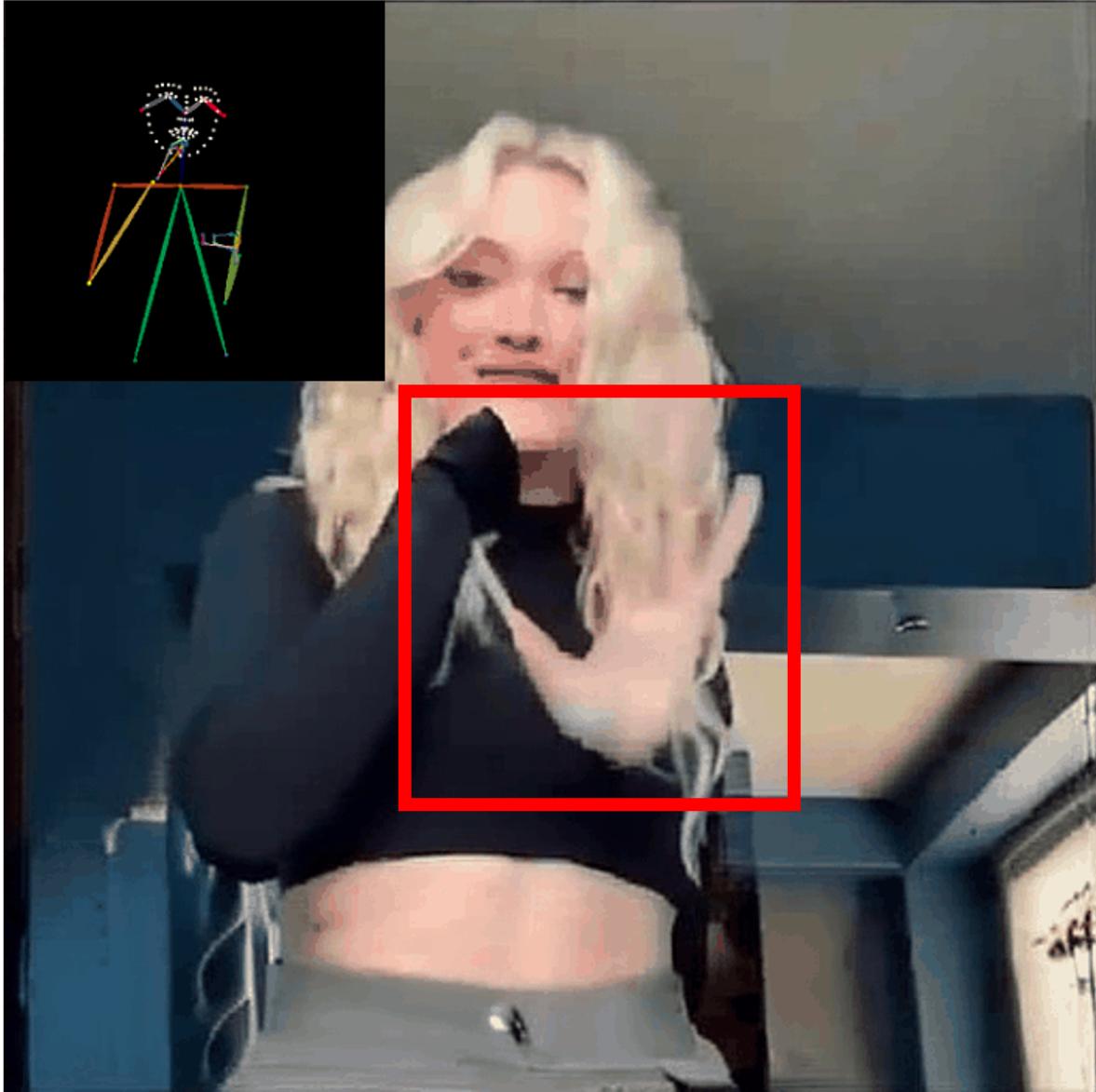} & 
        \includegraphics[width=0.33\linewidth]{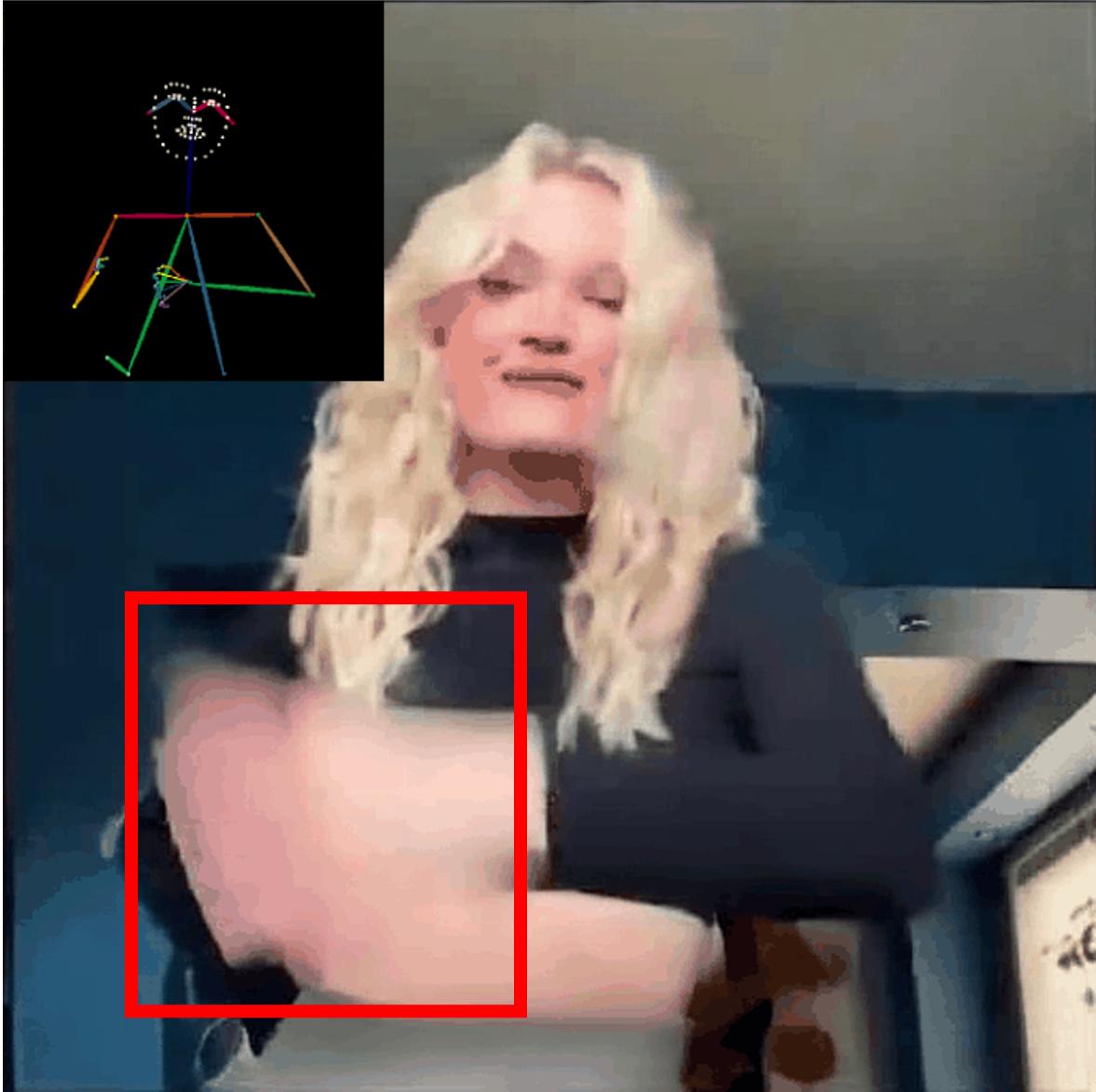} 
    \end{tabular}
    \caption{Samples generated from our reproduced Animate Anyone. Animate Anyone suffers from unstable generation if the condition pose is corrupted, as shown in the first two rows. Also, even if the condition pose is correct,  Animate Anyone generates blur and unrealistic hands, as shown in the last two rows.}
    \label{fig:issues}
\end{figure}

\begin{figure*}[t]
    \centering
    \addtolength{\tabcolsep}{-5.5pt}
    \begin{tabular}{ccccc}
        Ref Image & Gen Frame 1 & Gen Frame 2 & Gen Frame 3 & Gen Frame 4 \\
        \includegraphics[width=0.2\linewidth]{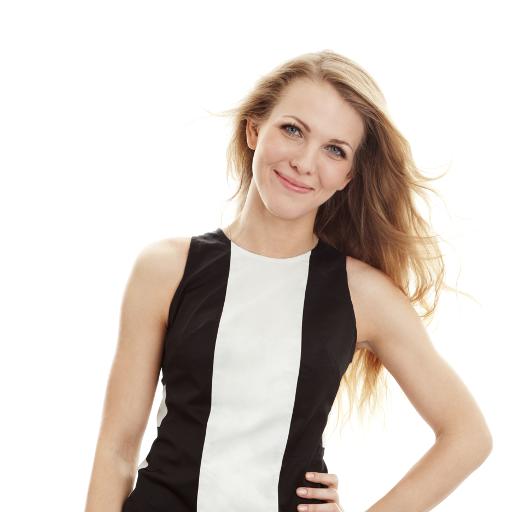} &
        \includegraphics[width=0.2\linewidth]{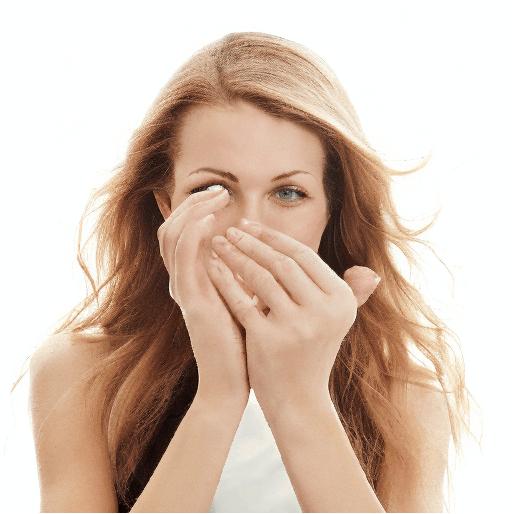} & \includegraphics[width=0.2\linewidth]{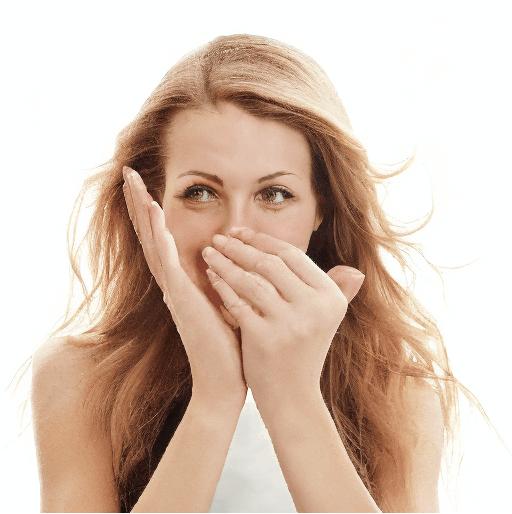} & \includegraphics[width=0.2\linewidth]{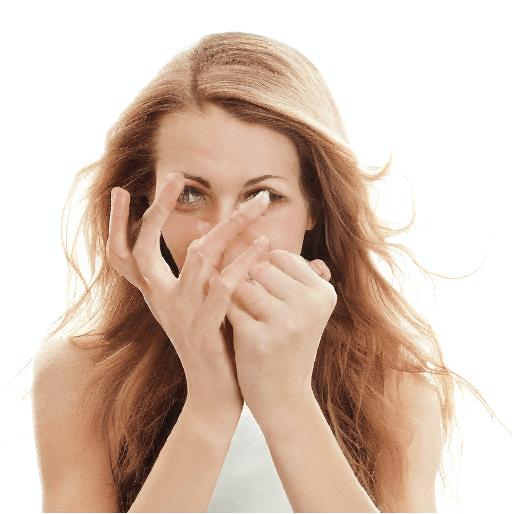} &
        \includegraphics[width=0.2\linewidth]{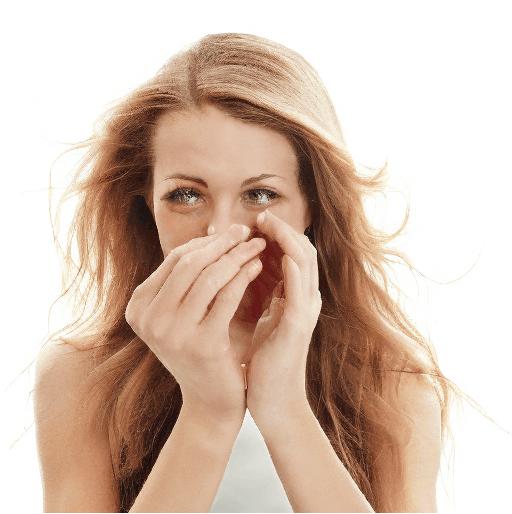} \\
        \includegraphics[width=0.2\linewidth]{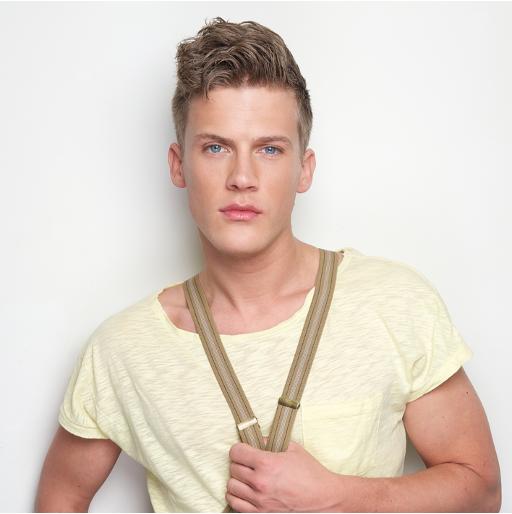} &
        \includegraphics[width=0.2\linewidth]{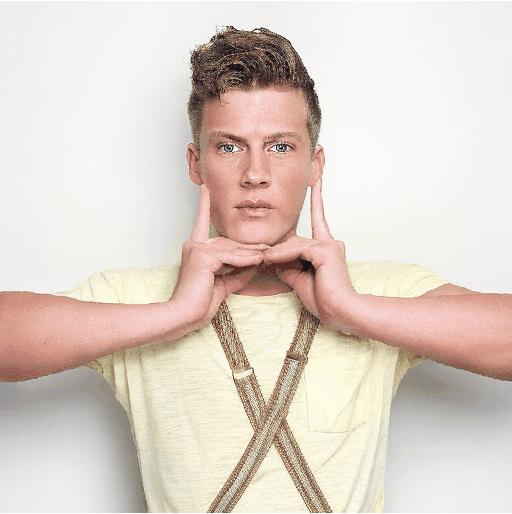} & \includegraphics[width=0.2\linewidth]{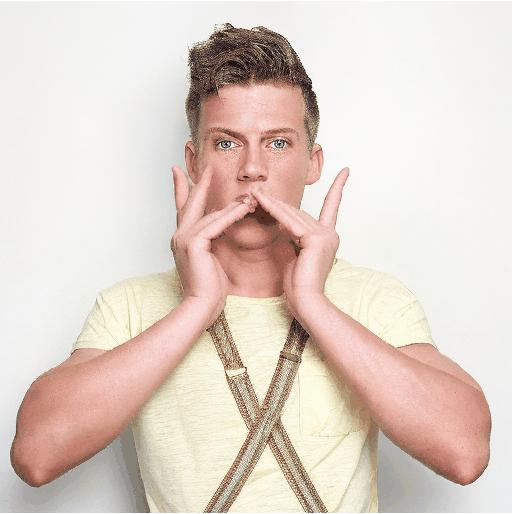} & \includegraphics[width=0.2\linewidth]{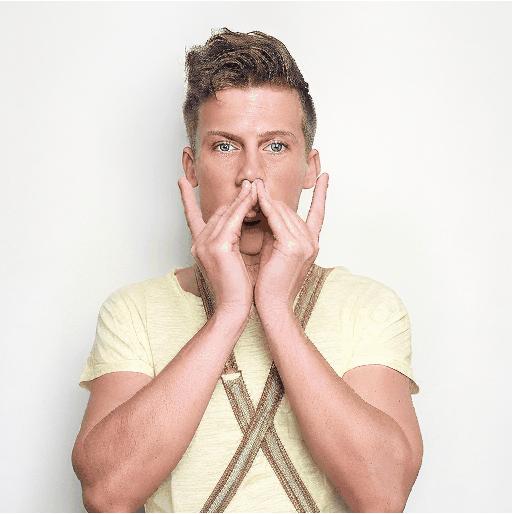} &
        \includegraphics[width=0.2\linewidth]{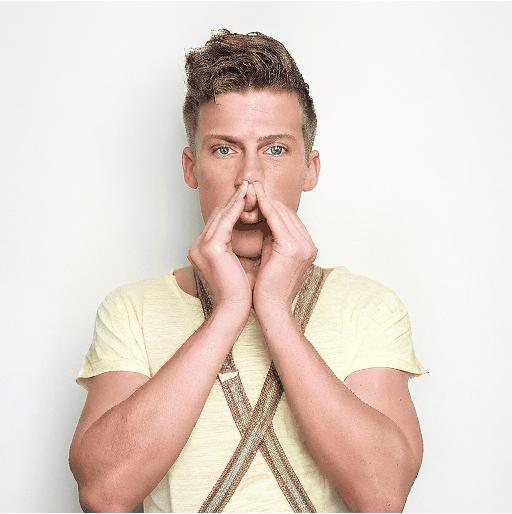} \\
        \includegraphics[width=0.2\linewidth]{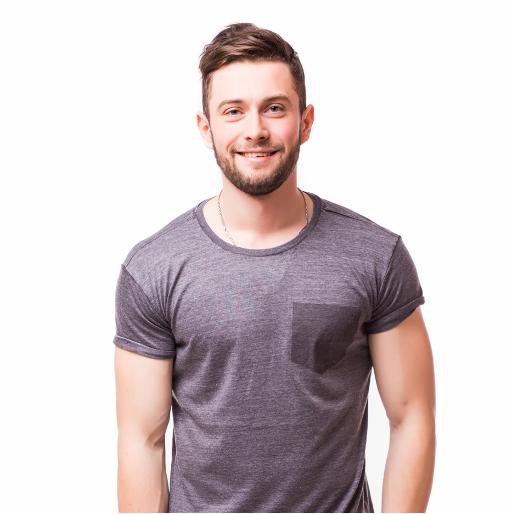} &
        \includegraphics[width=0.2\linewidth]{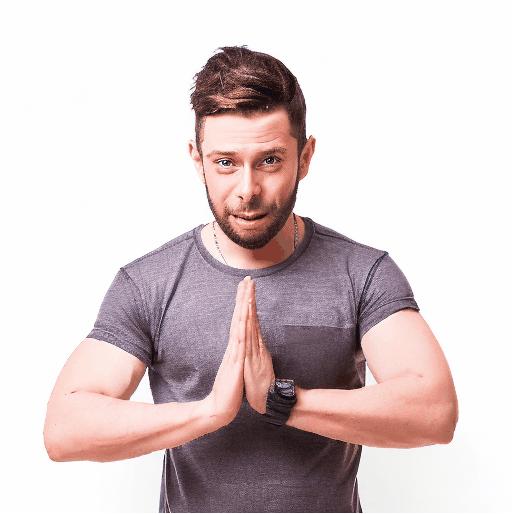} & \includegraphics[width=0.2\linewidth]{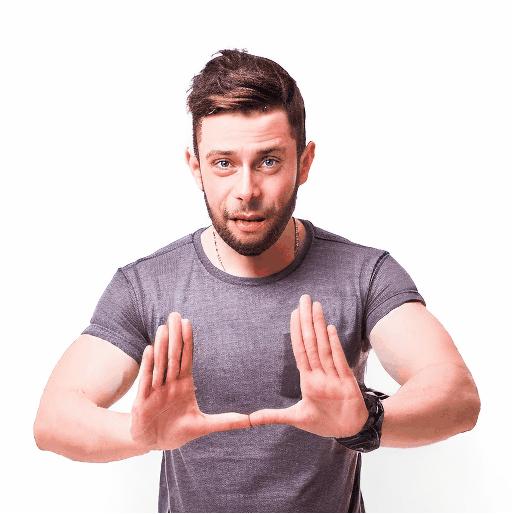} & \includegraphics[width=0.2\linewidth]{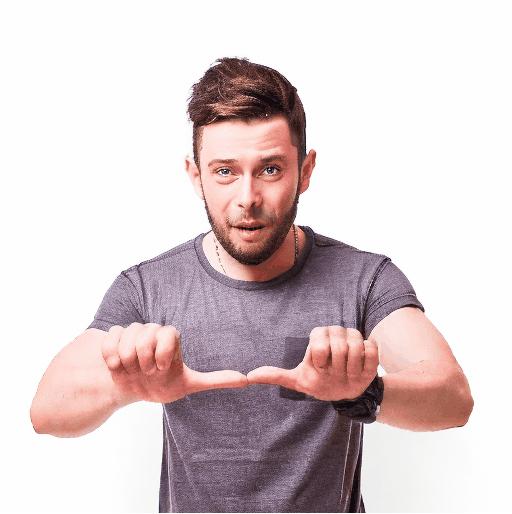} &
        \includegraphics[width=0.2\linewidth]{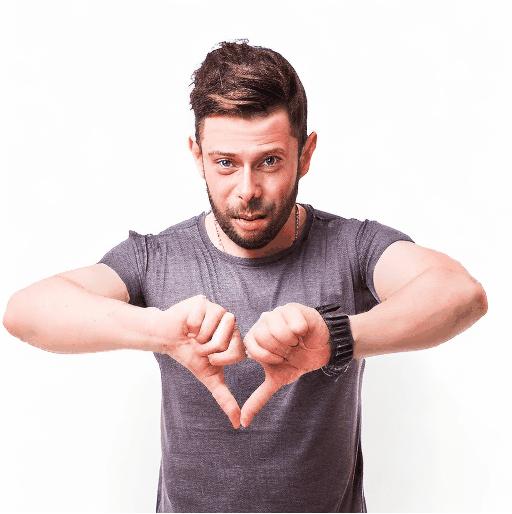} 
    \end{tabular}
    \caption{Samples generated from RealisDance. As can be seen, the generated results achieve high-quality hands even for complex gestures.}
    \label{fig:qulitative_results_1}
\end{figure*}

\vspace{1mm}
\noindent\textbf{Unstable generation.} 
The pose sequences used in existing training and inference are estimated from real video data, and thus, may be corrupted due to false detection. As shown in Figure~\ref{fig:issues}, the generation quality degrades significantly when the pose condition is corrupted. One existing solution is manually selecting high-quality pose sequences, which does not fundamentally address the underlying issue.

\vspace{1mm}
\noindent\textbf{Poor hands.}
Existing methods use OpenPose~\cite{open_pose, disco}, DWPose~\cite{dwpose, aa}, Densepose~\cite{densepose, magic_pose}, or SMPL~\cite{smpl, champ} to drive the video generation. However, as shown in Figure~\ref{fig:issues}, it is difficult to generate realistic hands conditioned on any of the above poses due to lack of 3D/depth information~\cite{dwpose, open_pose, densepose} or inaccurate pose conditions~\cite{open_pose, densepose, smpl}.

\vspace{1mm}
\noindent\textbf{Video shaking.} As mentioned above, the pose sequence is extracted using pose estimation methods, like DWPose. However, such pose estimation methods are applied to static images, which inevitably introduces inter-frame shaking to the estimated pose sequences. Although existing character animation methods integrate temporal attention~\cite{animate_diff} into the main UNet to smooth video, we find that the influence of pose sequence control is so dominant that temporal attention used in the main UNet is insufficient to mitigate video shaking completely.

\vspace{1mm}
In this paper, we propose RealisDance to deal with the above problems.
For robust generation, RealisDance employs an adaptive pose gating module to integrate three distinct pose sequences that are unlikely to be corrupted simultaneously. When one pose sequence is corrupted, RealisDance can adaptively reduce the corresponding gate weight and still obtain the correct generation driven by the other two.
To improve hand quality, HaMeR, a state-of-the-art 3D gesture estimation method, is adopted as one of the three pose types to provide accurate 3D and depth information of hands. Thanks to the HaMeR sequences, RealisDance can generate realistic hands even for complex gestures. 
To ensure smooth video, RealisDance not only applies temporal attention to the main UNet, but also inserts it into the pose guidance network, mitigating video shaking from both generation and condition aspects.
Moreover, we introduce the pose shuffle augmentation during training, further improving generation robustness and video smoothness.
Equipped with reference UNet and the above pose control modifications, RealisDance surpasses existing methods by a large margin in qualitative comparisons, achieving robust generation, realistic hands, and smooth video. Figure~\ref{fig:qulitative_results_1} shows several samples generated from RealisDance.

\begin{figure*}[t]
    \centering
    \includegraphics[width=\linewidth]{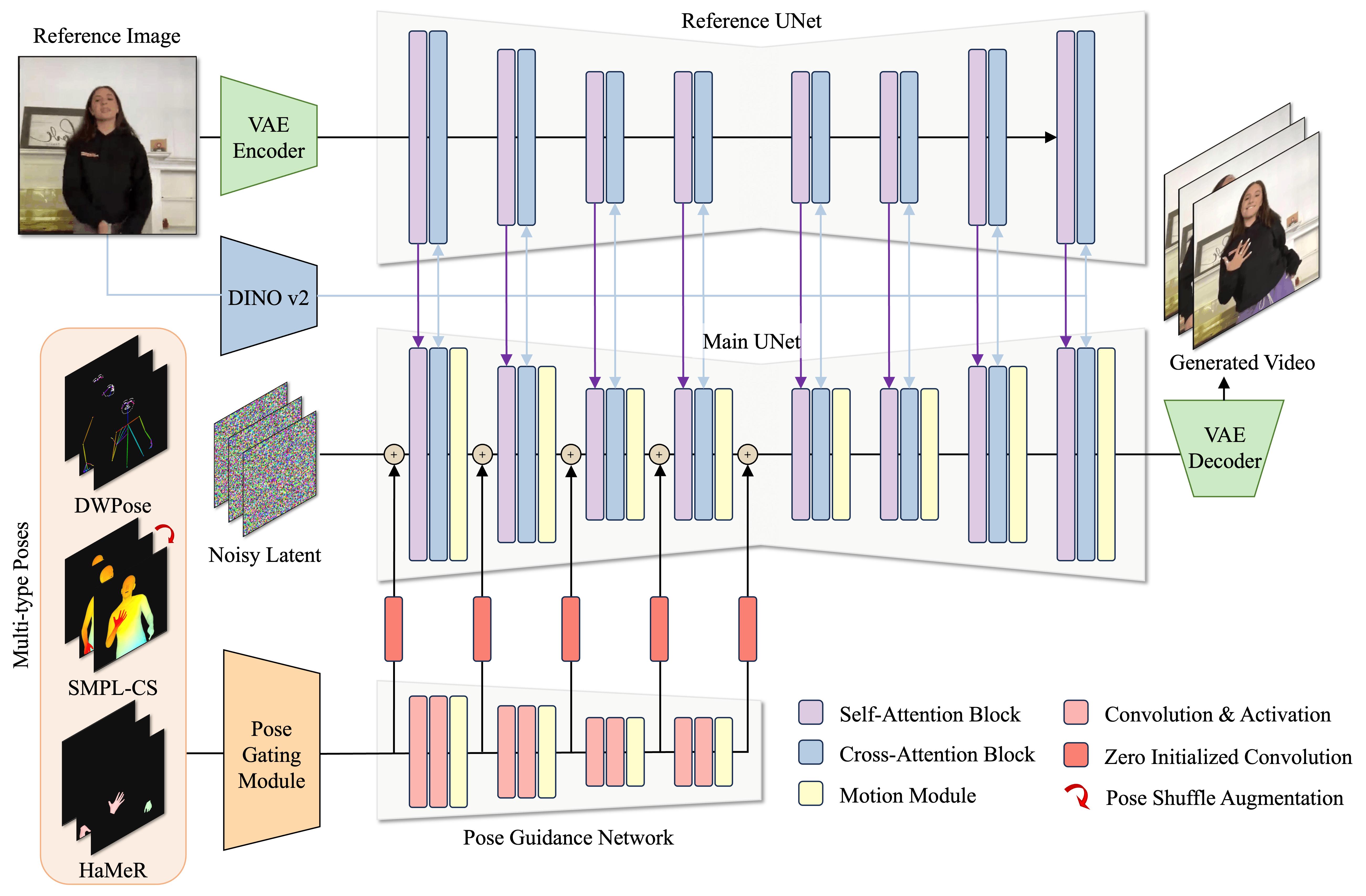}
    \caption{Architecture of RealisDance. Thanks to multi-type poses, the pose gating module, the multi-layer pose guidance network, and the pose shuffle augmentation, RealisDance achieves robust generation, realistic hands, and smooth video.}
    \label{fig:arch}
\end{figure*}

\section{Method}
As we observe that the generation quality is highly correlated with pose control, this paper focuses on pose control and proposes RealisDance with four key modifications: multi-type poses, the pose gating module, the multi-layer pose guidance network, and the pose shuffle augmentation. The architecture of RealisDance is shown in Figure~\ref{fig:arch}.

\begin{figure}[t]
    \centering
    \addtolength{\tabcolsep}{-5.5pt}
    \begin{tabular}{cccc}
        \includegraphics[width=0.24\linewidth]{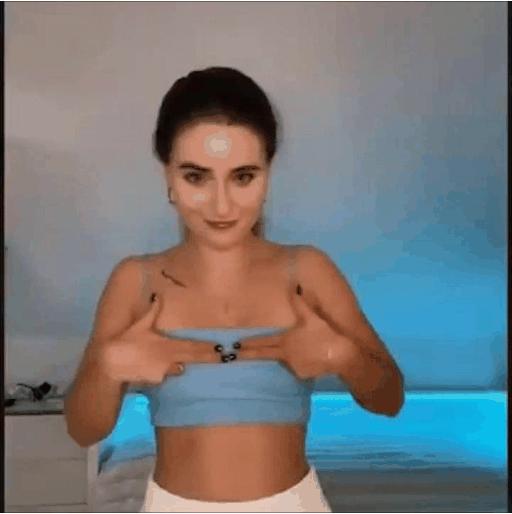} &
        \includegraphics[width=0.24\linewidth]{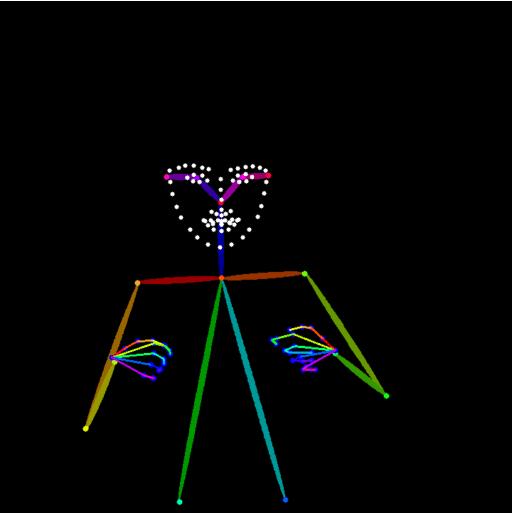} & \includegraphics[width=0.24\linewidth]{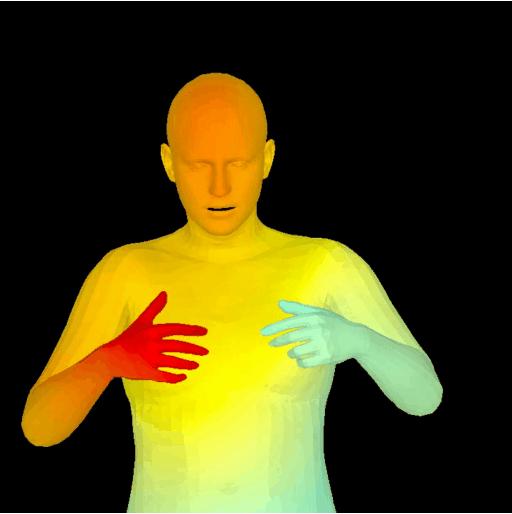} & \includegraphics[width=0.24\linewidth]{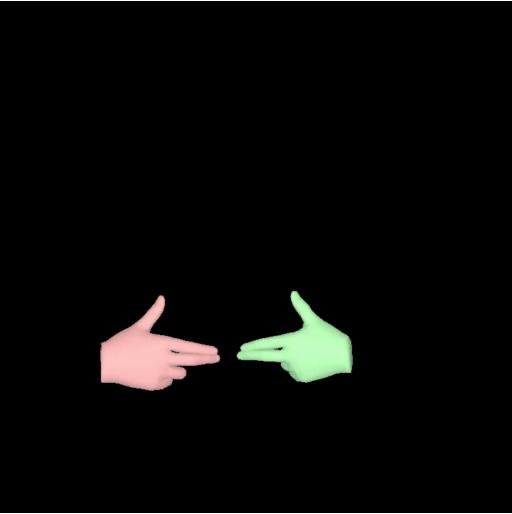} \\
        Ref Image & DWPose & SMPL-CS & HaMer
    \end{tabular}
    \caption{Illustration of three types of poses. SMPL-CS integrates 3D, depth, and continuous semantic information, and HaMer provides accurate 3D gesture estimation.}
    \label{fig:pose_show}
\end{figure}

\vspace{1mm}
\noindent\textbf{Multi-type poses.} RealisDance takes three types of poses as input: DWPose~\cite{dwpose}, SMPL-Colorful Surface (SMPL-CS), and HaMeR~\cite{hamer}. Figure~\ref{fig:pose_show} illustrates these three types of poses. 
DWPose, like OpenPose~\cite{open_pose}, utilizes 2D coordinates to annotate human body keypoints, hands, and face landmarks. During training and inference, DWPose is rendered as points and edges with different colors. 
SMPL-CS is based on SMPLer-X~\cite{smpler_x}. We use SMPLer-X to estimate the 3D parameters~\cite{smplx} of the target human body and render the surface with gradient color so that the rendering results integrate 3D, depth, and continuous semantic information simultaneously, avoiding cumbersome and redundant pose input. 
HaMeR is the key factor for RealisDance to generate realistic hands. HaMeR is a state-of-the-art 3D hand gesture estimation method that leverages the scaling-up capabilities of Transformers which is trained on large-scale hand datasets.
Compared with DWPose, HaMer can provide 3D and depth information to help RealisDance understand the spatial structure of gestures. Compared with SMPL-CS, HaMeR obtains more accurate estimation for complex gestures. Thanks to the HaMeR sequence, RealisDance gets a significant improvement in the quality of hand generation.

\begin{figure}[t]
    \centering
    \includegraphics[width=0.9\linewidth]{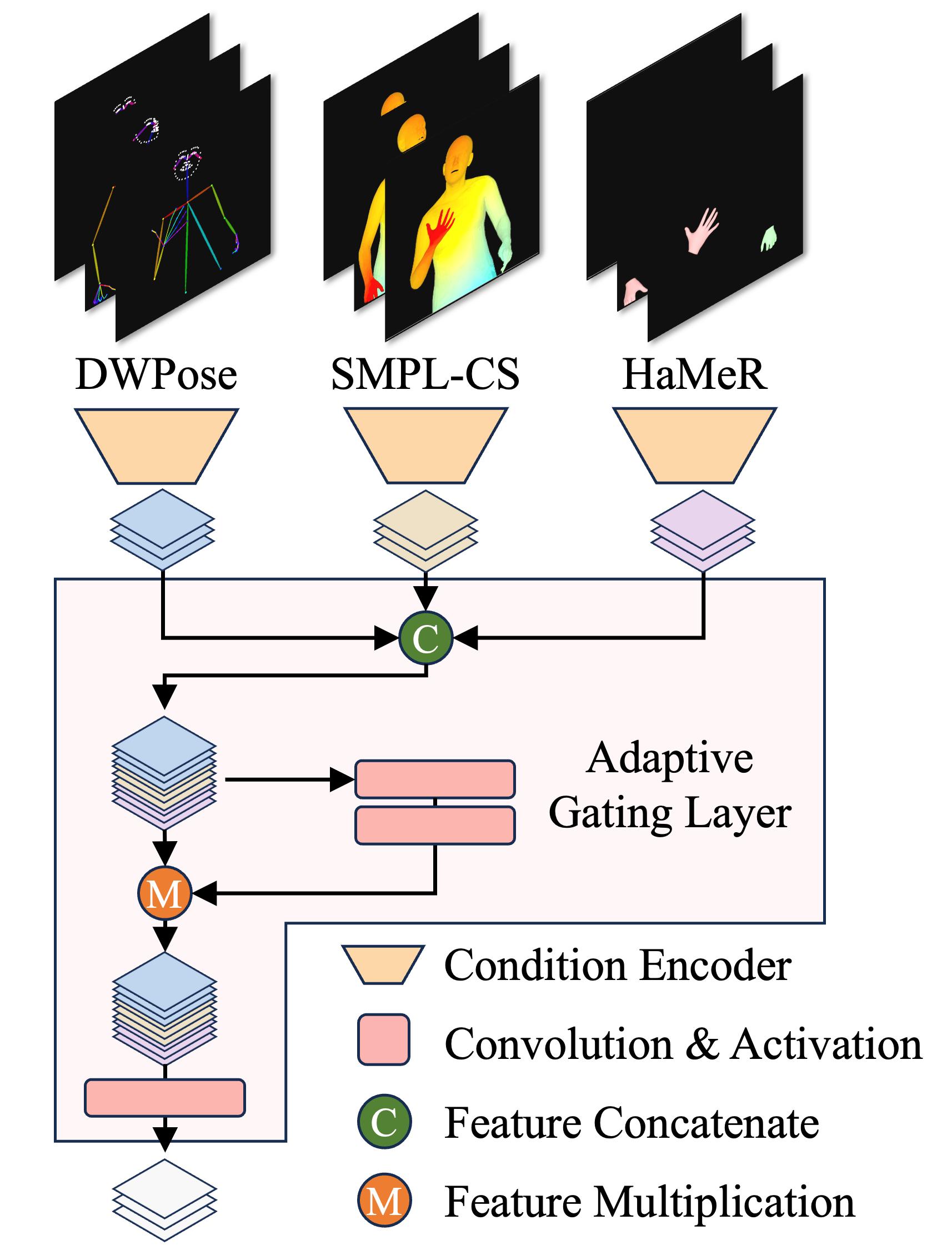}
    \caption{Architecture of pose gating module. In practice, we implement three individual condition encoders using one encoder with grouped convolution for faster speed.}
    \label{fig:pose_gating}
\end{figure}

\vspace{1mm}
\noindent\textbf{Pose gating module.}
The pose gating module includes three individual condition encoders to embed three types of poses respectively and an adaptive gating layer to merge three encoded features. Condition encoder only includes several convolutional layers and SiLU activation layers~\cite{silu}, just like the one used in ControlNet~\cite{controlnet}. The adaptive gating layer first concatenates three features and then feeds the concatenated feature to a simple bypass to get gating weights for each feature at each pixel. At last, the concatenated feature multiplied by the gating weights is fed into a 1 $\times$ 1 convolution to get the merged feature. Figure~\ref{fig:pose_gating} shows the architecture of the pose gating module. 

\vspace{1mm}
\noindent\textbf{Pose guidance network.}
We observed that the pose guider used in Animate Anyone~\cite{aa} is too shallow to effectively convey the pose information, while using ControlNet~\cite{controlnet} like Magic Animate~\cite{magic_animate} makes the whole model too heavy. Therefore, we propose a lightweight pose guidance network to convey pose information effectively and efficiently. More importantly, the pose guidance network is equipped with motion modules~\cite{animate_diff} to smooth video from the condition aspect. Specifically, the pose guidance network contains four blocks. Each block consists of two convolutional layers with SiLU activations and one motion module at the end. For the first three blocks, the stride of the first convolutional layers is set to 2 to downsample the feature map. The pose guidance network collects the input feature maps and the feature maps at the end of each block, feeds them into the corresponding zero-initialized convolutional layers, and sums the output with the corresponding feature maps in the main UNet encoder. As there are skip connections in UNet, the added condition information can naturally be conveyed to the UNet decoder. Please refer to Figure~\ref{fig:arch} for more details.
Thanks to the motion module, the pose guidance network not only can smooth video from the condition aspect, but is also more robust to incorrect pose frames in the pose sequence.

\noindent\textbf{Pose shuffle augmentation.}
The training of RealisDance is divided into two stages. The first stage is dedicated to image finetuning, the second stage focuses on motion learning. In the second stage, we introduce the pose shuffle augmentation to further improve the model robustness to incorrect pose frames. The pose shuffle augmentation randomly swaps two pose frames in the pose sequence, which forces the motion module to identify incorrect pose frames and utilize inter-frame information to obtain the correct generation.

\vspace{1mm}
\noindent\textbf{Comparisons with recent research.}
Consistent with recent research~\cite{aa, magic_animate, magic_pose}, RealisDance employs the reference UNet to ensure good character consistency. Beyond this, RealisDance incorporates a series of pose control modifications to facilitate robust generation, better hand fidelity, and smooth video.

The concurrent work Champ~\cite{champ} also applies multi-type poses to improve generation quality. Nonetheless, its reliance on SMPL as the sole source of depth, normal, and semantic poses introduces a significant vulnerability: if the SMPL sequence is corrupted, all other types of pose sequences will suffer corruption concurrently. In contrast, RealisDance incorporates three distinct types of pose sequences, significantly reducing the probability of simultaneous corruption. Besides, RealisDance introduces SMPL-CS to integrate 3D, depth, and continuous semantic information into a unified pose representation, eliminating redundant pose conditions. RealisDance also leverages HaMeR sequences to ensure realistic hand generation.

\section{Experiments}
\subsection{Implementation Details}
RealisDance is trained in two stages: image finetuning and motion learning. In image finetuning, both the main UNet and the reference UNet are initialized from Real Vision v5.1, and all components are optimizable except for DINOv2~\cite{dinov2} and motion modules. In motion learning, only motion modules, which are initialized from AnimateDiff~\cite{animate_diff}, are optimizable. The pose shuffle rate is set to 5e-2. For both two stages, the learning rate is set to 5e-5. The zero-SNR~\cite{zero_snr}, min-SNR~\cite{min_snr}, and classifier-free guidance (CFG)~\cite{cfg} are enabled. The unconditional drop rate is set to 1e-2. We use window shifting in temporal for long sequence generation.

\begin{figure}[t]
    \centering
    \addtolength{\tabcolsep}{-5.5pt}
    \begin{tabular}{ccccc}
        & Frame 1 & Frame 2 & Frame 3 & Frame 4\\
        \rotatebox{90}{~~~~~~~AA} &
        \includegraphics[width=0.22\linewidth]{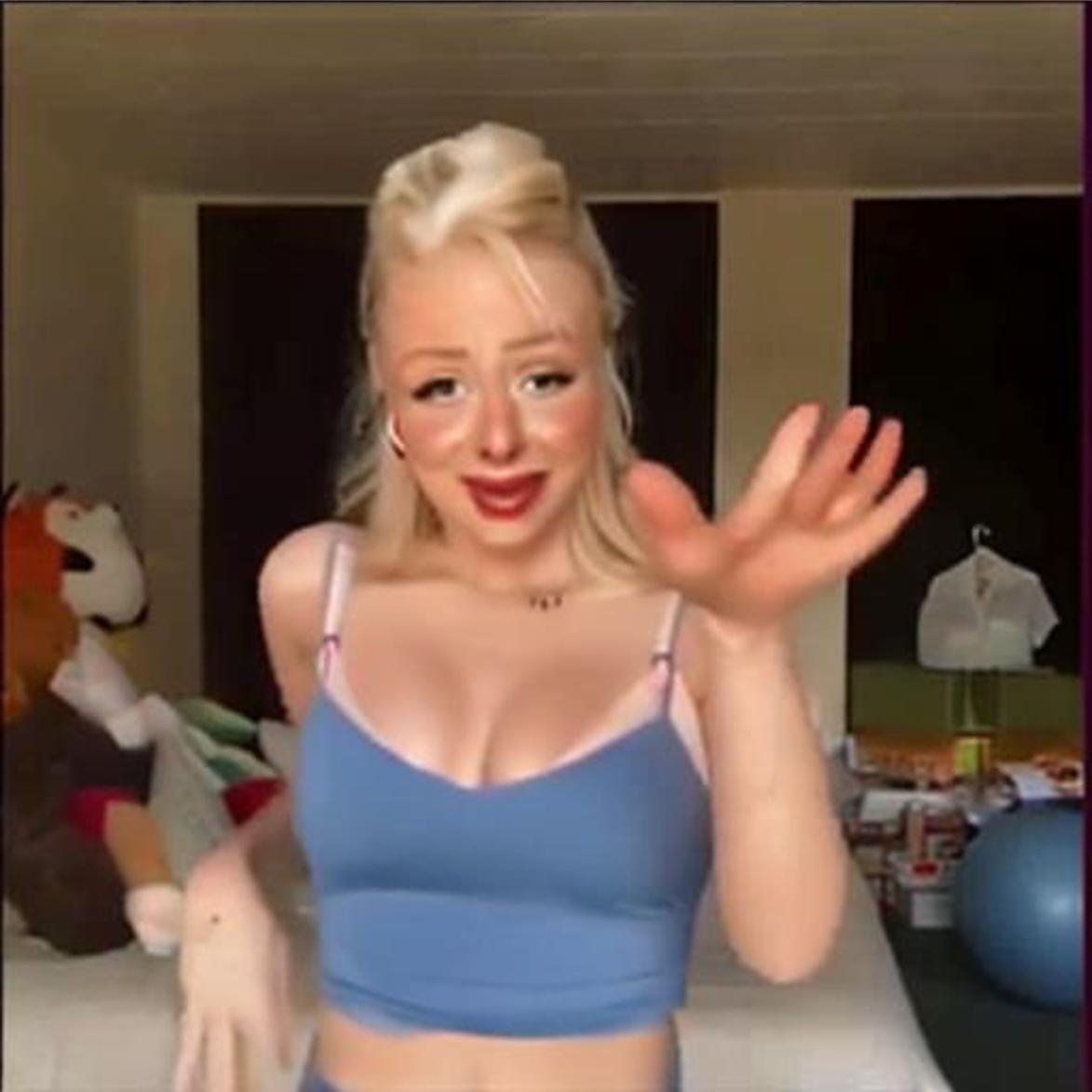} &
        \includegraphics[width=0.22\linewidth]{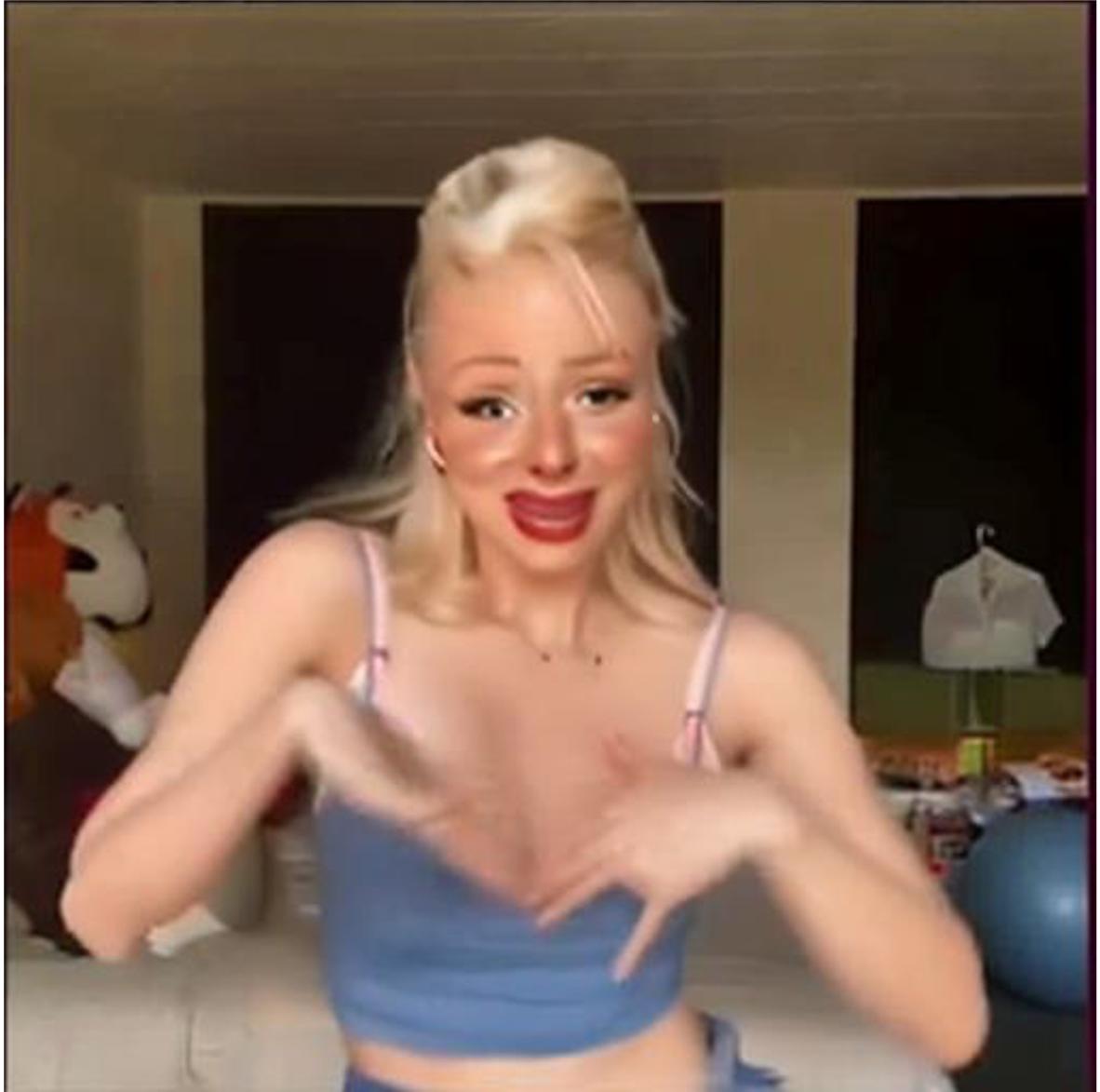} & \includegraphics[width=0.22\linewidth]{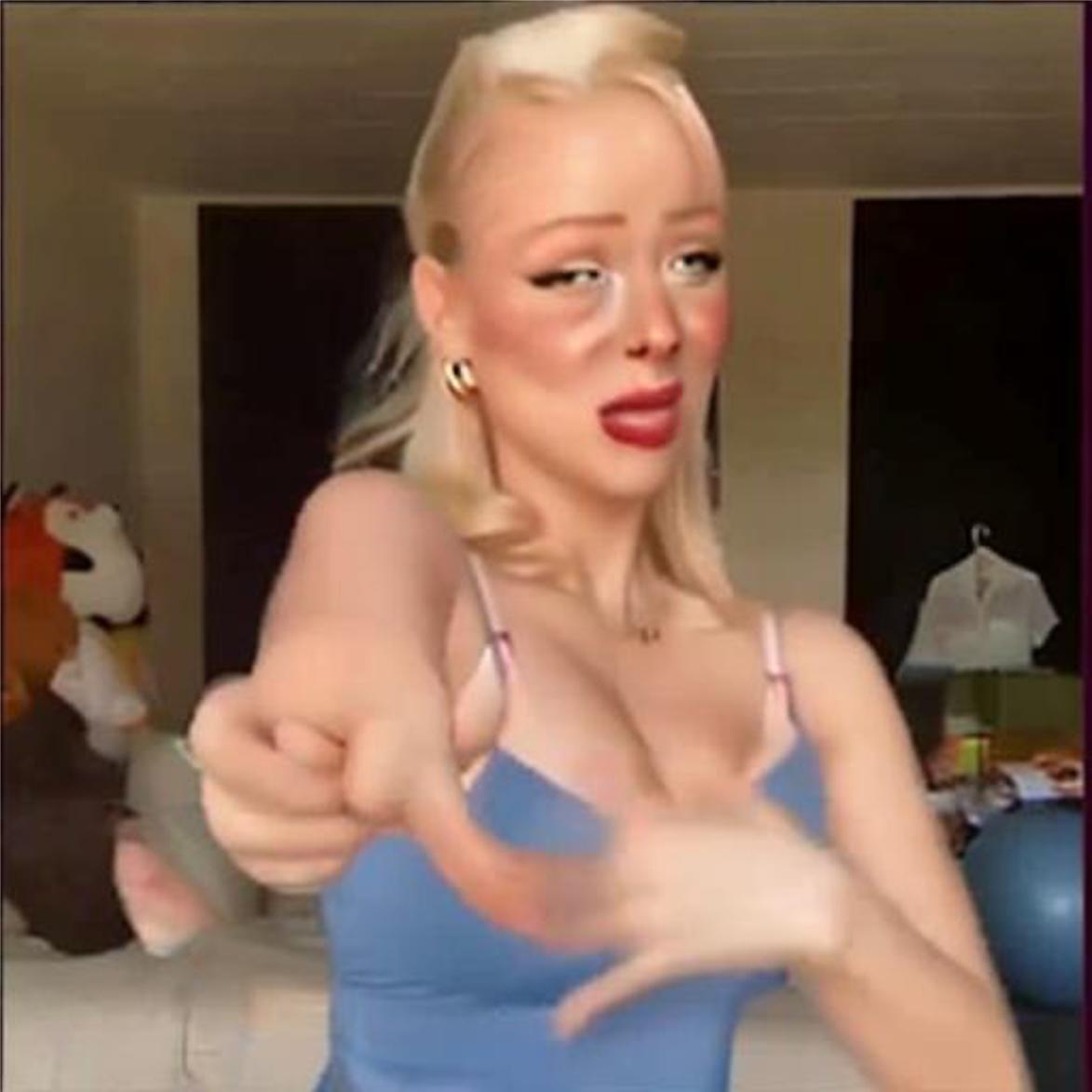} & \includegraphics[width=0.22\linewidth]{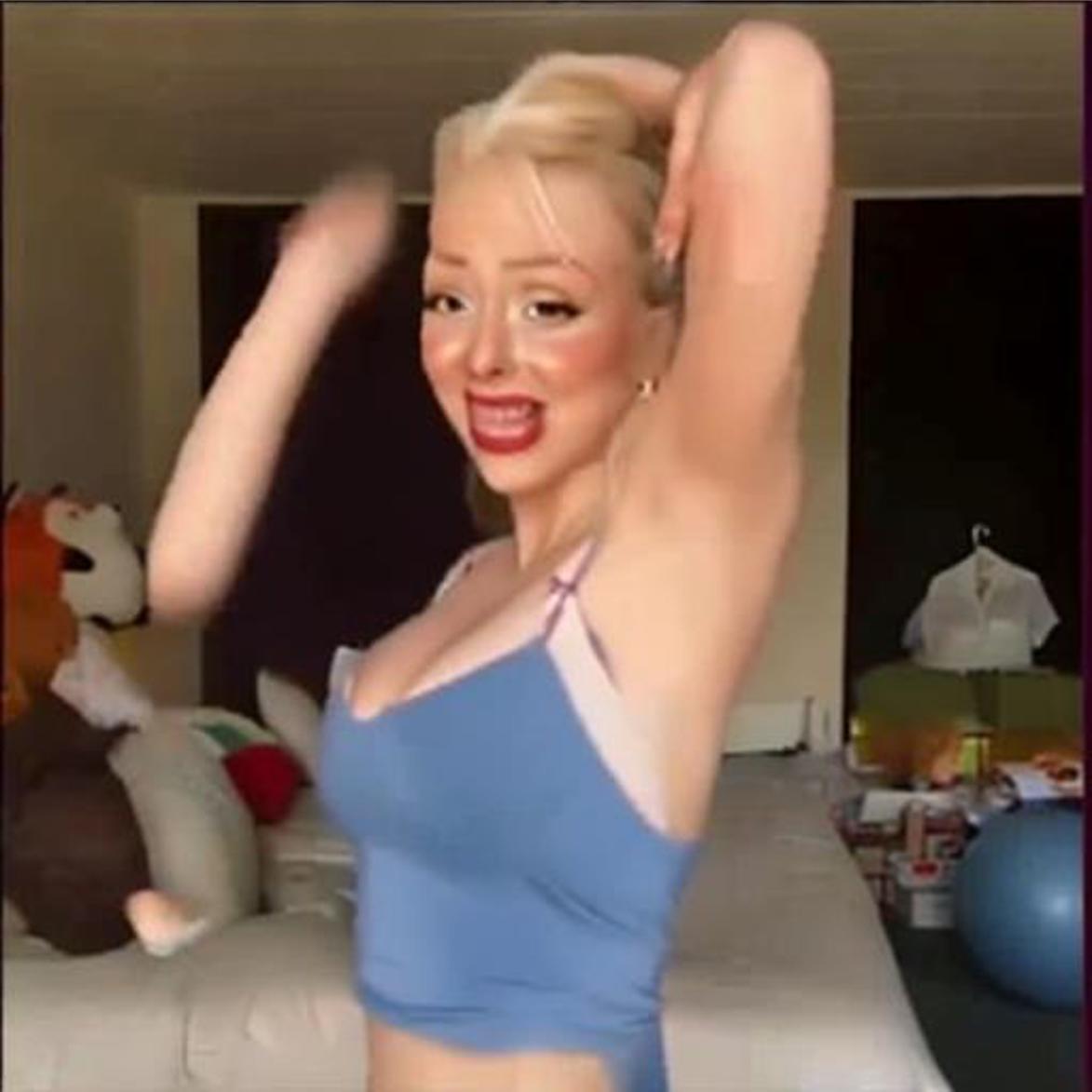} \\
        \rotatebox{90}{~~~~~~{Ours}} &
        \includegraphics[width=0.22\linewidth]{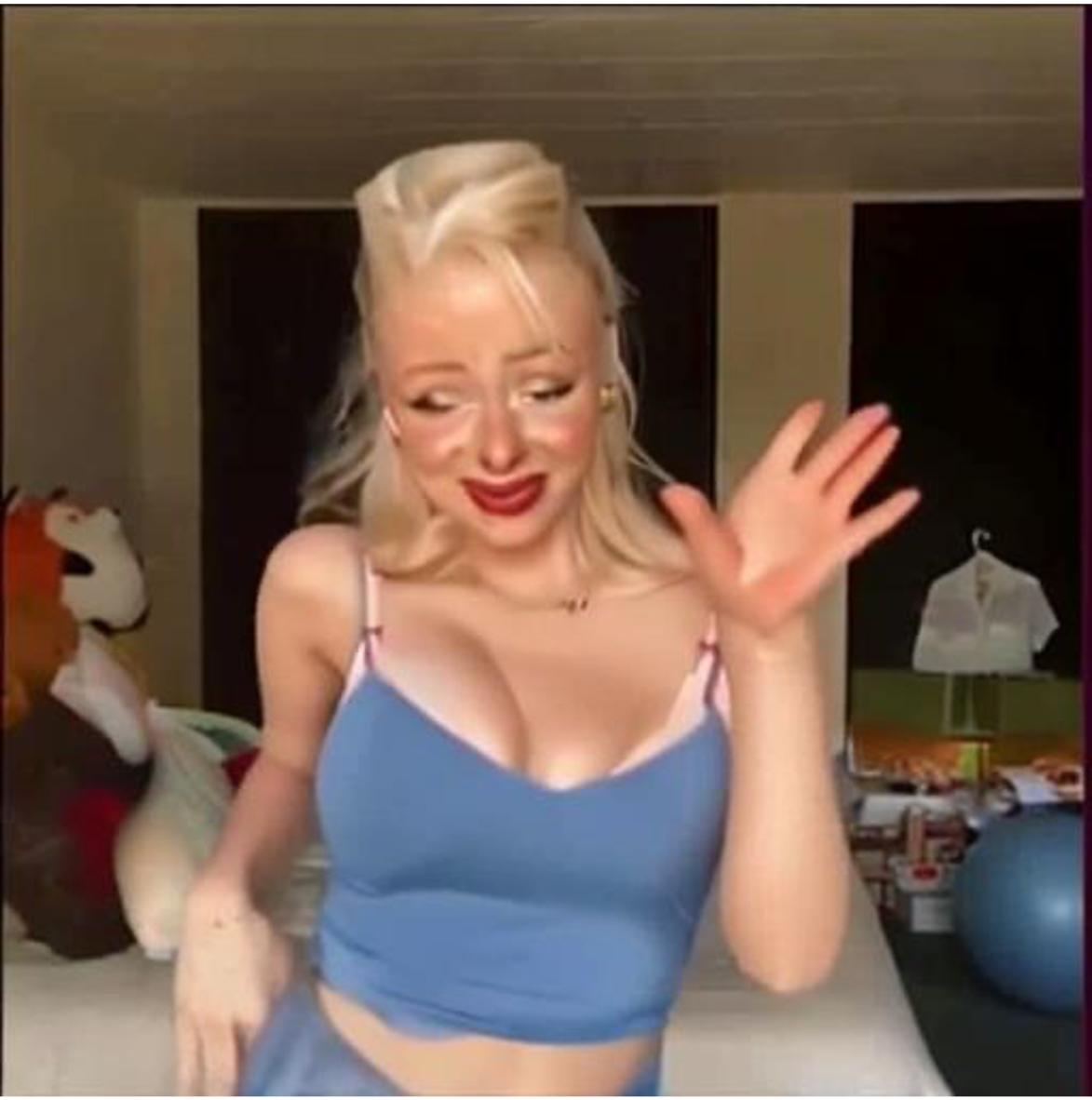} &
        \includegraphics[width=0.22\linewidth]{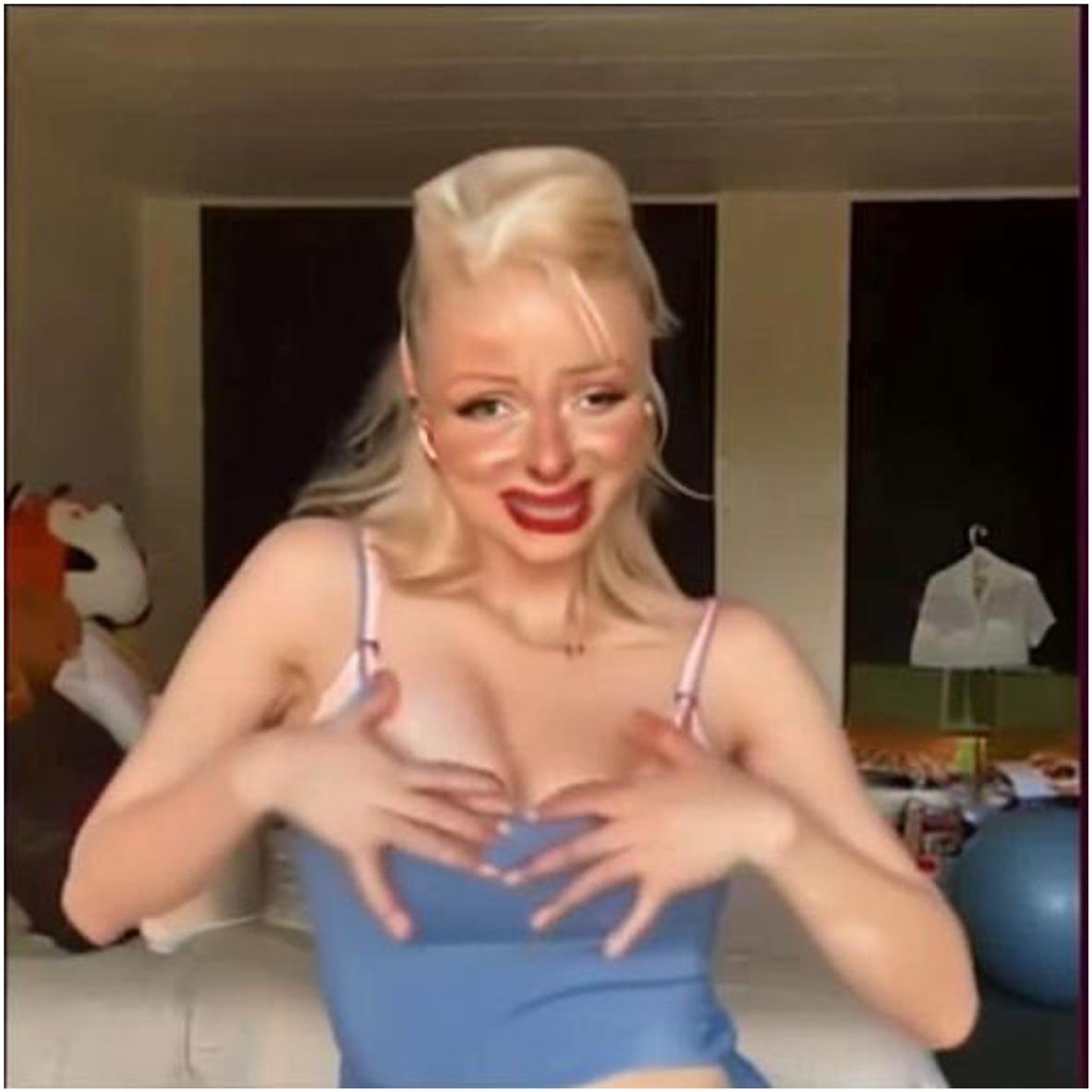} & \includegraphics[width=0.22\linewidth]{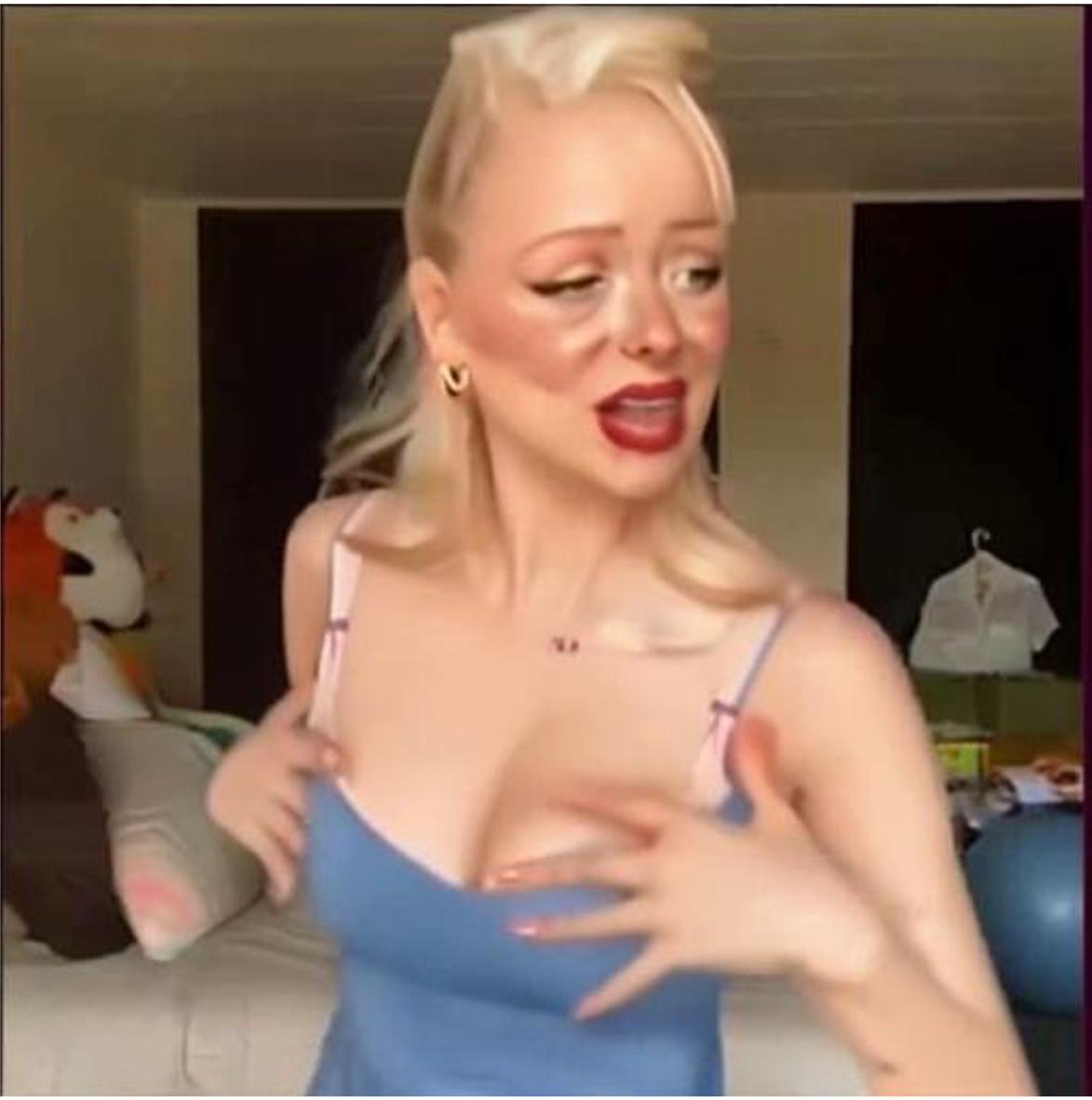} & \includegraphics[width=0.22\linewidth]{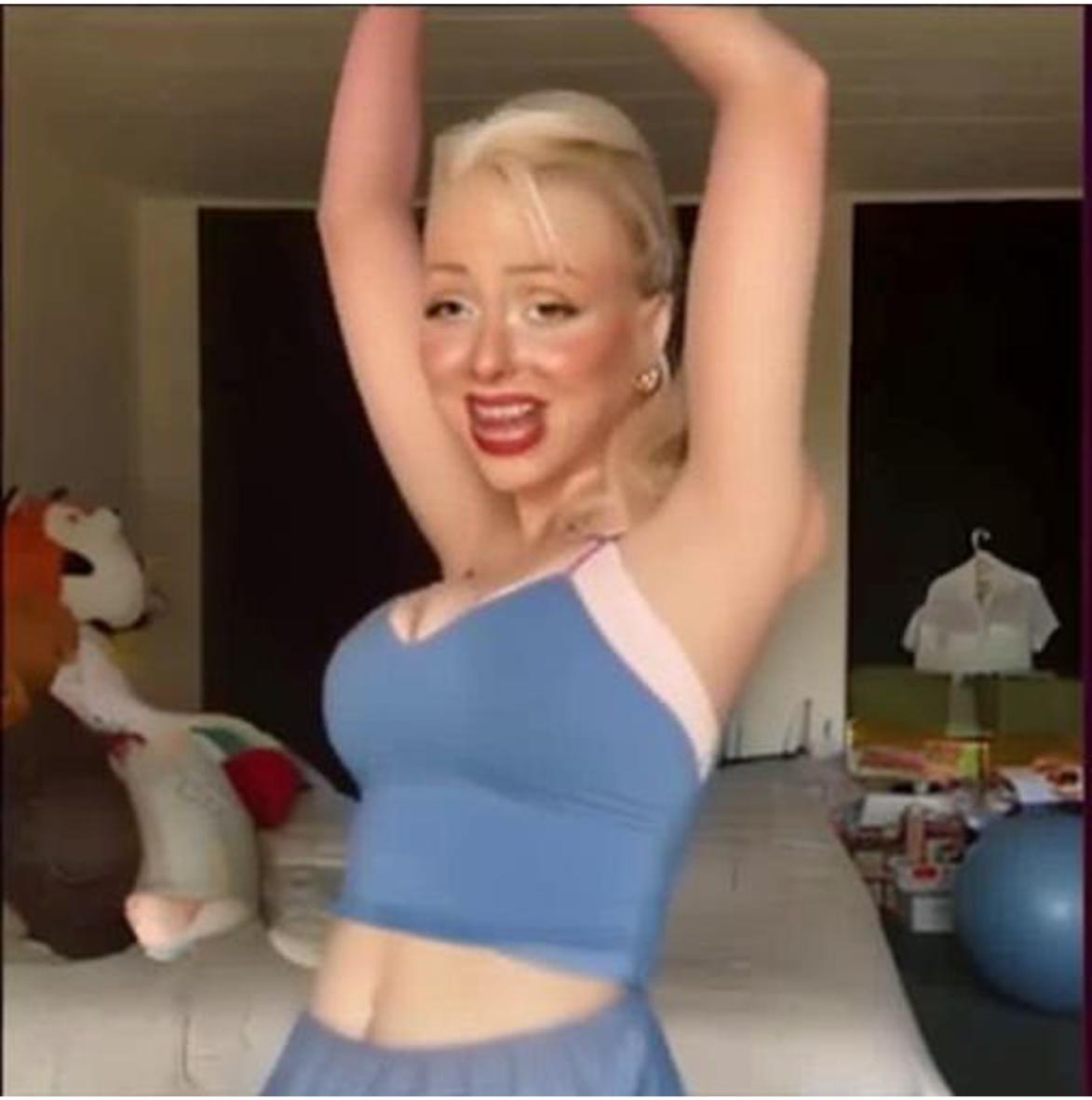} \\
        \rotatebox{90}{~~~~~~~AA} &
        \includegraphics[width=0.22\linewidth]{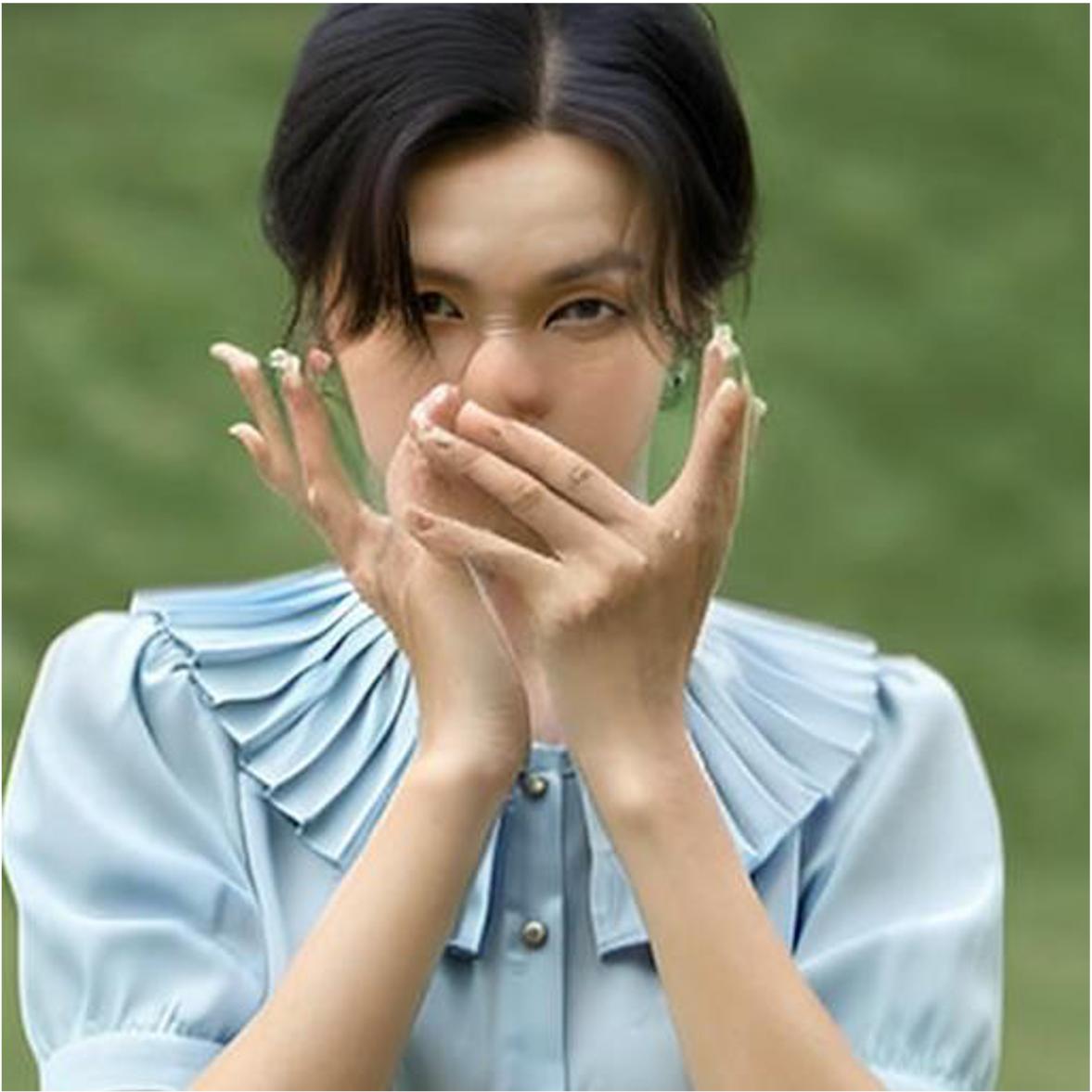} &
        \includegraphics[width=0.22\linewidth]{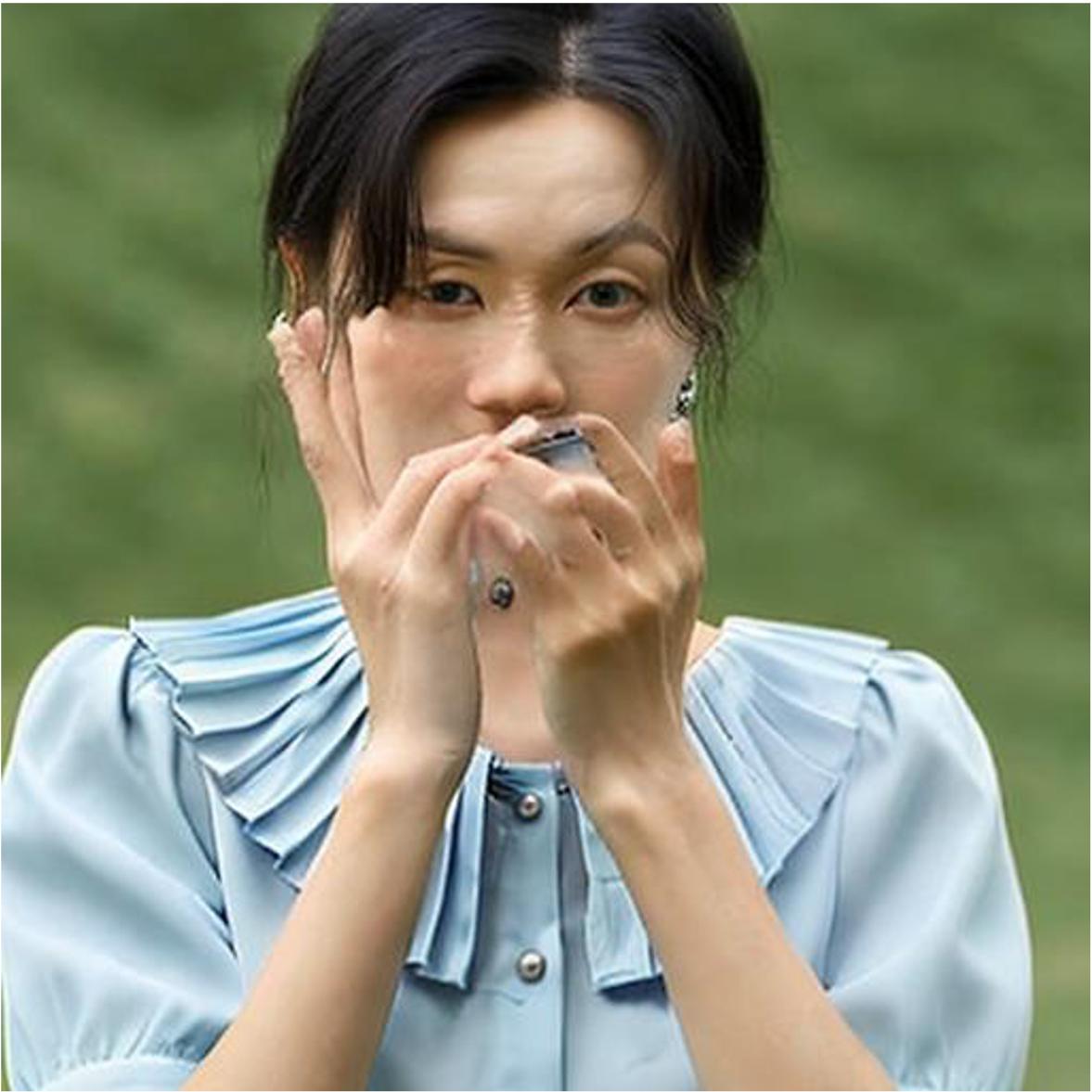} & \includegraphics[width=0.22\linewidth]{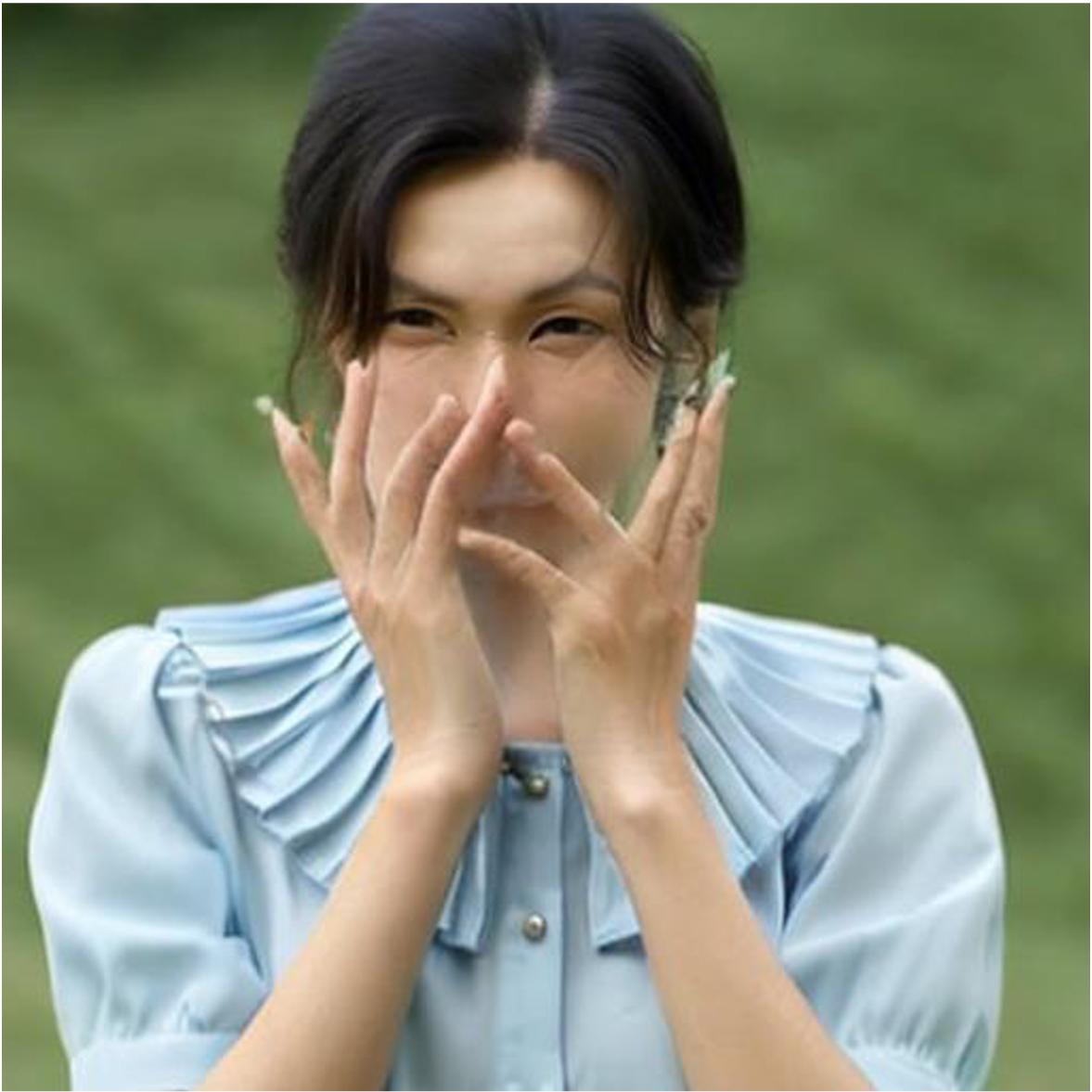} & \includegraphics[width=0.22\linewidth]{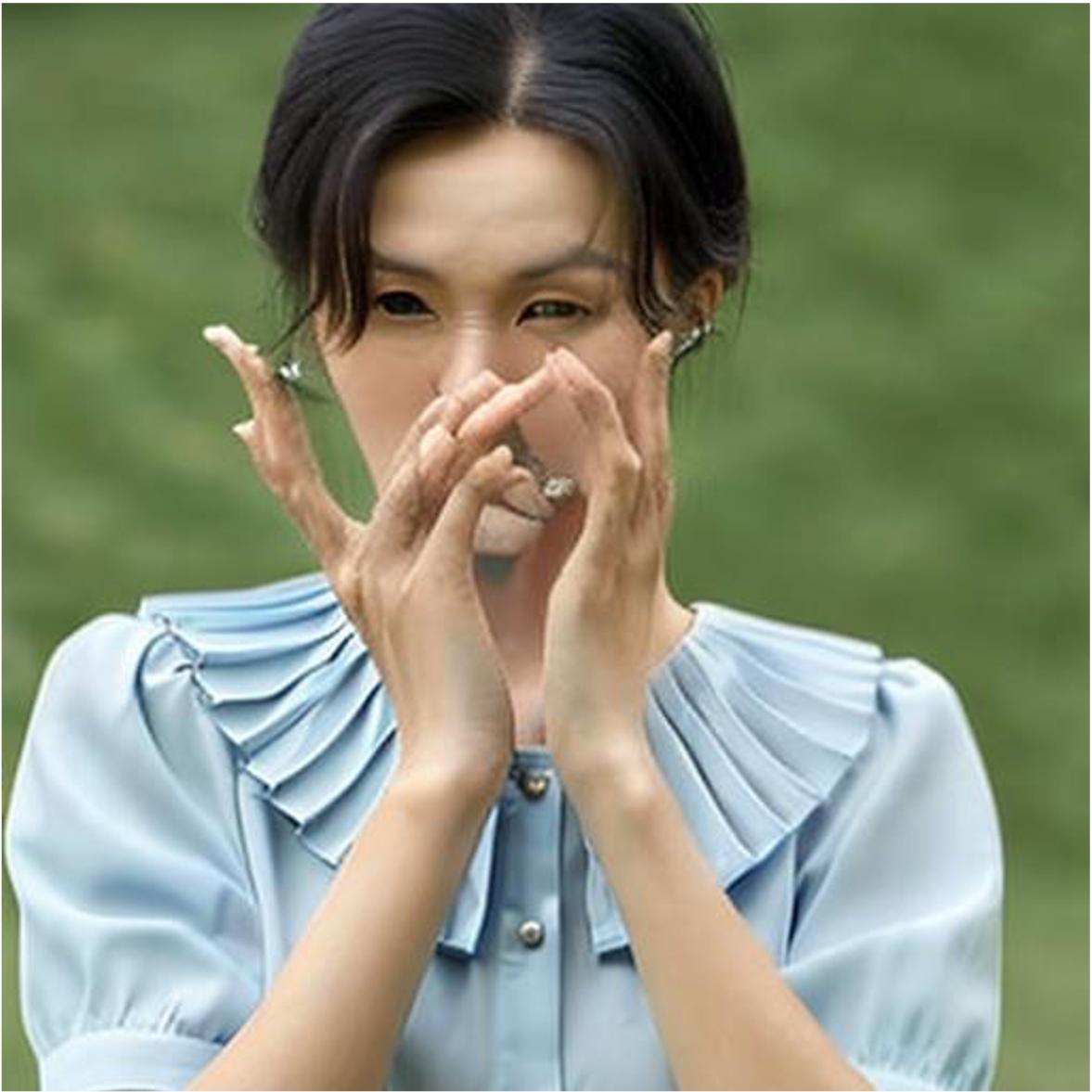} \\
        \rotatebox{90}{~~~~~~{Ours}} &
        \includegraphics[width=0.22\linewidth]{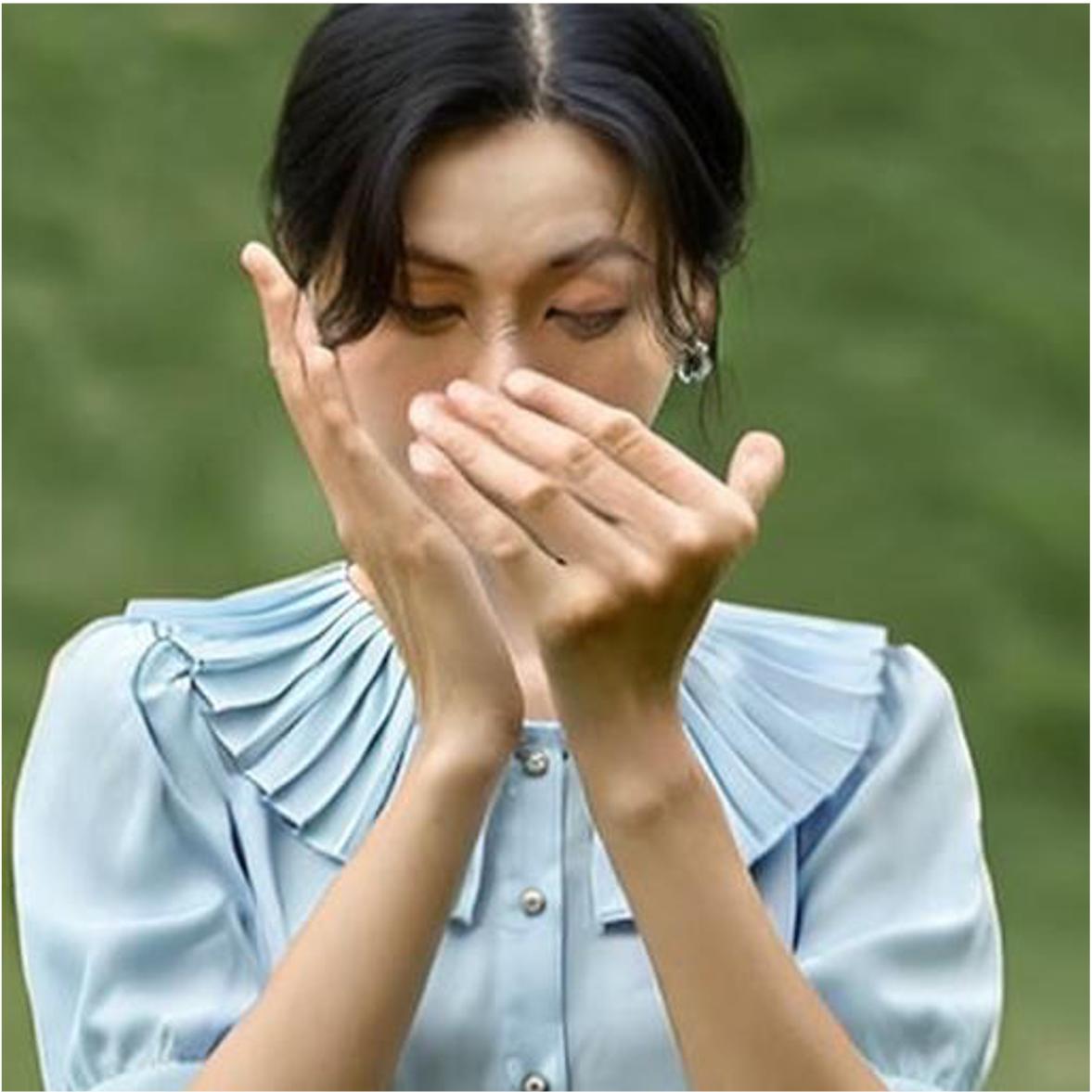} &
        \includegraphics[width=0.22\linewidth]{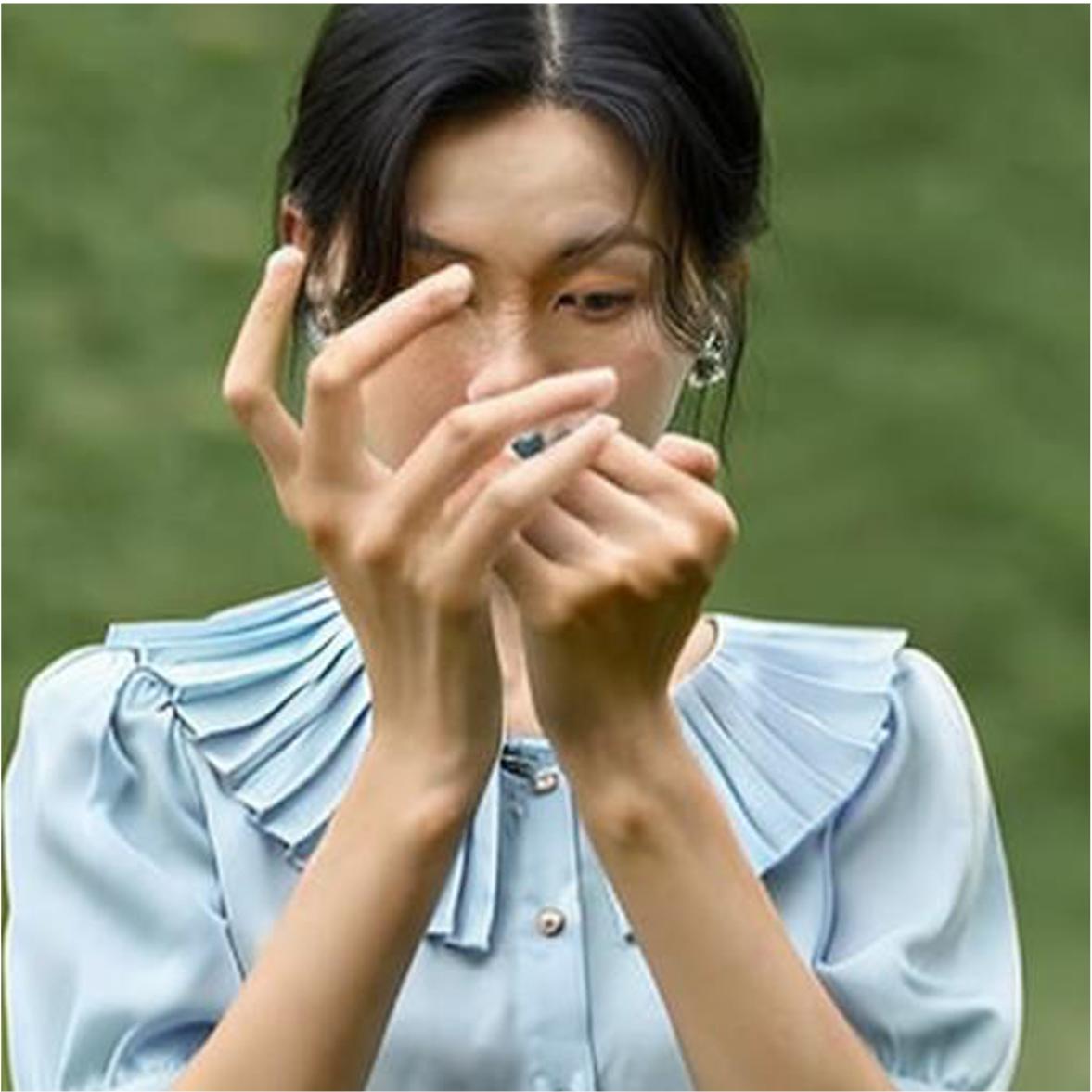} & \includegraphics[width=0.22\linewidth]{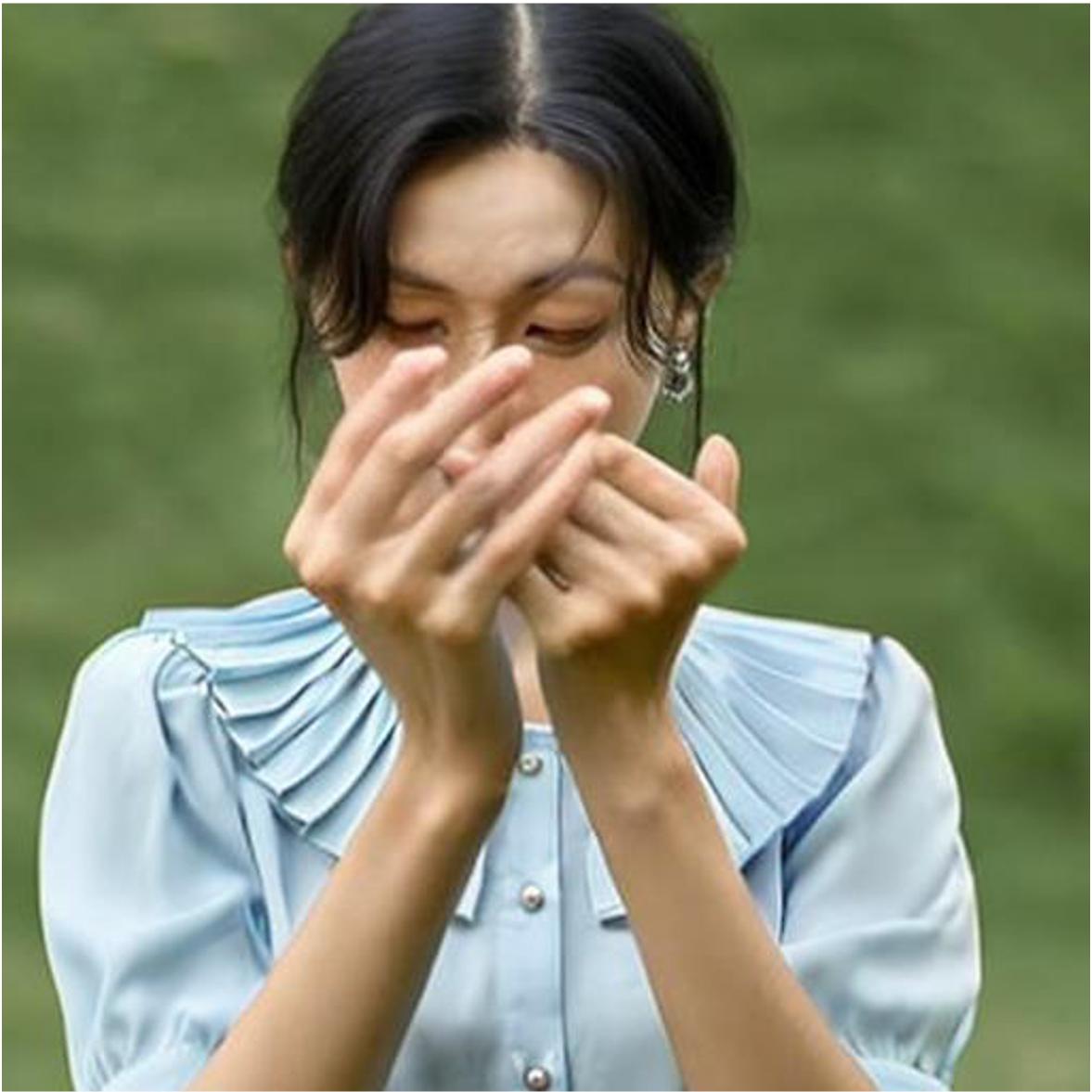} & \includegraphics[width=0.22\linewidth]{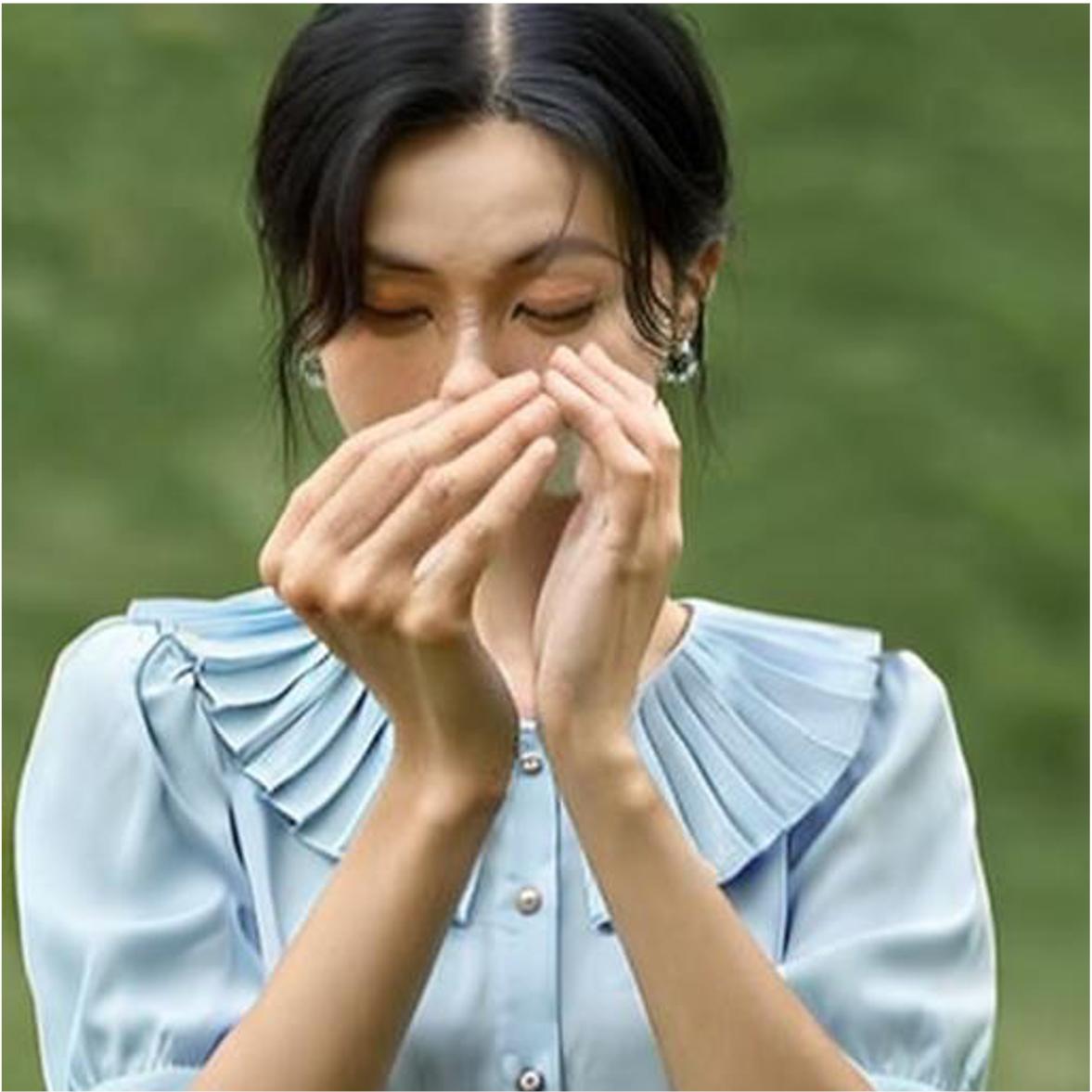} \\
        \rotatebox{90}{~~~~~~~AA} &
        \includegraphics[width=0.22\linewidth]{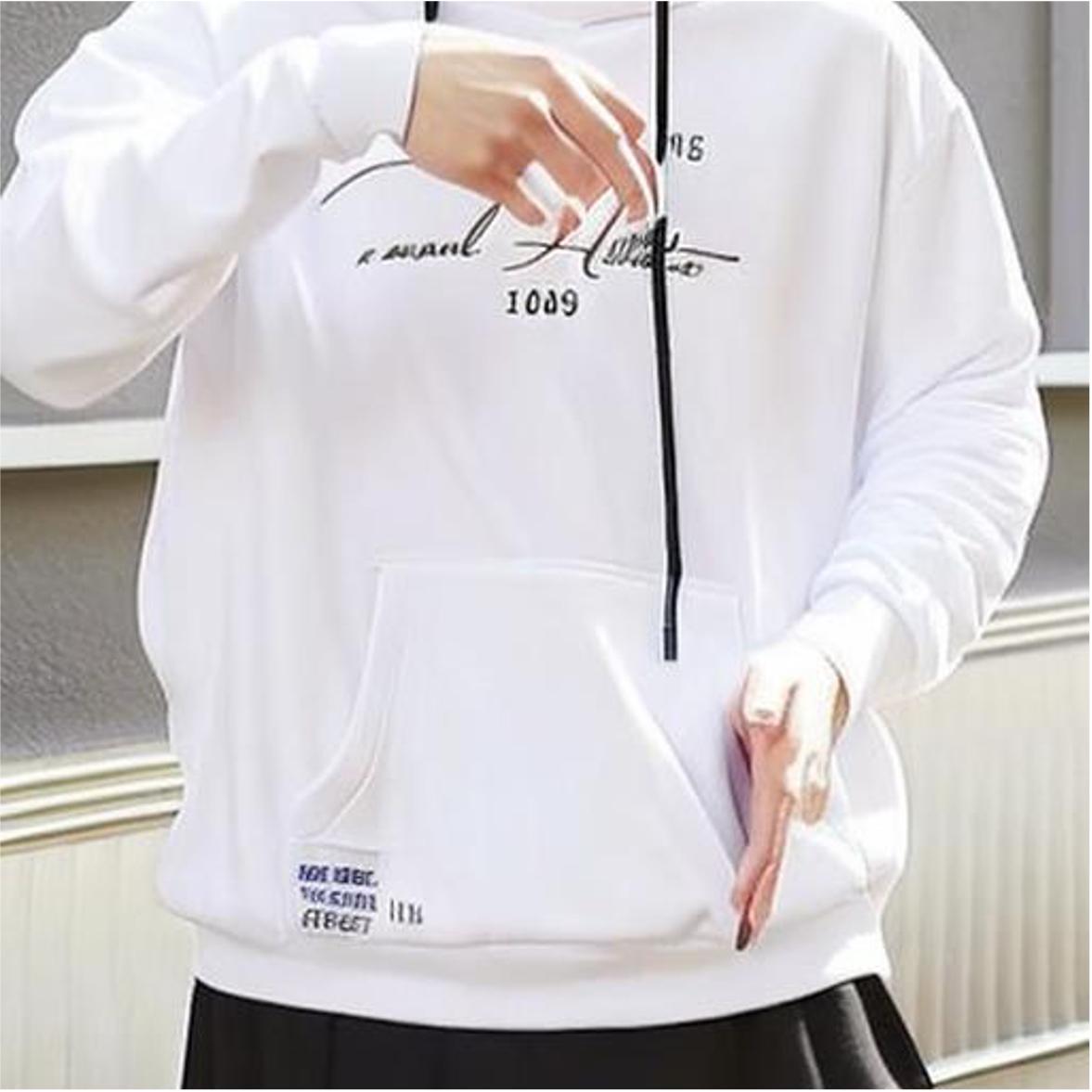} &
        \includegraphics[width=0.22\linewidth]{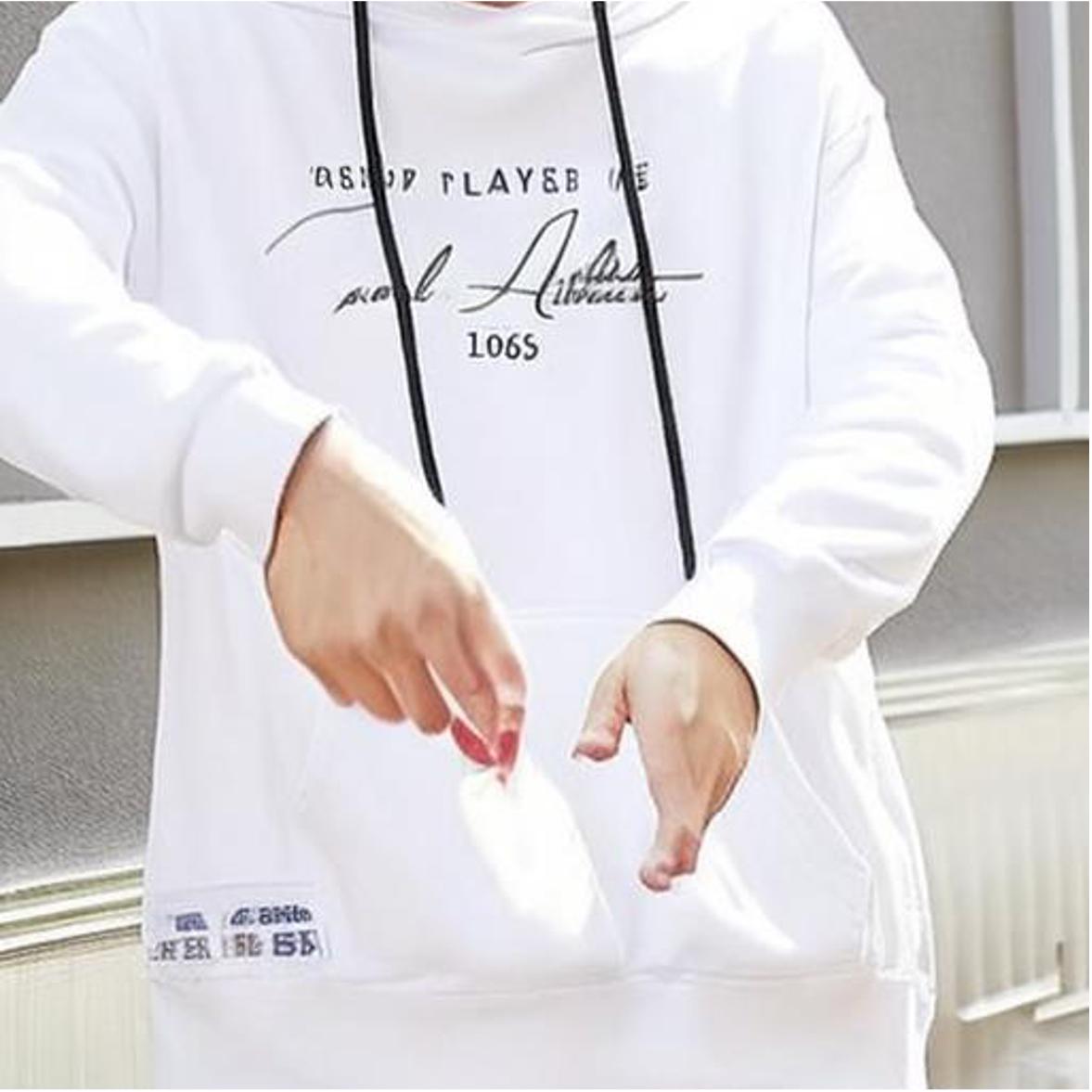} & \includegraphics[width=0.22\linewidth]{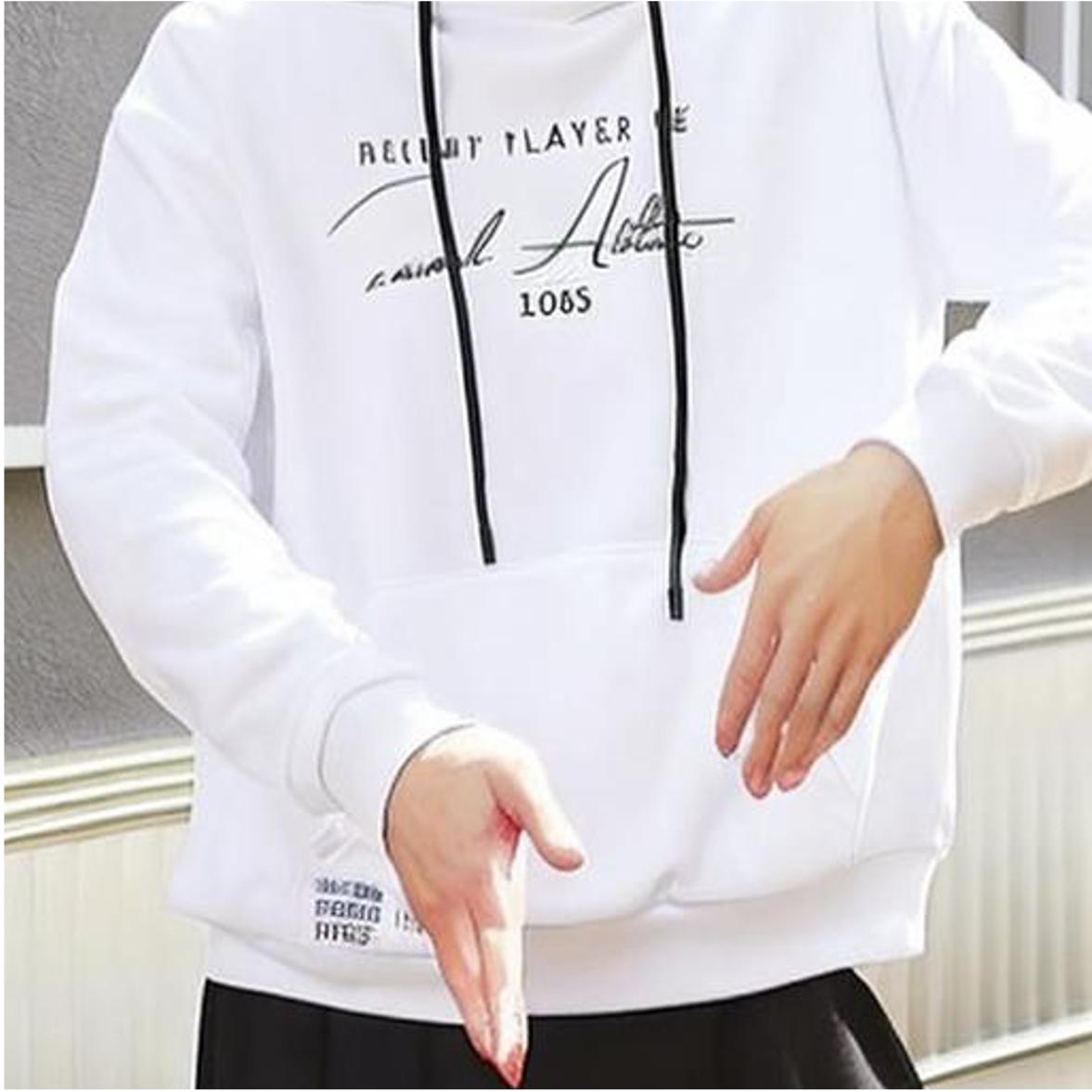} & \includegraphics[width=0.22\linewidth]{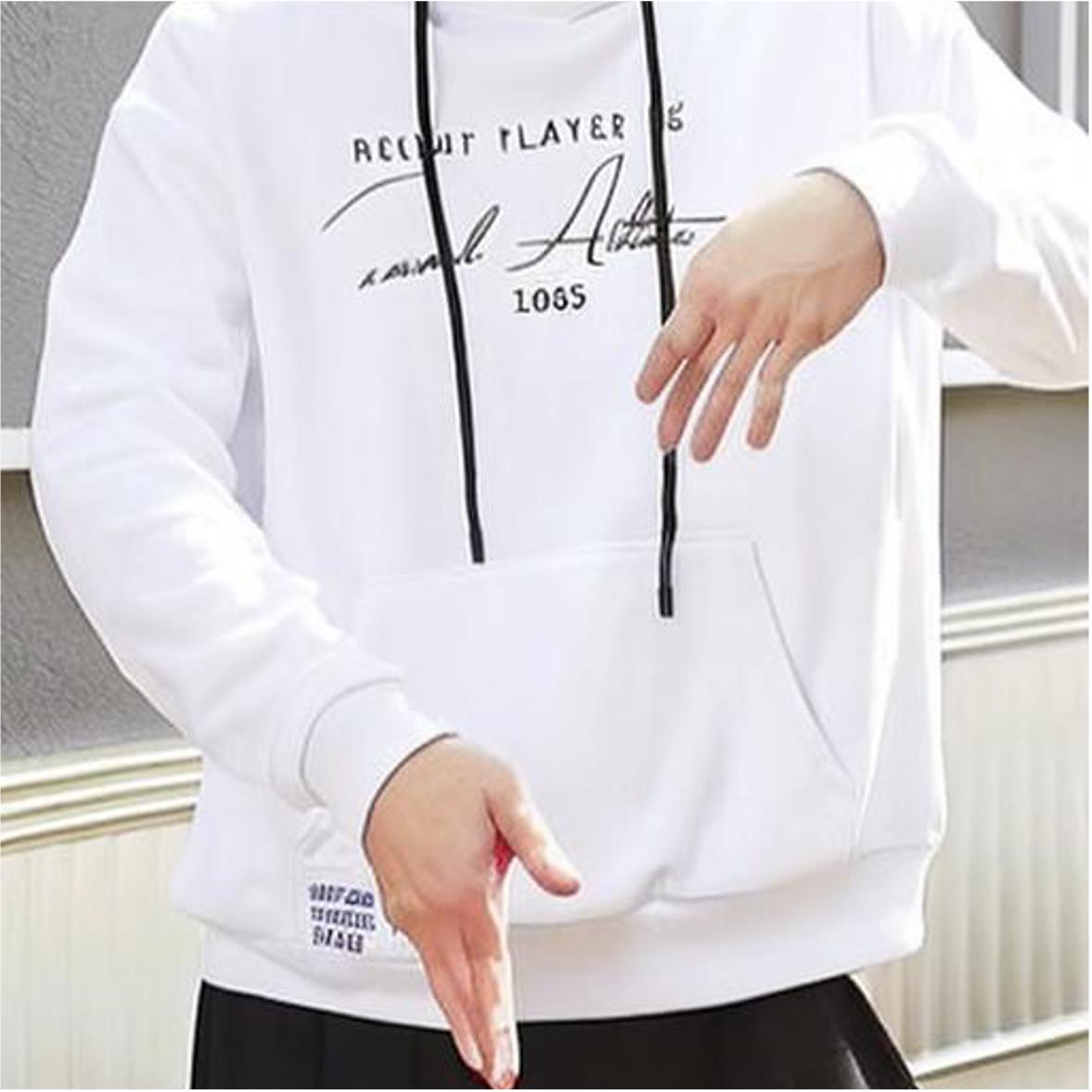} \\
        \rotatebox{90}{~~~~~~{Ours}} &
        \includegraphics[width=0.22\linewidth]{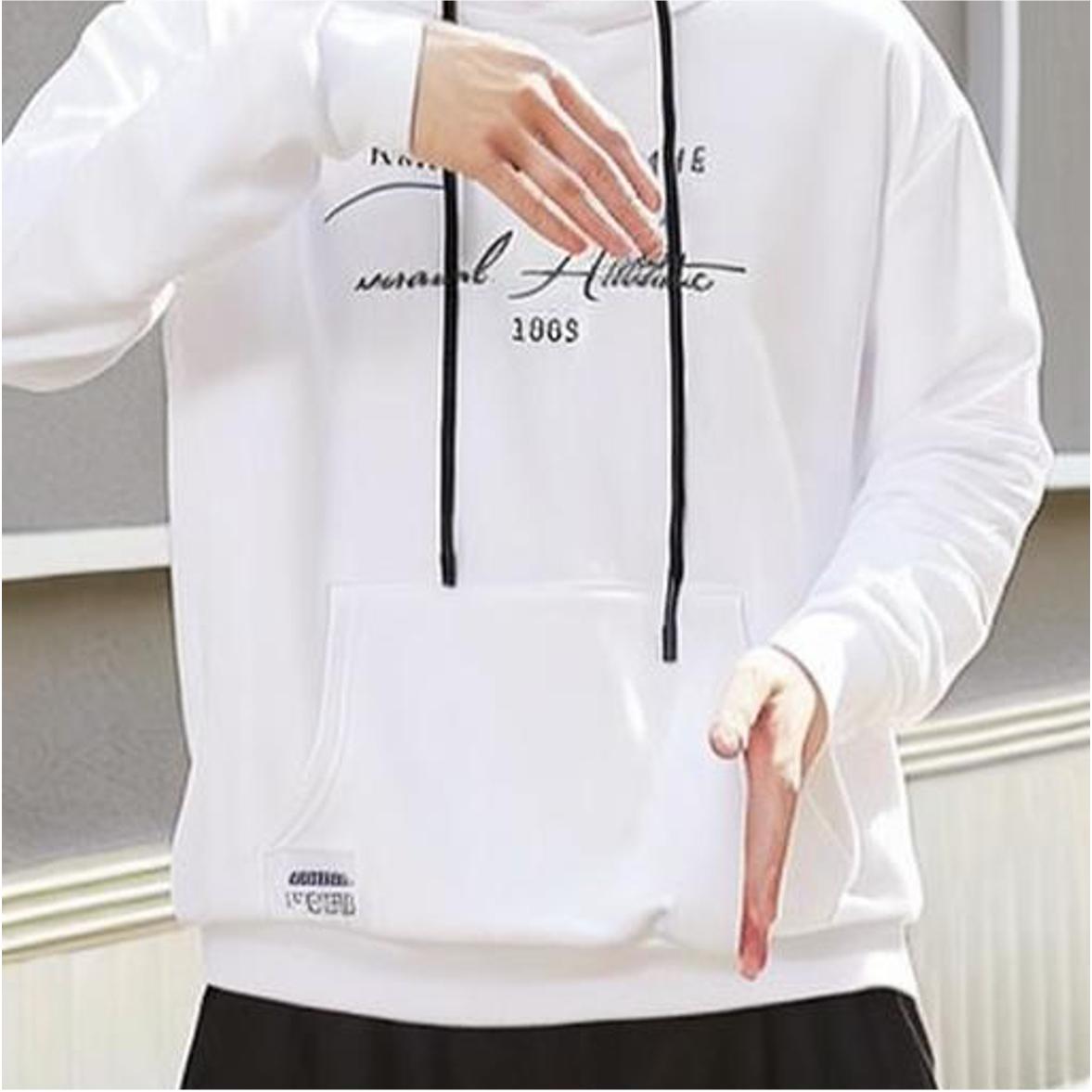} &
        \includegraphics[width=0.22\linewidth]{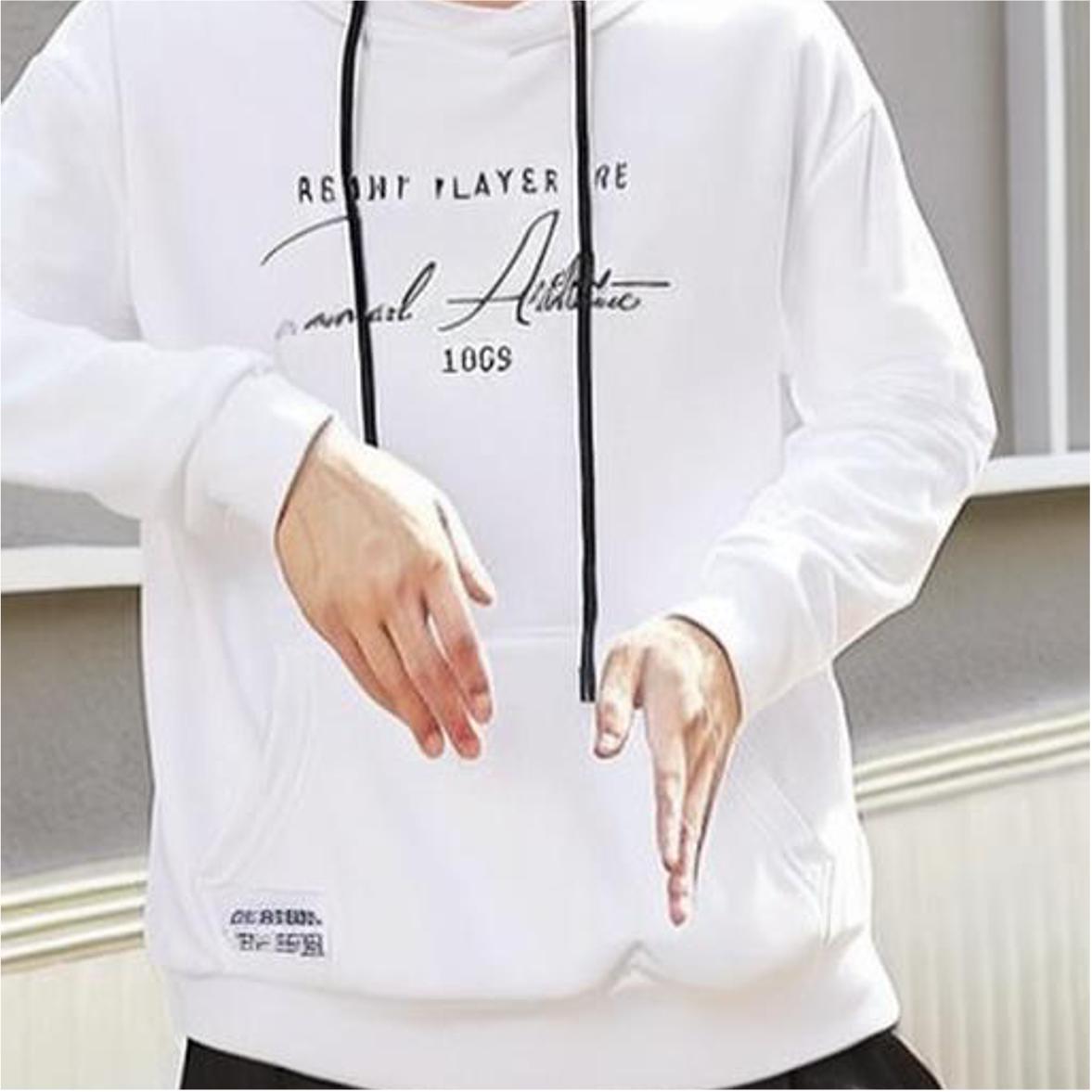} & \includegraphics[width=0.22\linewidth]{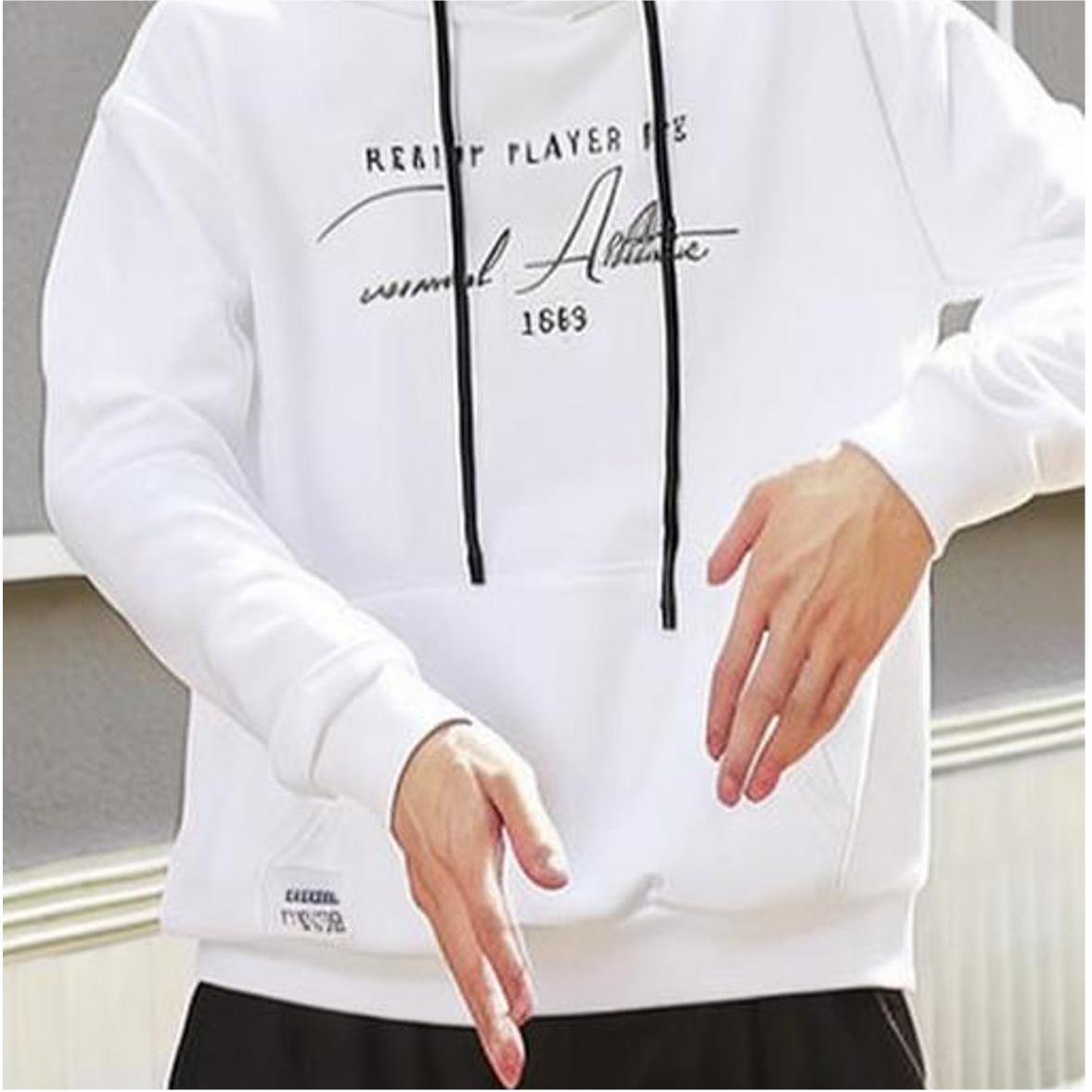} & \includegraphics[width=0.22\linewidth]{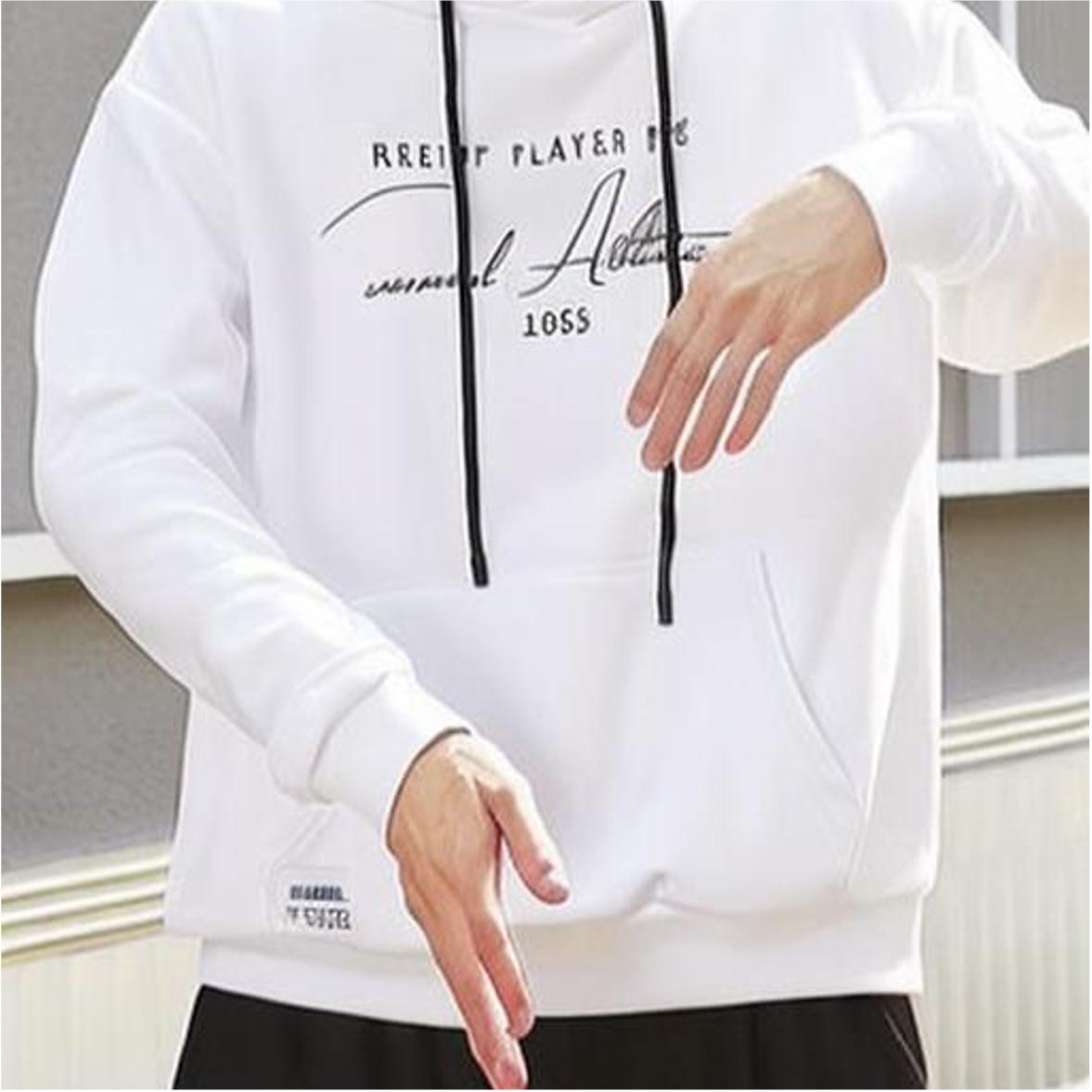} \\
        \rotatebox{90}{~~~~~~~AA} &
        \includegraphics[width=0.22\linewidth]{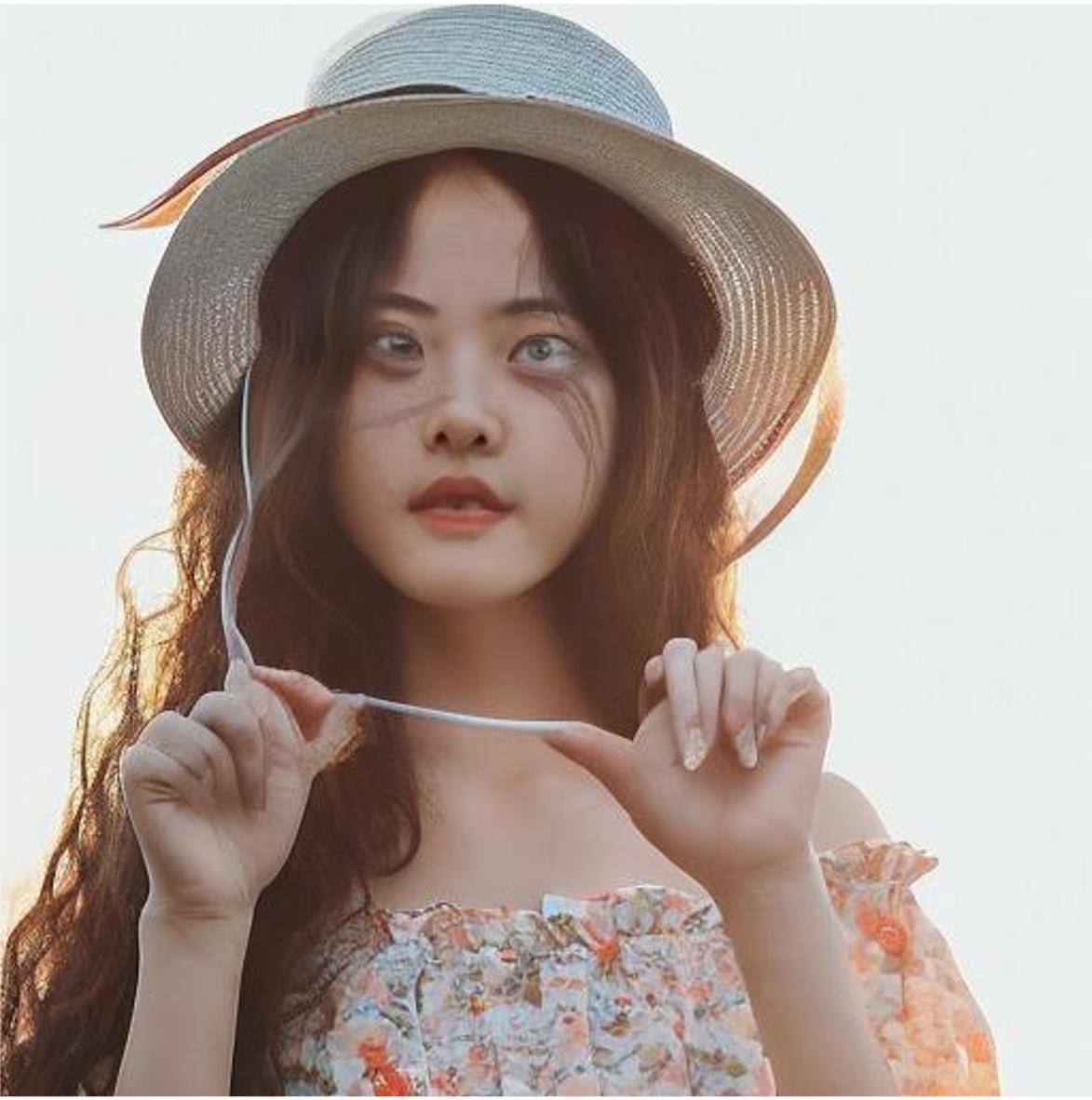} &
        \includegraphics[width=0.22\linewidth]{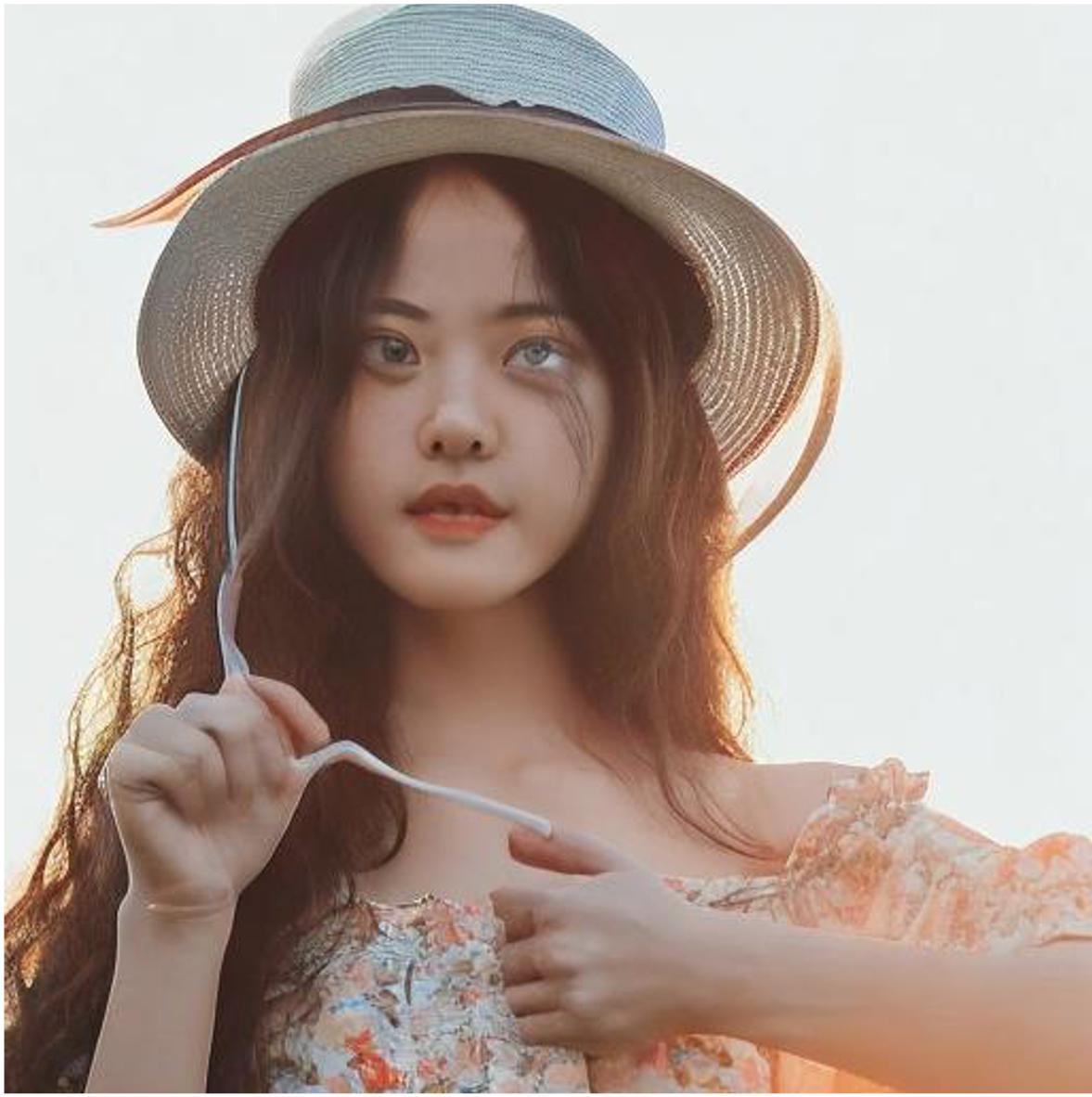} & \includegraphics[width=0.22\linewidth]{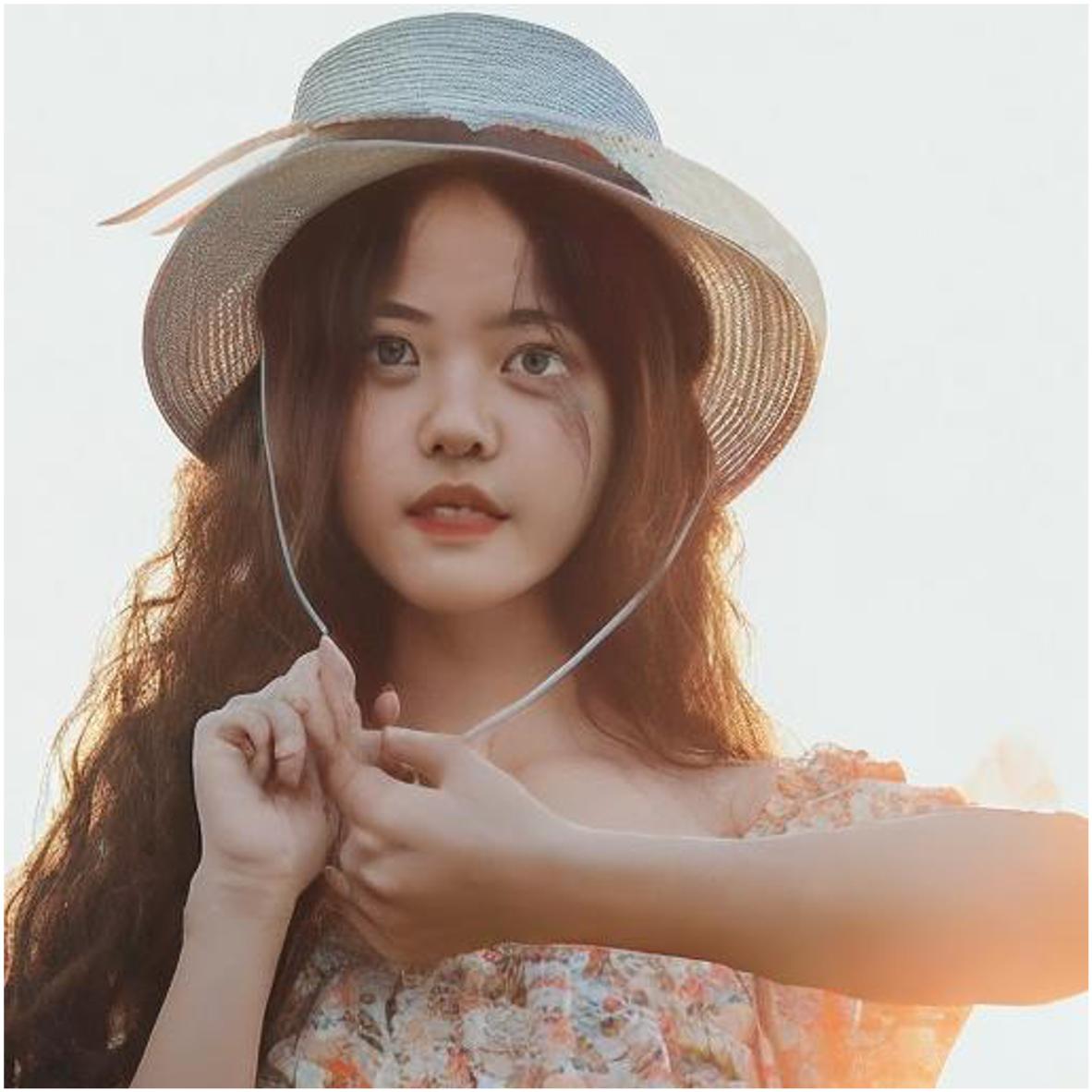} & \includegraphics[width=0.22\linewidth]{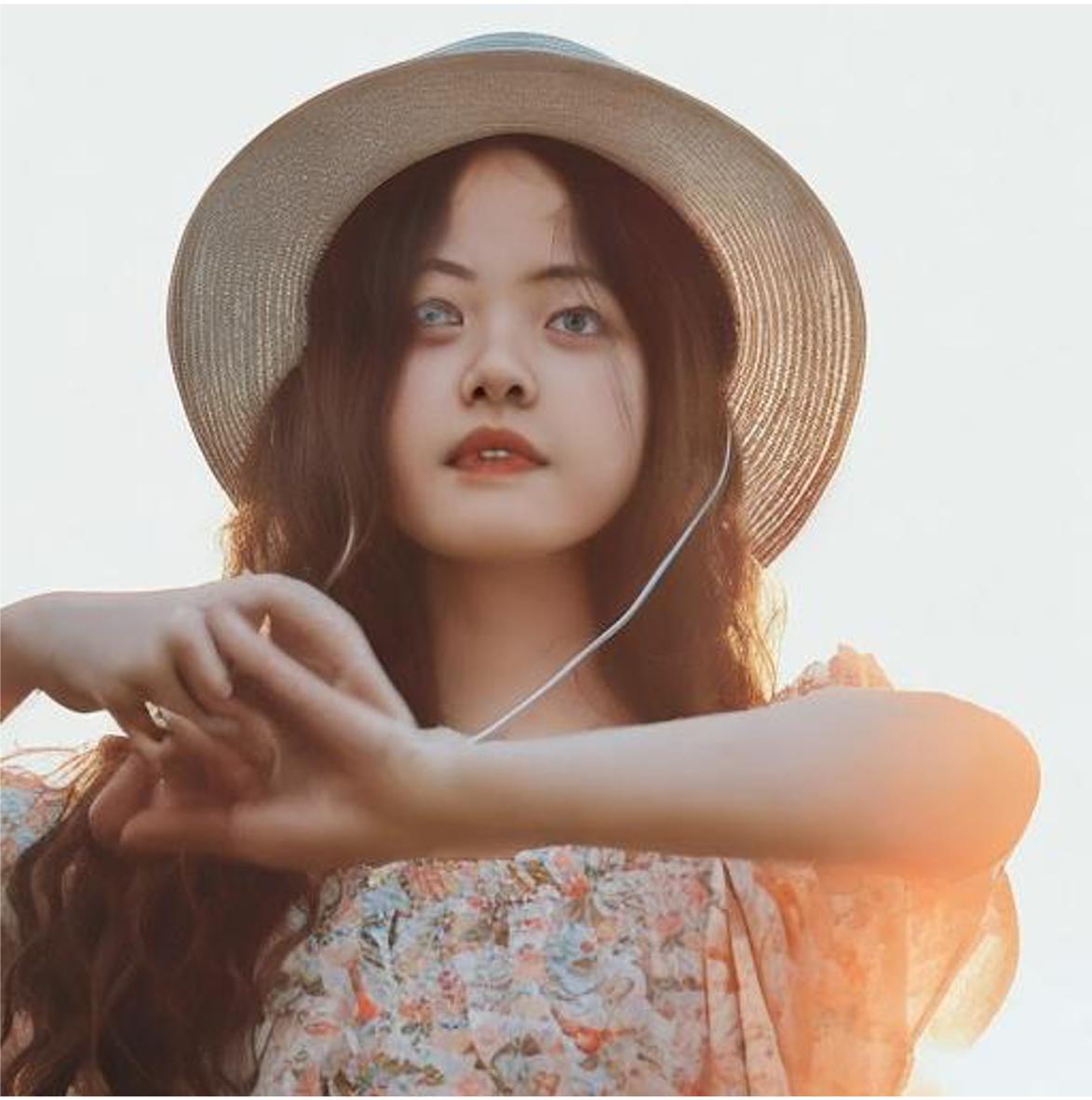} \\
        \rotatebox{90}{~~~~~~{Ours}} &
        \includegraphics[width=0.22\linewidth]{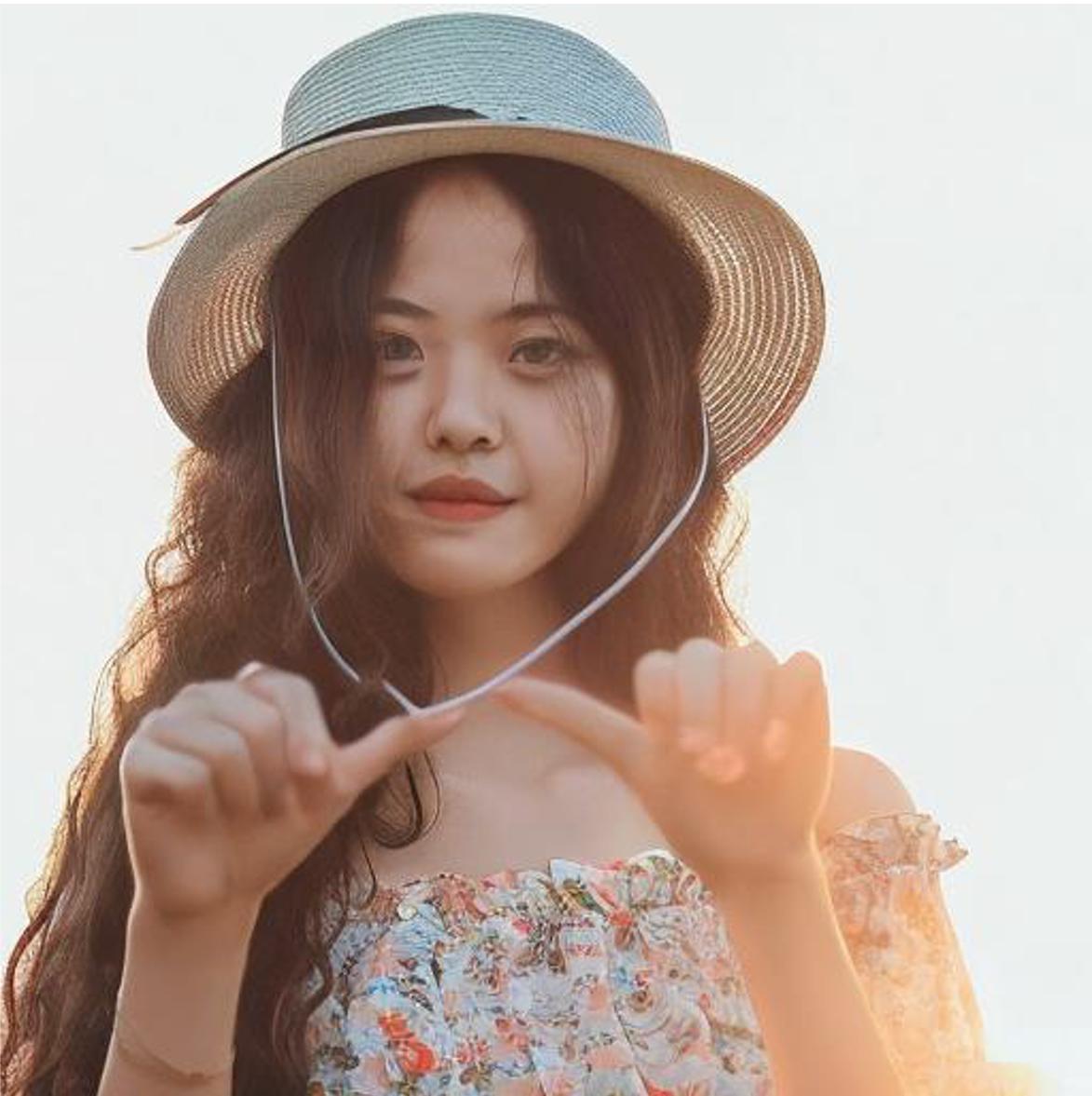} &
        \includegraphics[width=0.22\linewidth]{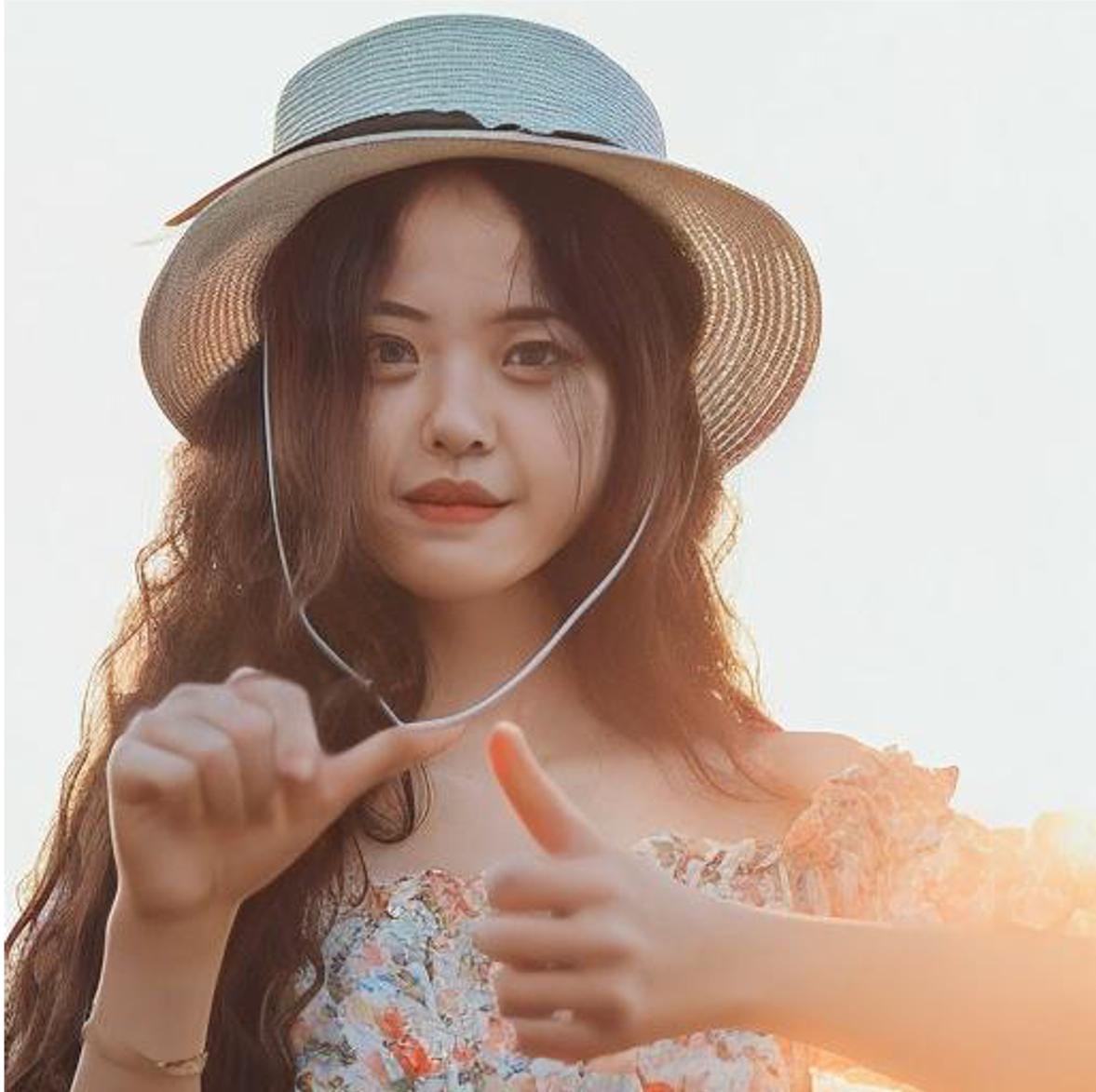} & \includegraphics[width=0.22\linewidth]{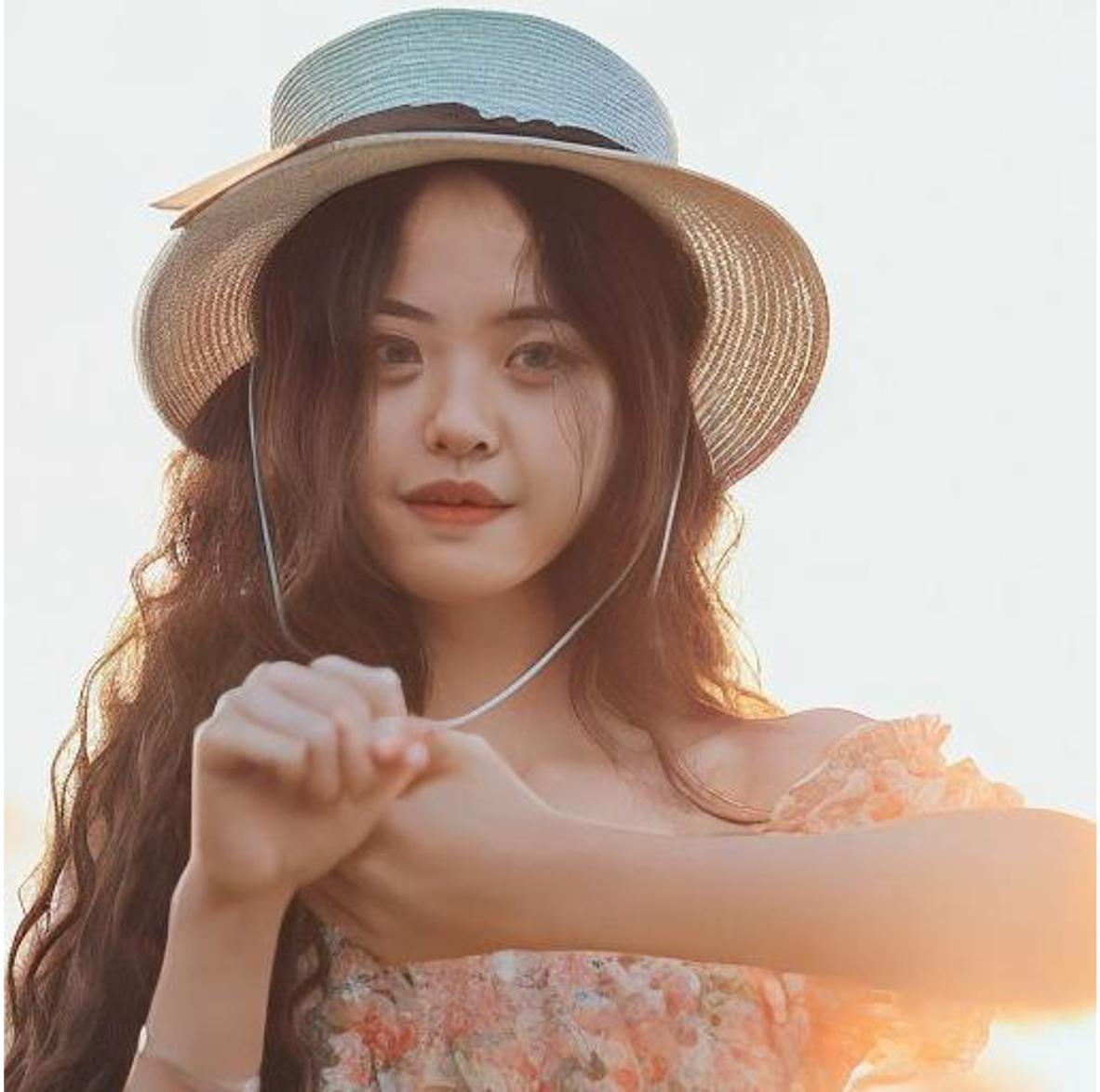} & \includegraphics[width=0.22\linewidth]{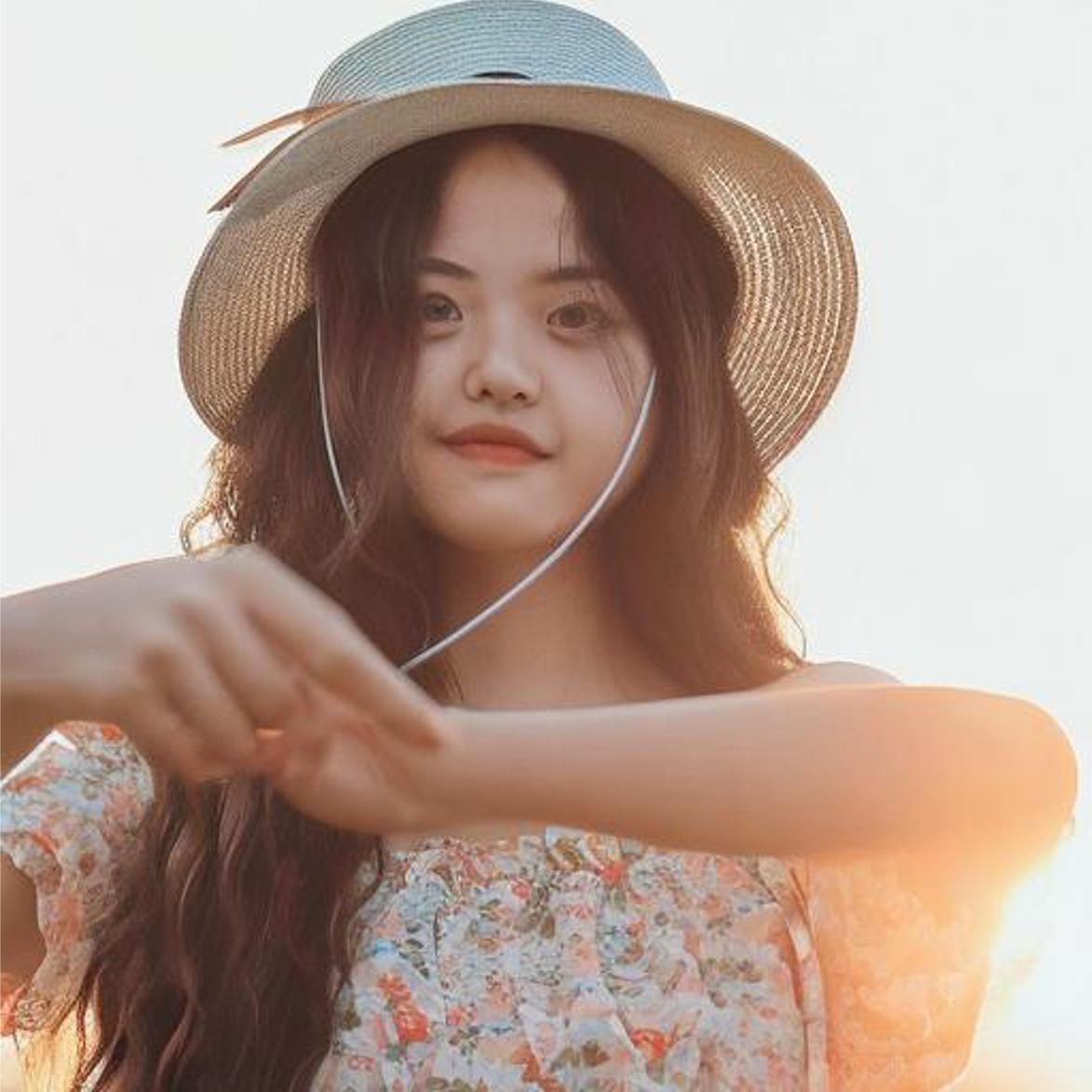} \\
    \end{tabular}
    \caption{Comparisons of overall quality between RealisDance and our reproduced Animate Anyone.}
    \label{fig:overall_quality}
\end{figure}

\subsection{Qualitative Comparisons}
In this section, we qualitatively evaluate the proposed RealisDance by comparing it with our reproduced Animate Anyone from overall quality, generation stability, hands quality, and video smoothness. The reference images are collected from the dataset and the internet, which can only be used for the academy.

\vspace{1mm}
\noindent\textbf{Overall quality.}
Figure~\ref{fig:overall_quality} shows comparisons of overall quality. It can be seen that RealisDance significantly improves hand quality and its results contain fewer pose artifacts due to better robustness. See case 1 frame 3 and frame 4, our reproduced Animate Anyone generates artifacts due to the corrupted DWPose, while RealisDance can still obtain correct results driven by the uncorrupted SMPL-CS.

\begin{figure}[t]
    \centering
    \addtolength{\tabcolsep}{-5.5pt}
    \begin{tabular}{cccc}
        Ref Img & DWPose & AA & {Ours} \\
        \includegraphics[width=0.22\linewidth]{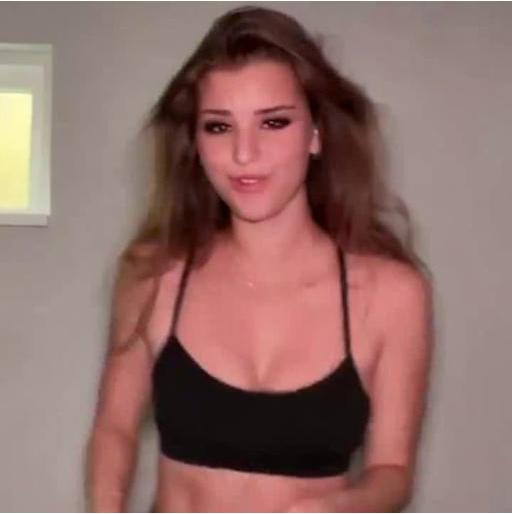} &
        \includegraphics[width=0.22\linewidth]{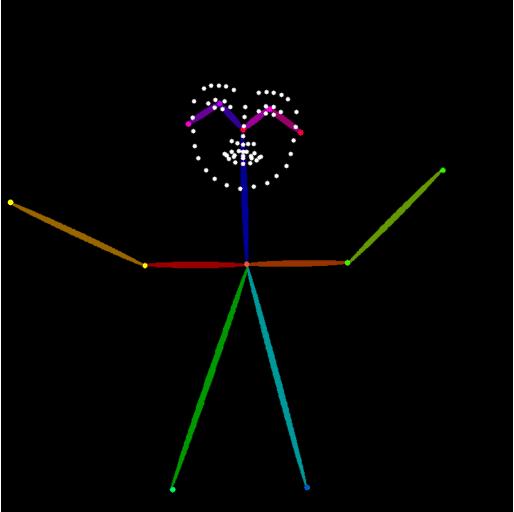} & \includegraphics[width=0.22\linewidth]{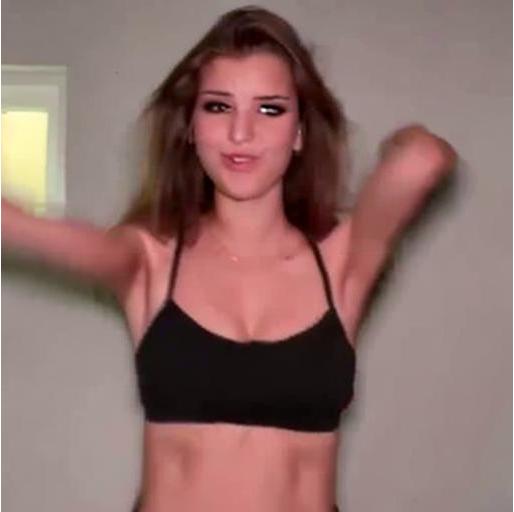} & \includegraphics[width=0.22\linewidth]{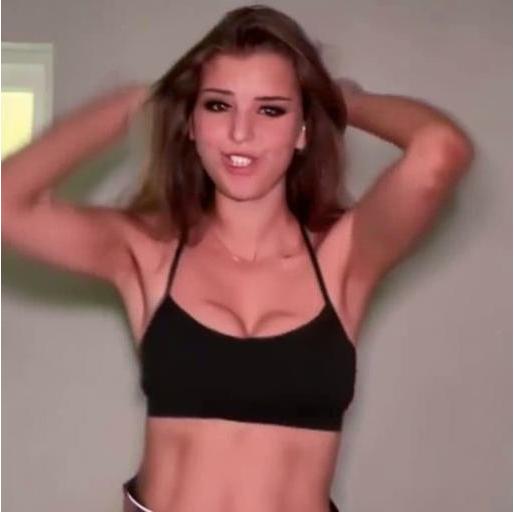} \\
      
        \includegraphics[width=0.22\linewidth]{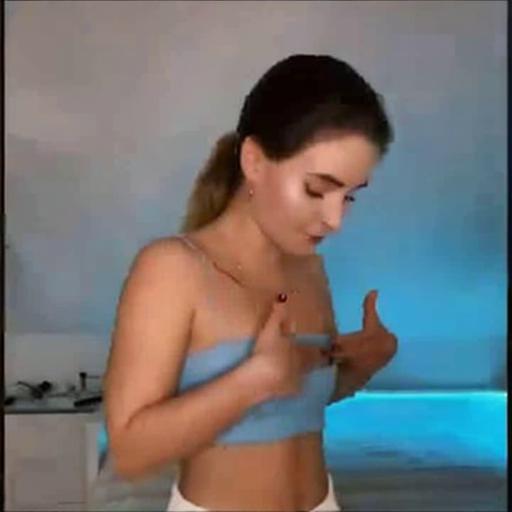} &
        \includegraphics[width=0.22\linewidth]{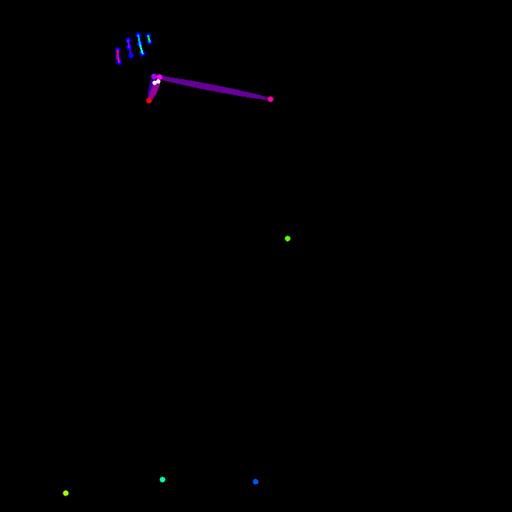} & \includegraphics[width=0.22\linewidth]{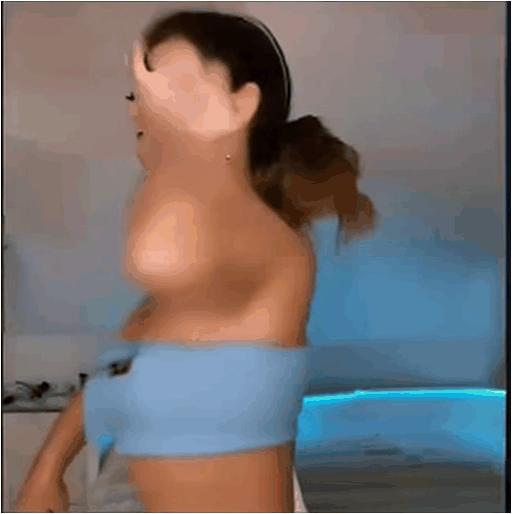} & \includegraphics[width=0.22\linewidth]{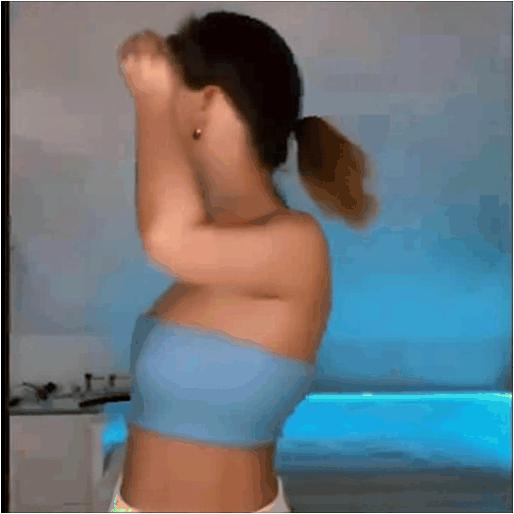} \\ 
      
        \includegraphics[width=0.22\linewidth]{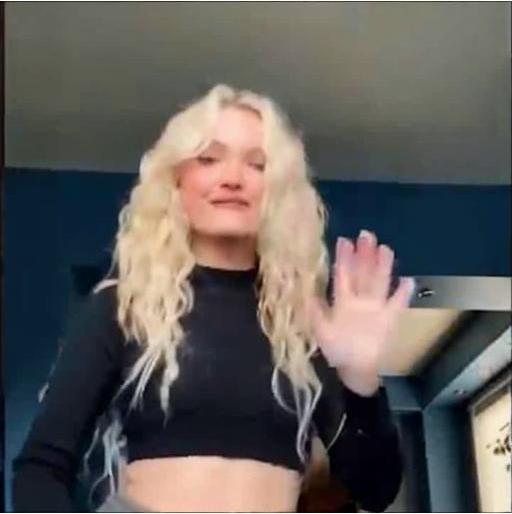} &
        \includegraphics[width=0.22\linewidth]{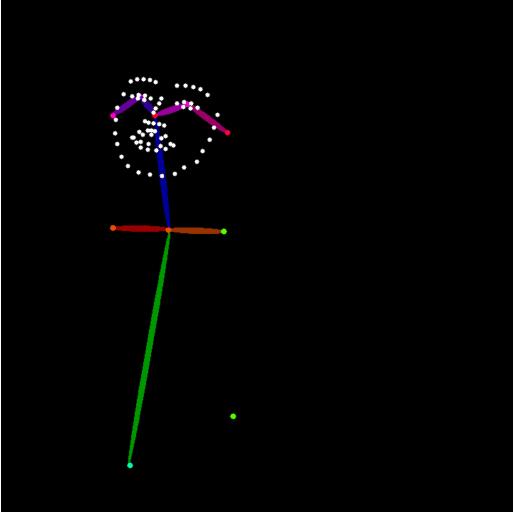} & \includegraphics[width=0.22\linewidth]{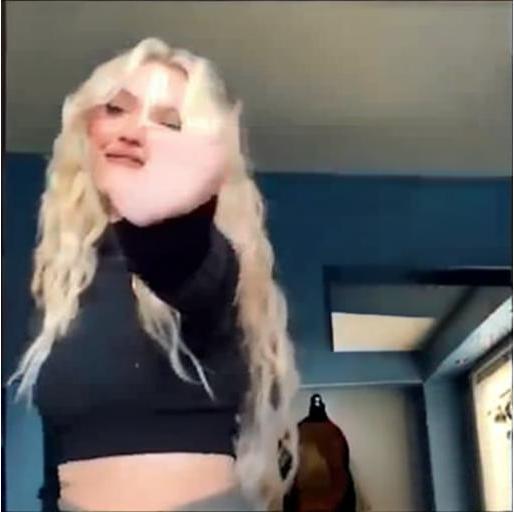} & \includegraphics[width=0.22\linewidth]{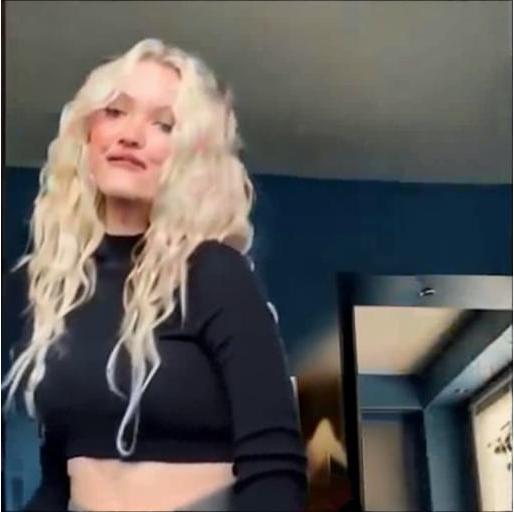} \\ 
    \end{tabular}
    \caption{Comparisons of generation stability when DWPose is corrupted.}
    \label{fig:robust}
\end{figure}

\vspace{1mm}
\noindent\textbf{Generation stability.}
To further evaluate generation stability, Figure~\ref{fig:robust} shows several samples when DWPose is severely corrupted. In this situation, Animate Anybody leveraging only DWPose will generate artifacts (broken arm in case 1) or blurry frames (cases 2 and 3). However, RealisDance can achieve much better results based on information from the other two poses.

\begin{figure*}[t]
    \centering
    \addtolength{\tabcolsep}{-5.5pt}
    \begin{tabular}{cccccccccc}
        & Frame 1 & Frame 2 & Frame 3 & Frame 4 & ~ & Frame 1 & Frame 2 & Frame3 & Frame 4\\
        \rotatebox{90}{~~~~~~~AA} &
        \includegraphics[width=0.12\linewidth]{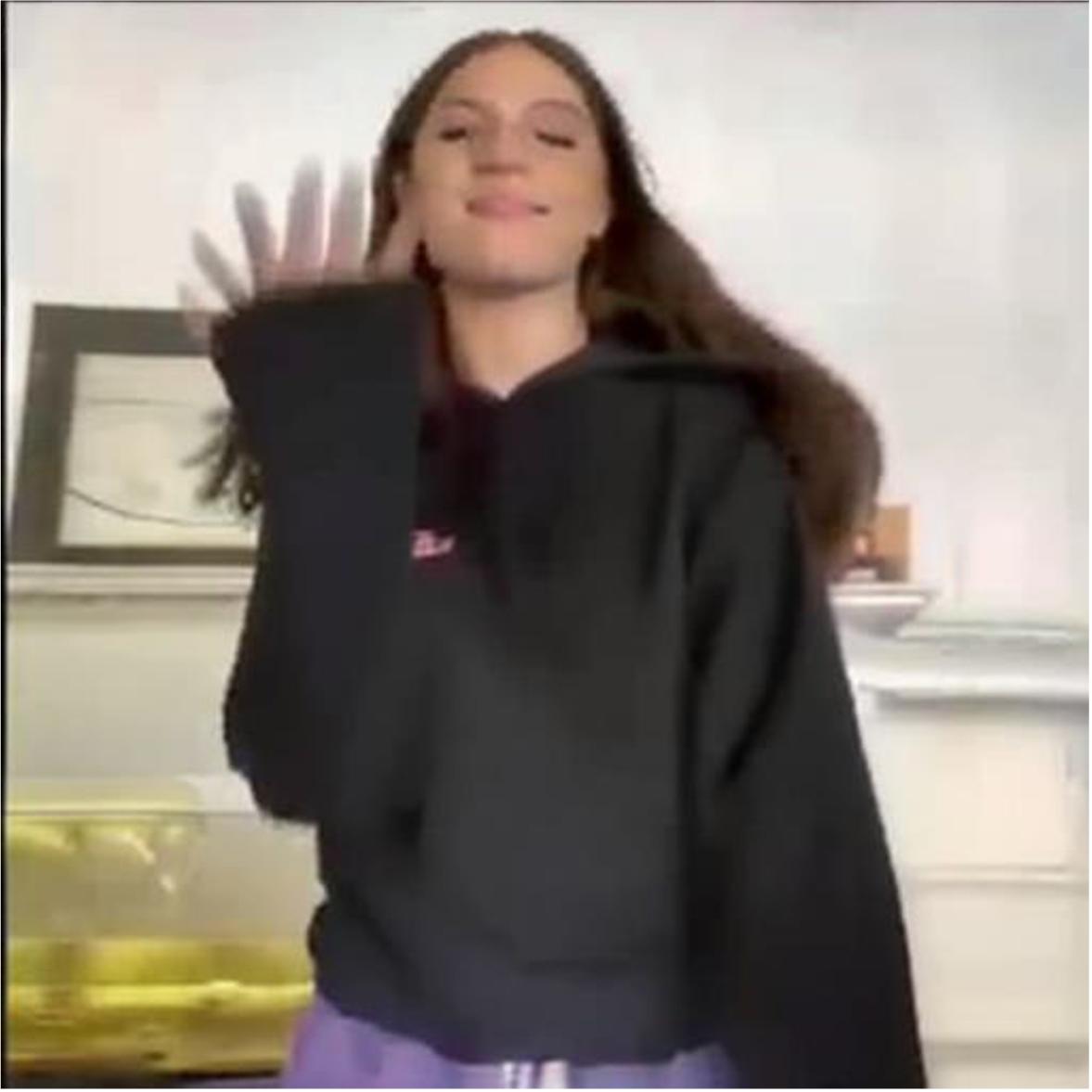} &
        \includegraphics[width=0.12\linewidth]{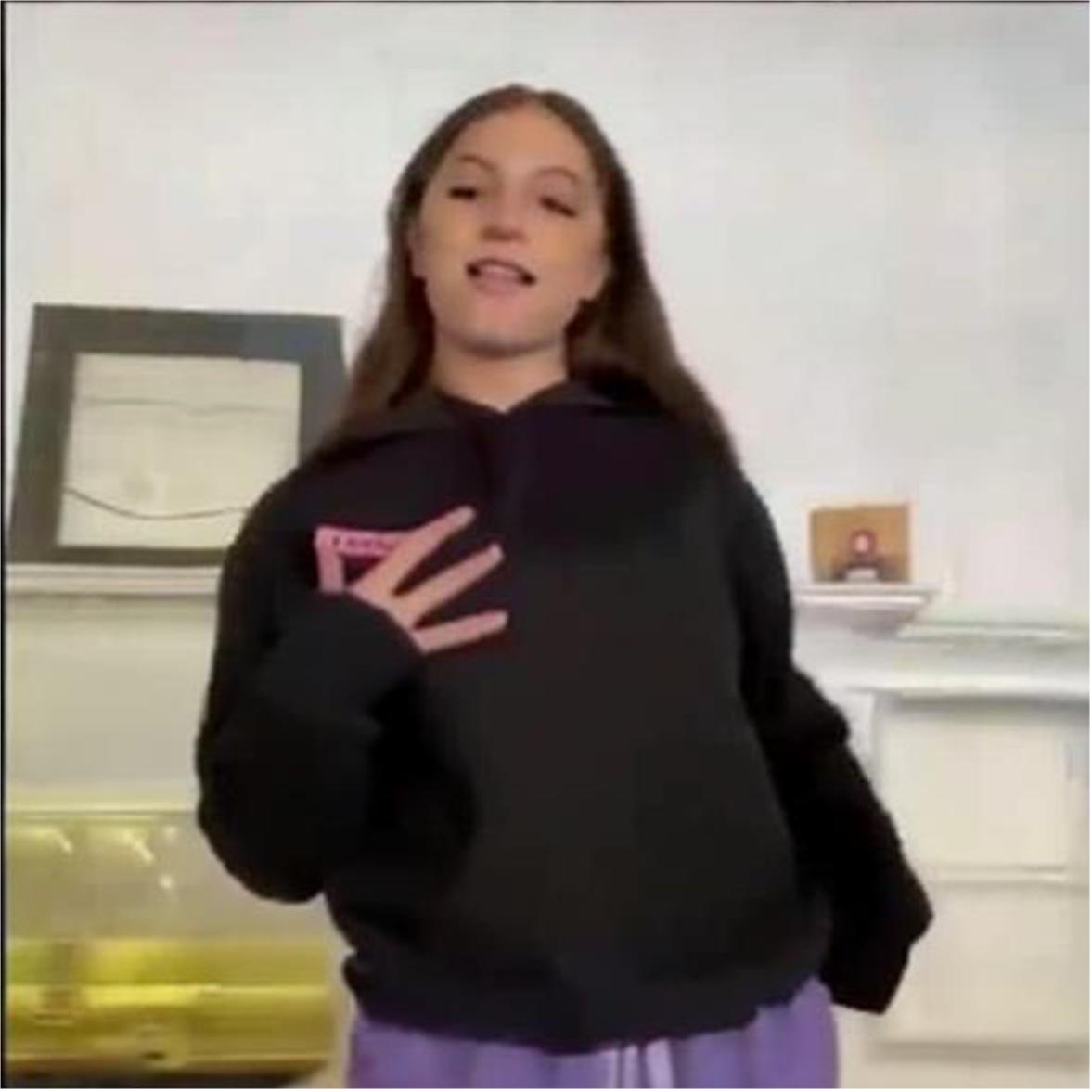} & 
        \includegraphics[width=0.12\linewidth]{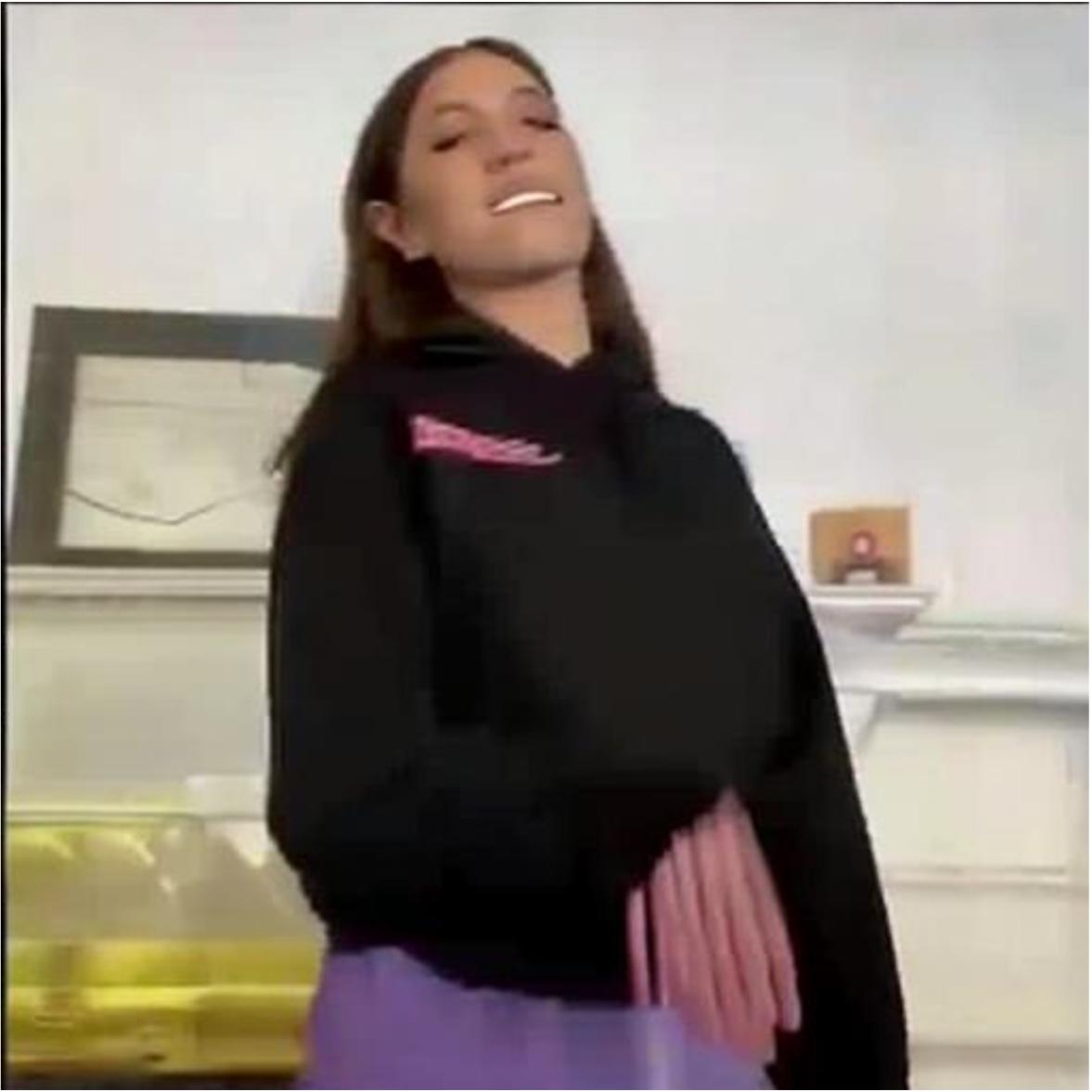} & 
        \includegraphics[width=0.12\linewidth]{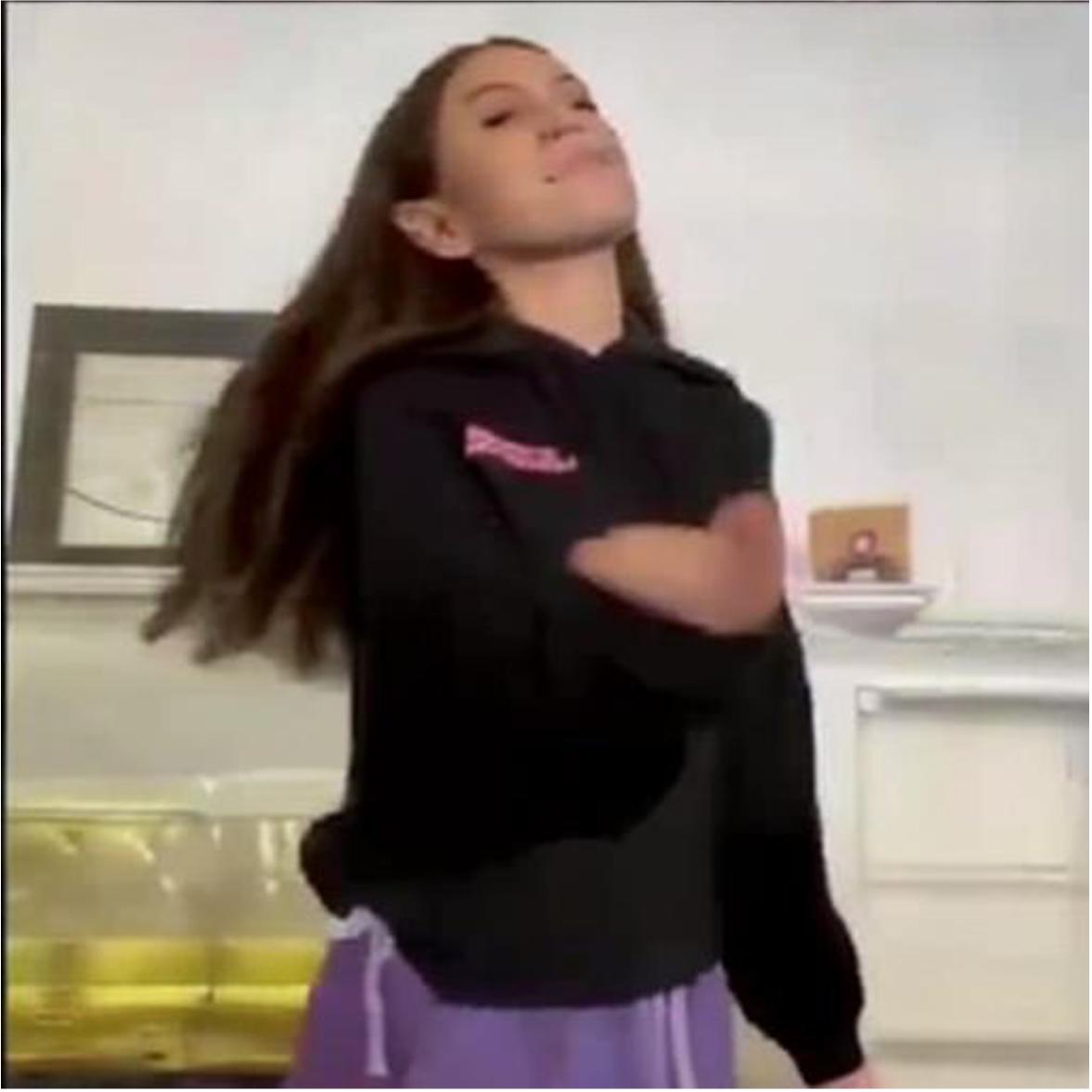} &
        ~&
        \includegraphics[width=0.12\linewidth]{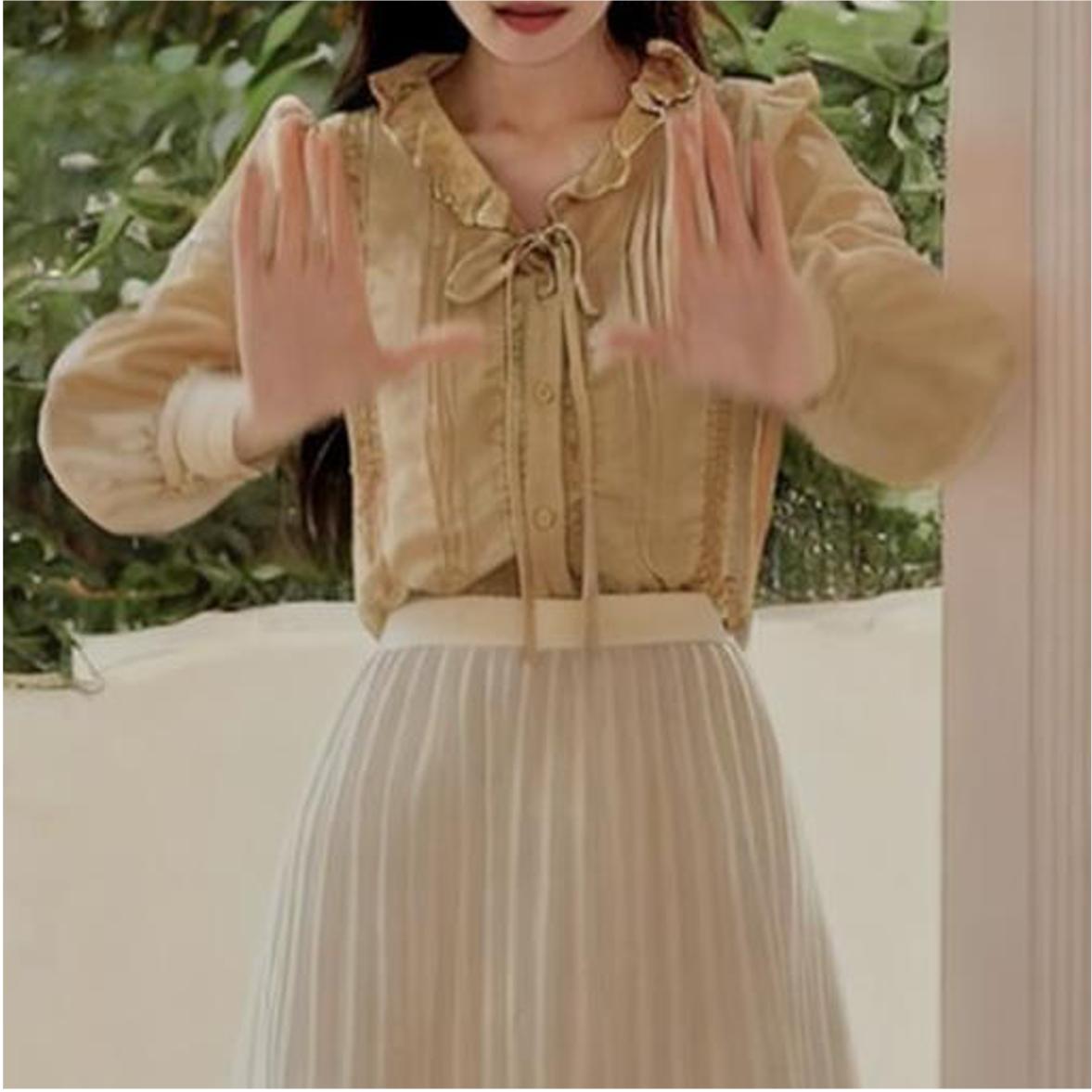} &
        \includegraphics[width=0.12\linewidth]{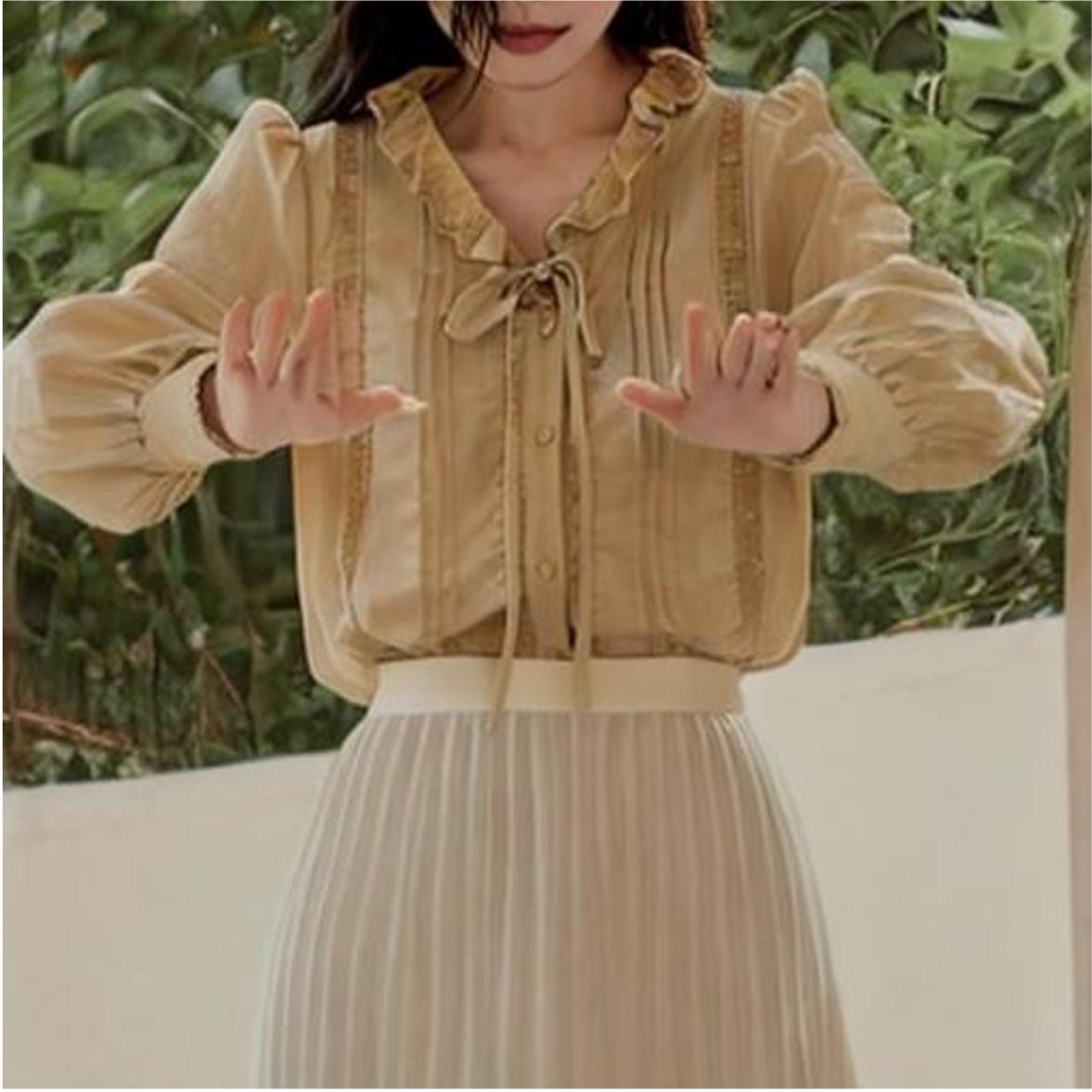} & 
        \includegraphics[width=0.12\linewidth]{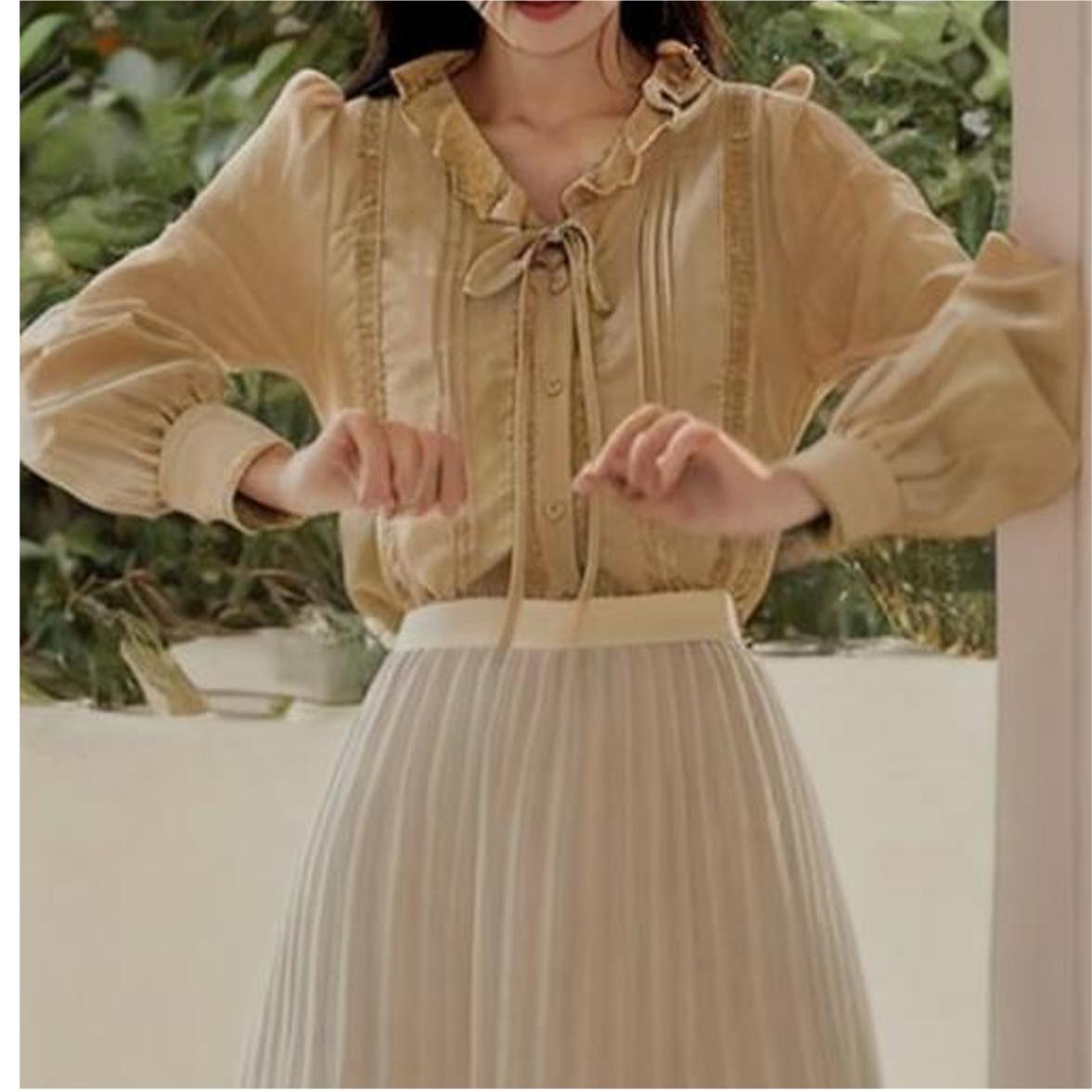} & 
        \includegraphics[width=0.12\linewidth]{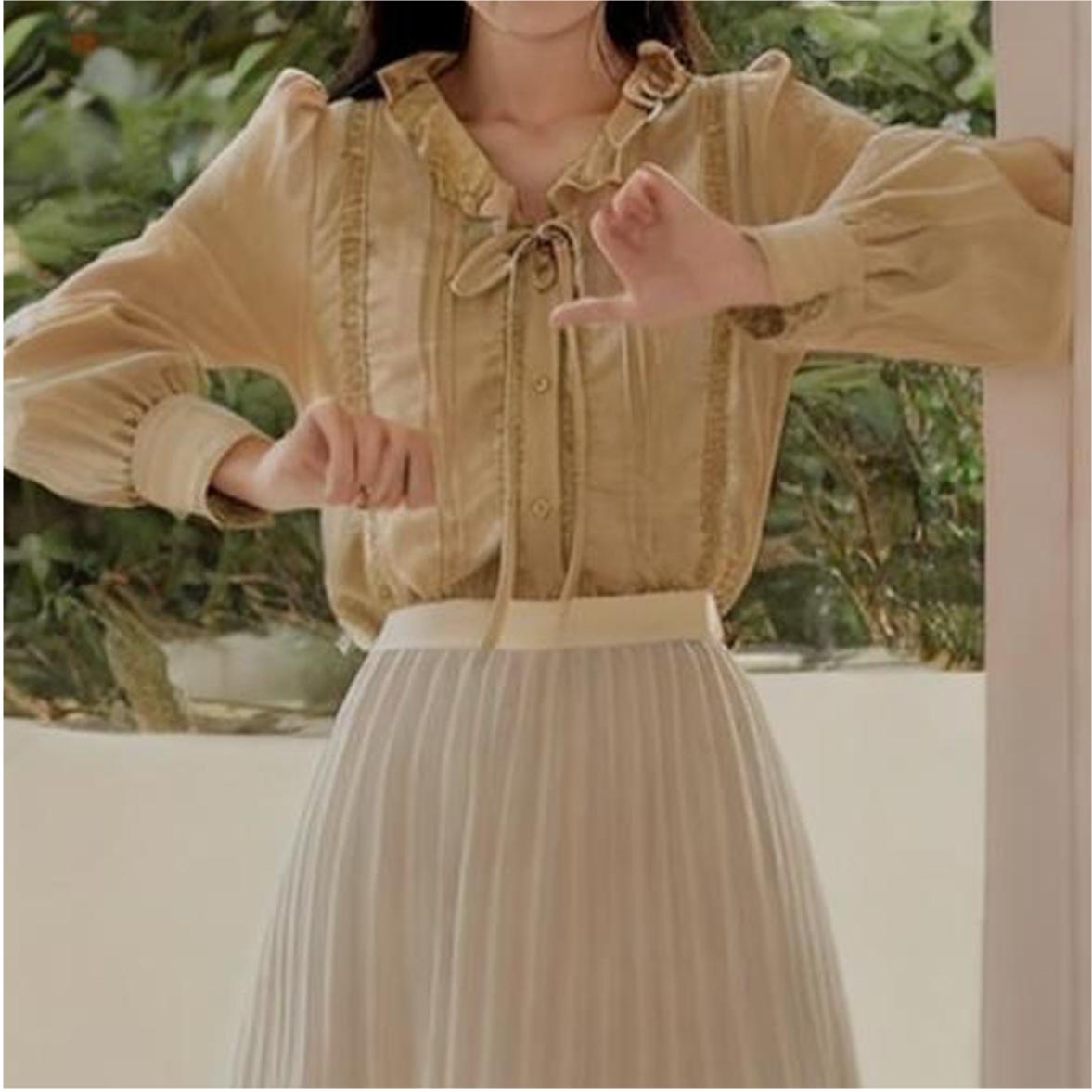} \\
        \rotatebox{90}{~~~~~~{Ours}} &
        \includegraphics[width=0.12\linewidth]{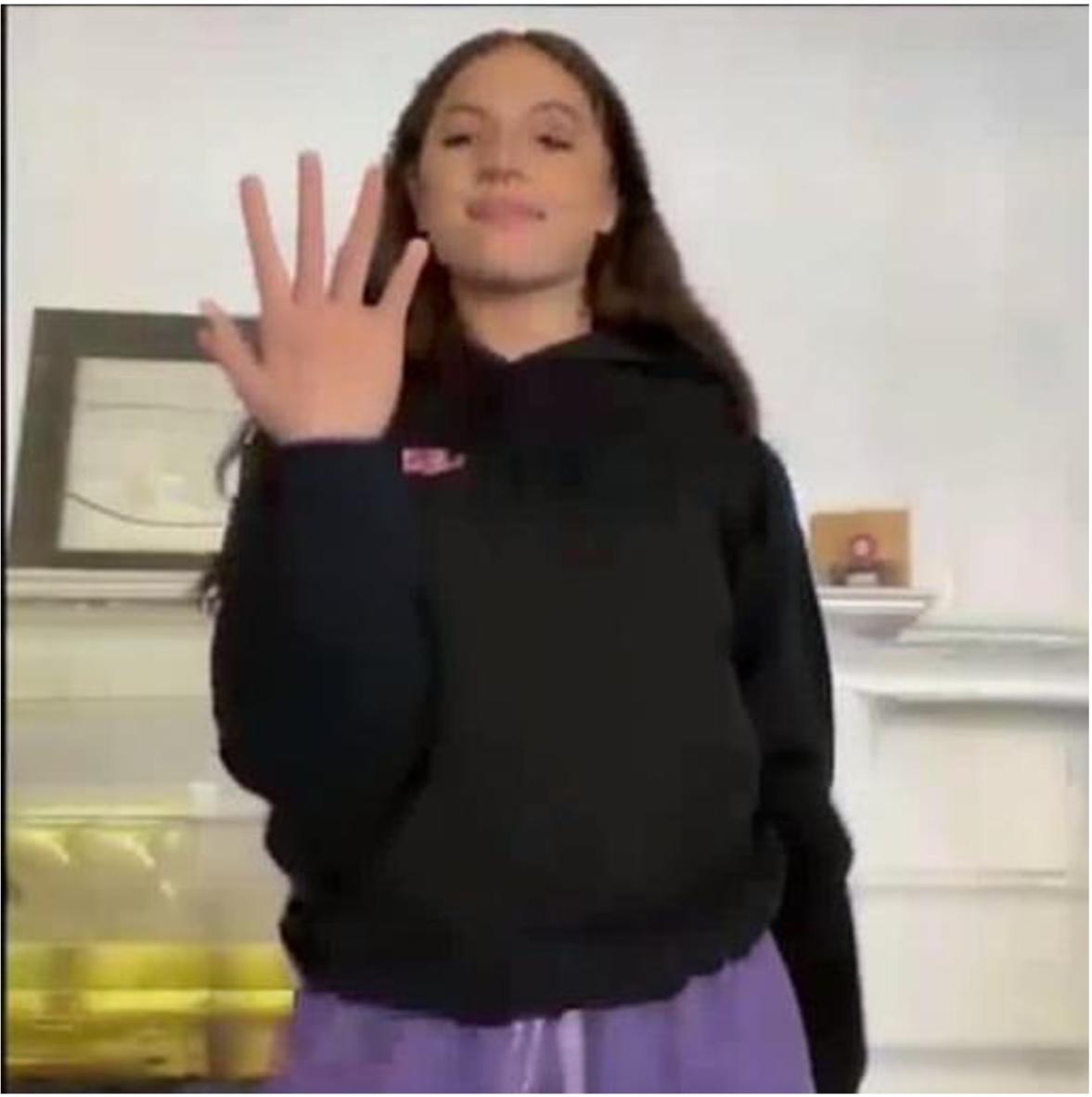} &
        \includegraphics[width=0.12\linewidth]{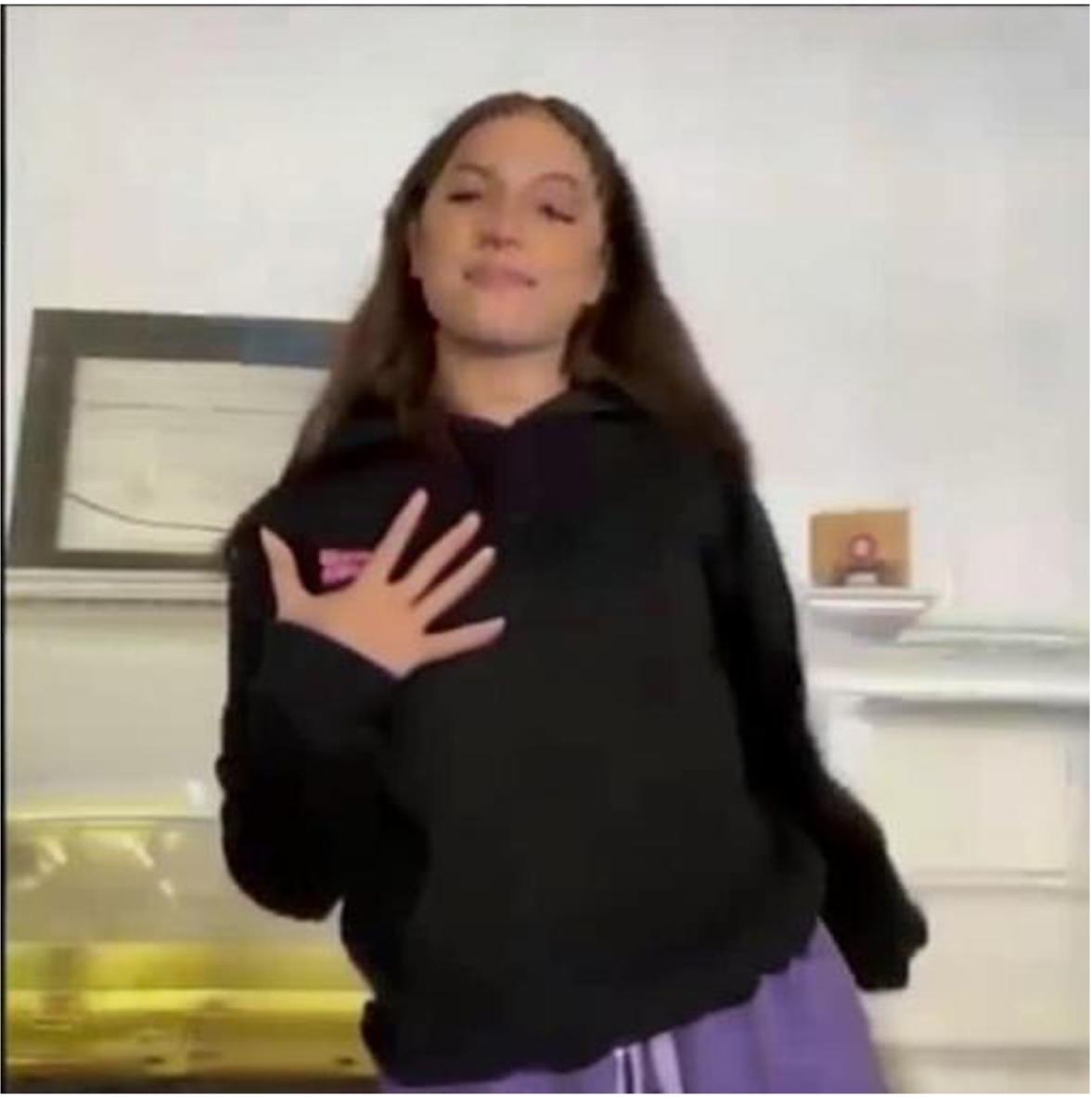} & 
        \includegraphics[width=0.12\linewidth]{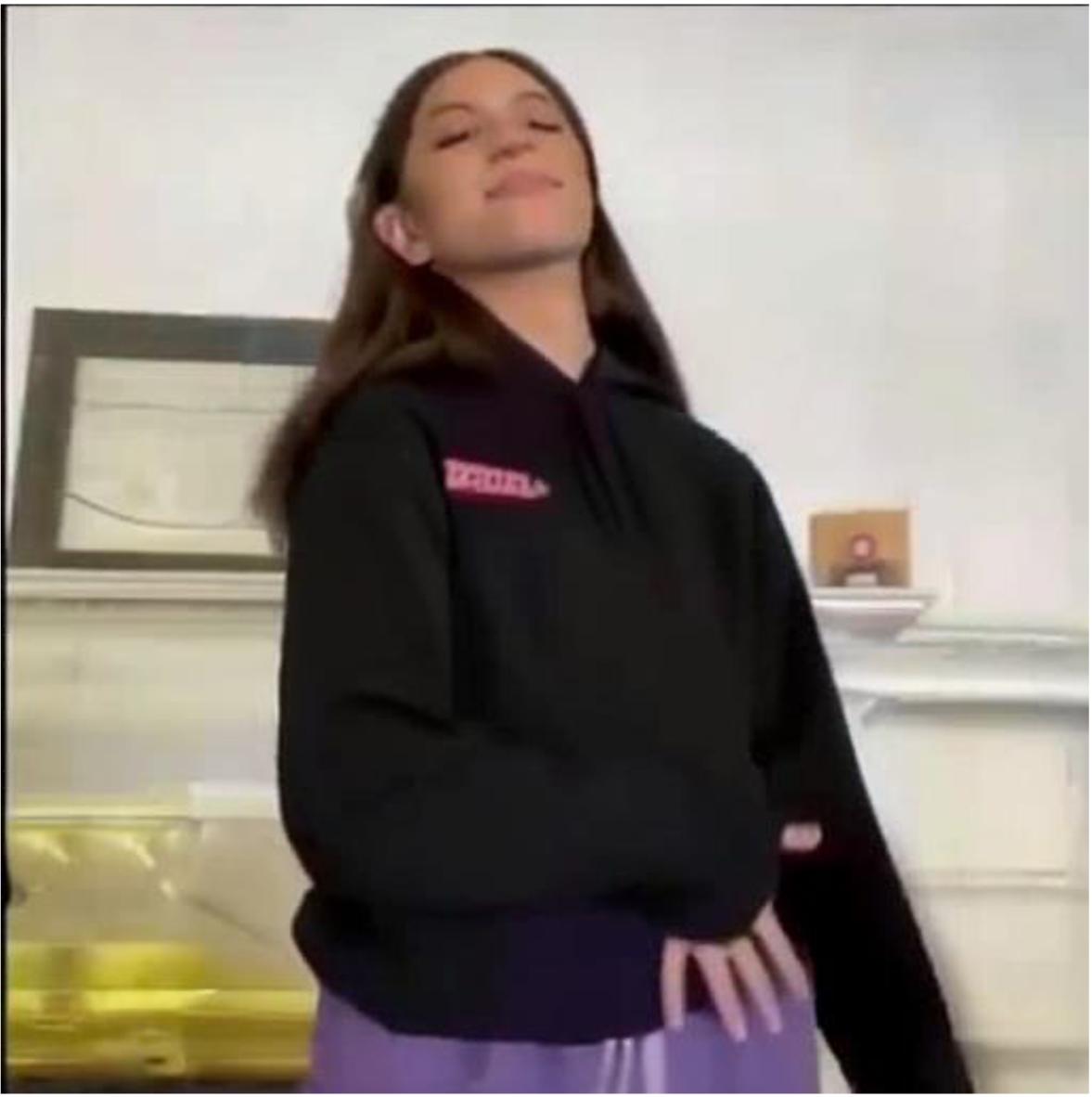} & 
        \includegraphics[width=0.12\linewidth]{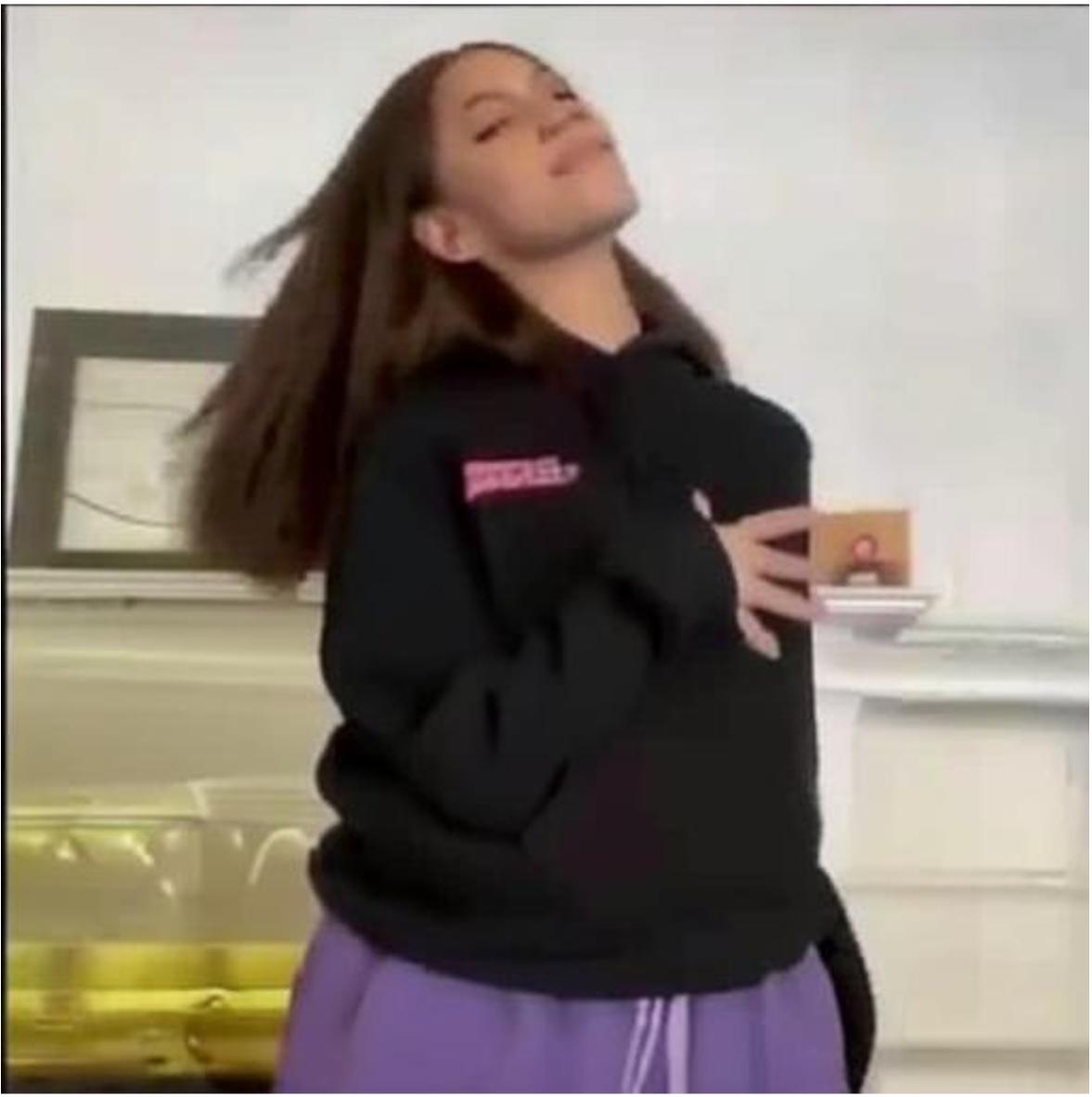} &
        ~&
        \includegraphics[width=0.12\linewidth]{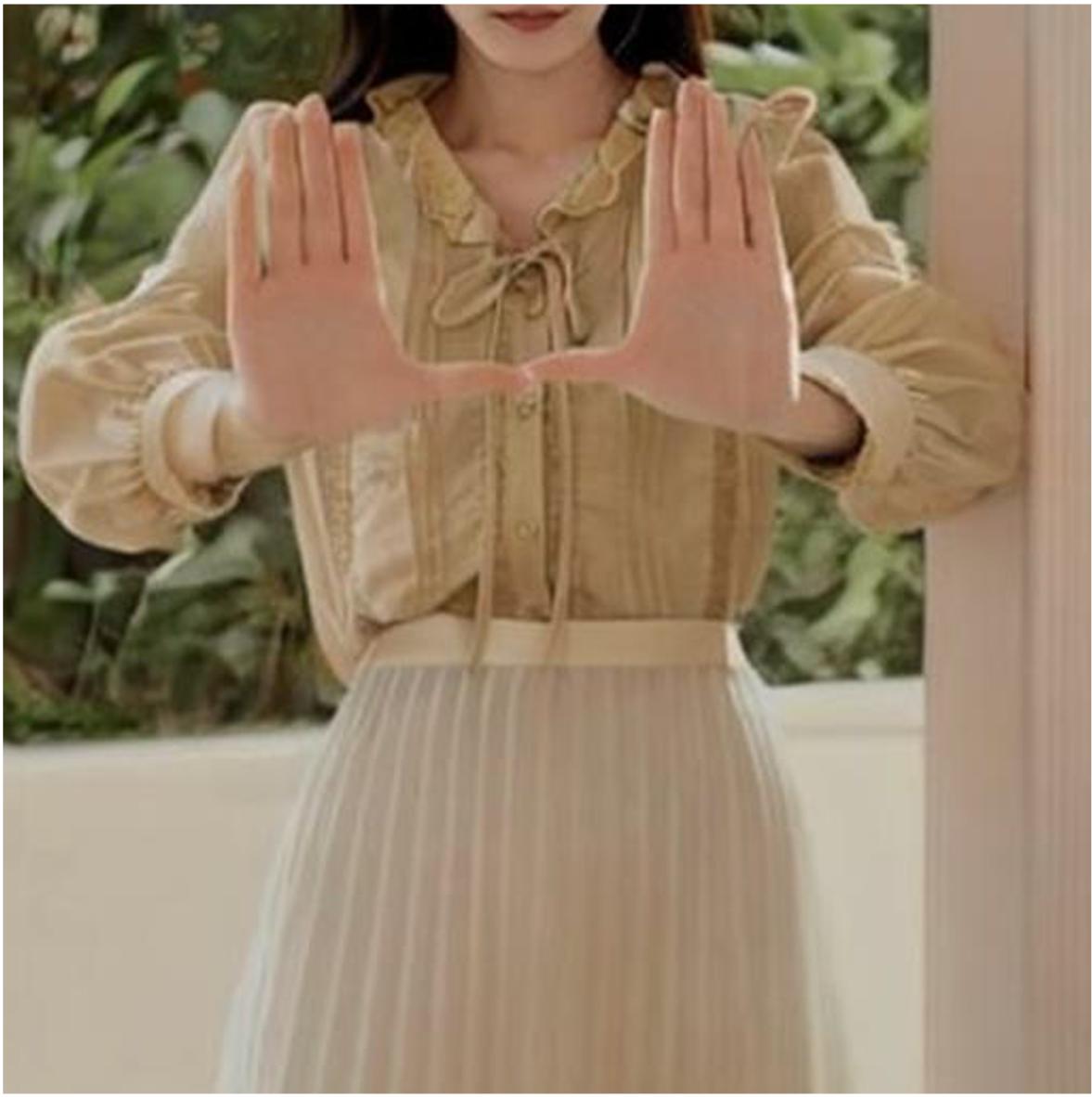} &
        \includegraphics[width=0.12\linewidth]{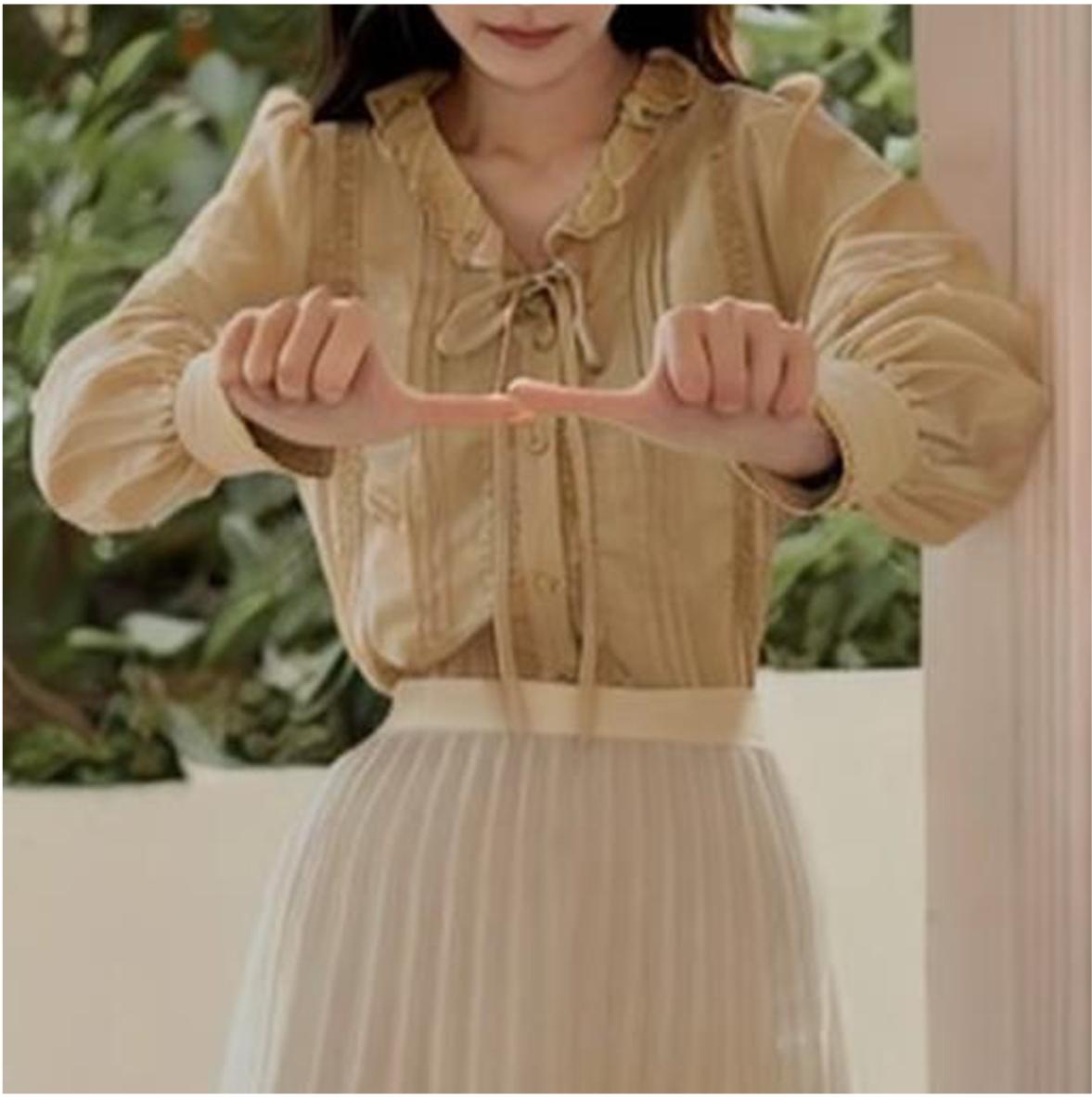} & 
        \includegraphics[width=0.12\linewidth]{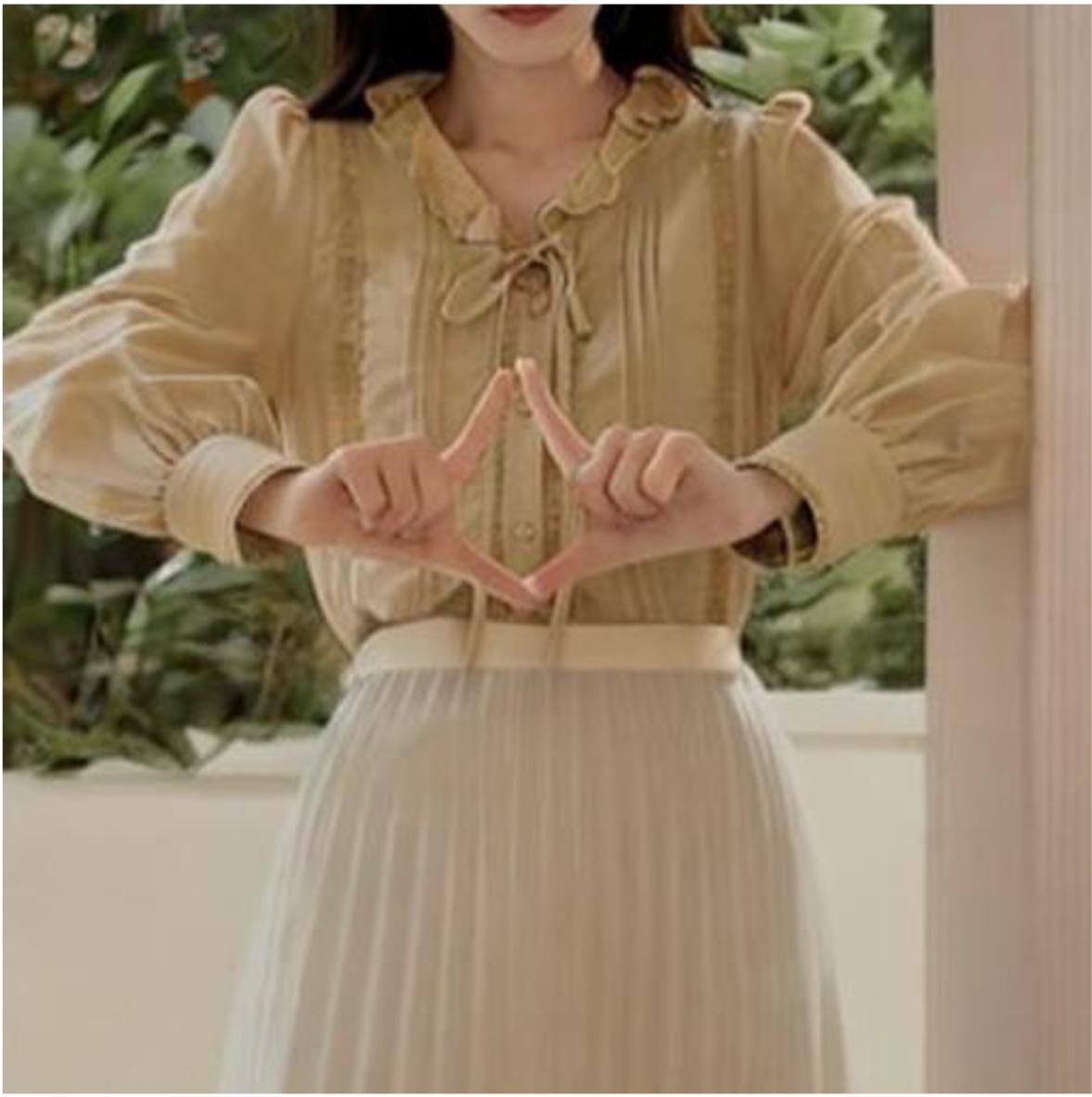} & 
        \includegraphics[width=0.12\linewidth]{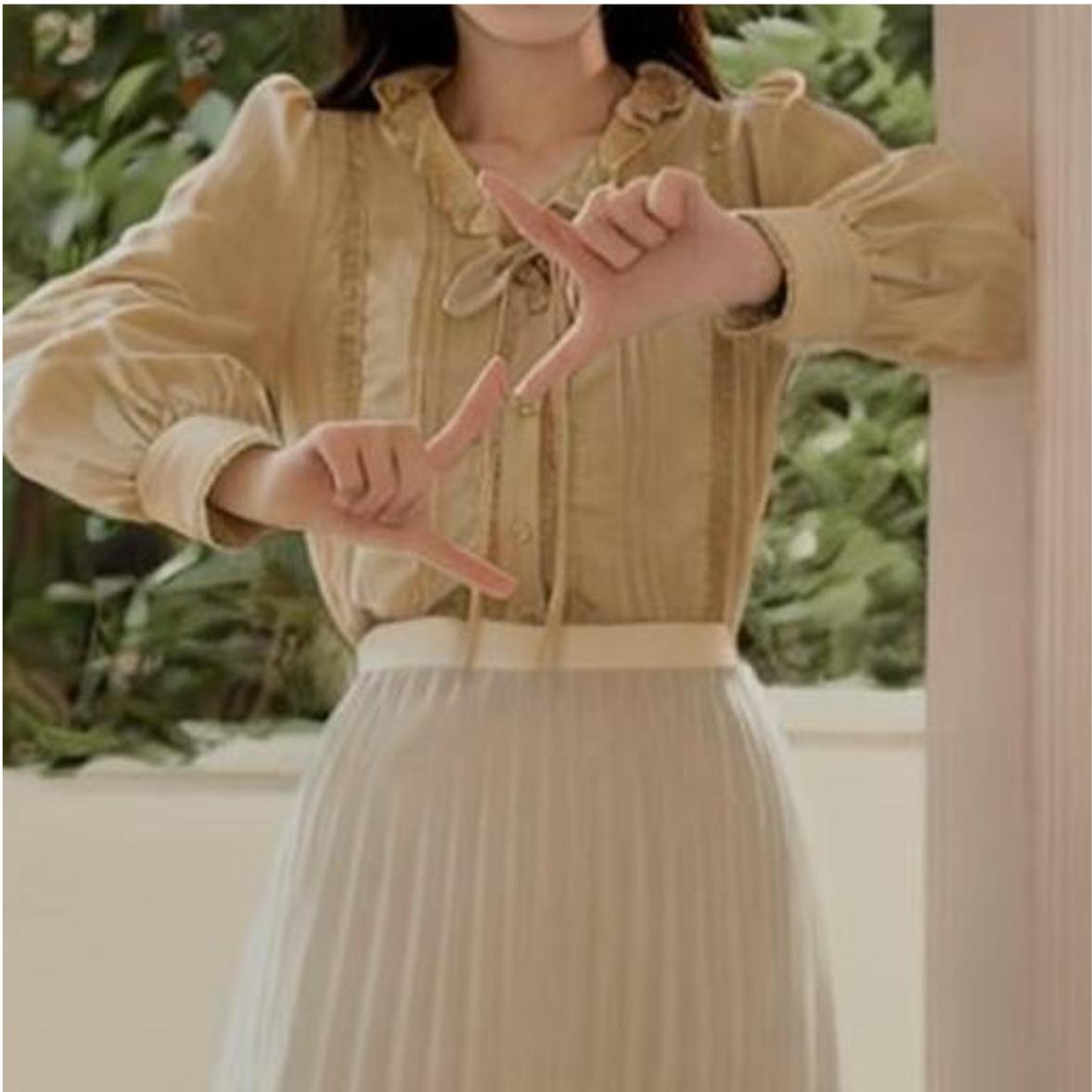} \\
        \rotatebox{90}{~~~~~~~AA} &
        \includegraphics[width=0.12\linewidth]{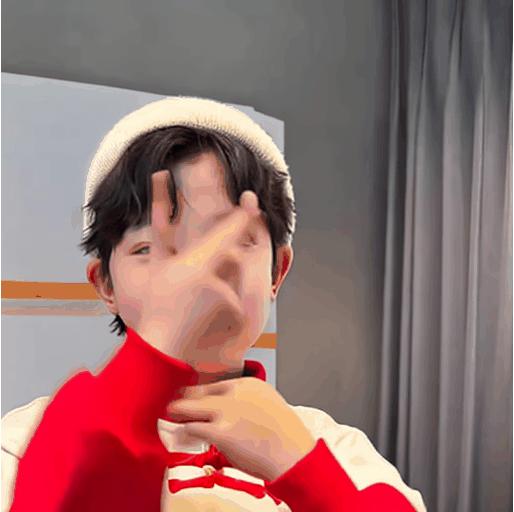} &
        \includegraphics[width=0.12\linewidth]{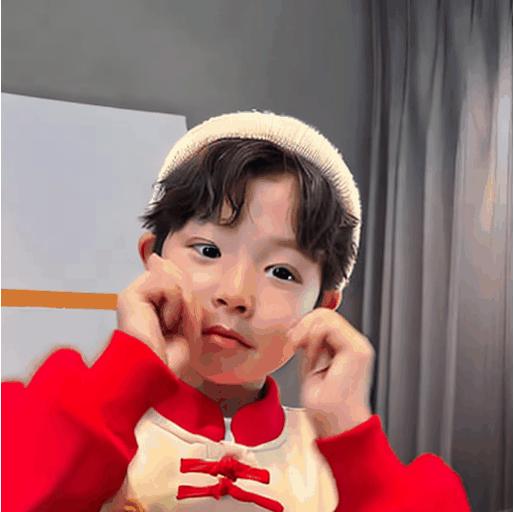} & 
        \includegraphics[width=0.12\linewidth]{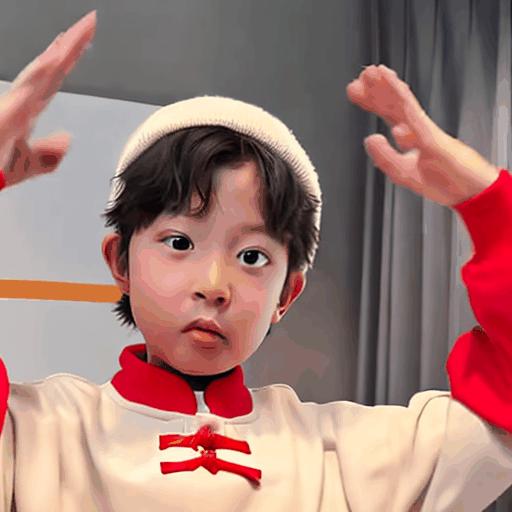} & 
        \includegraphics[width=0.12\linewidth]{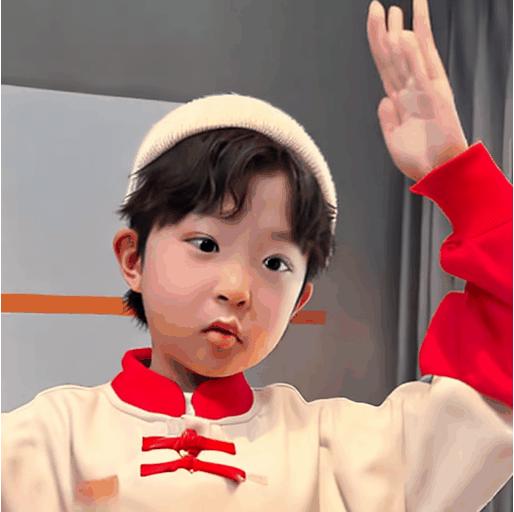} &
        ~&
        \includegraphics[width=0.12\linewidth]{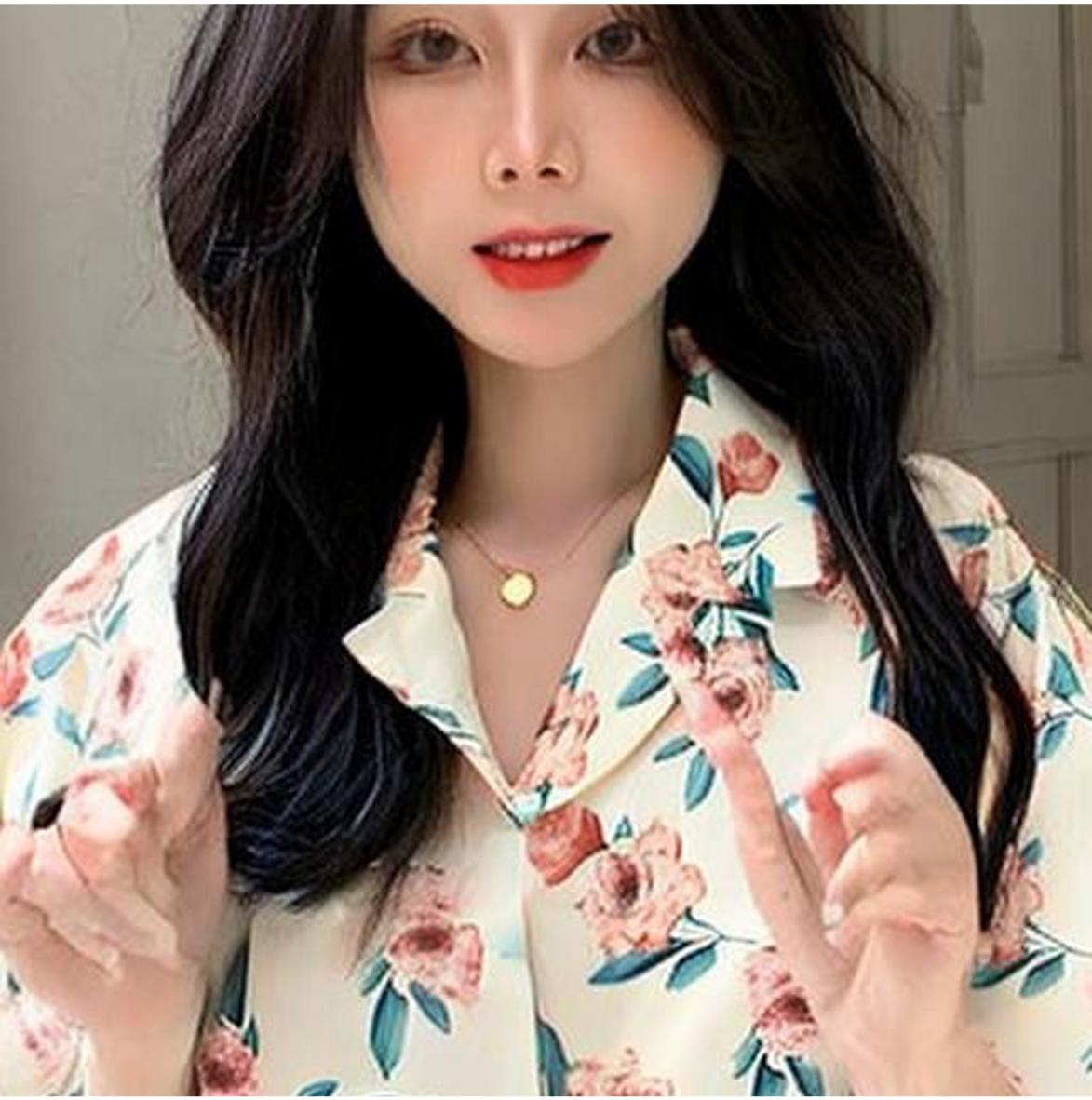} &
        \includegraphics[width=0.12\linewidth]{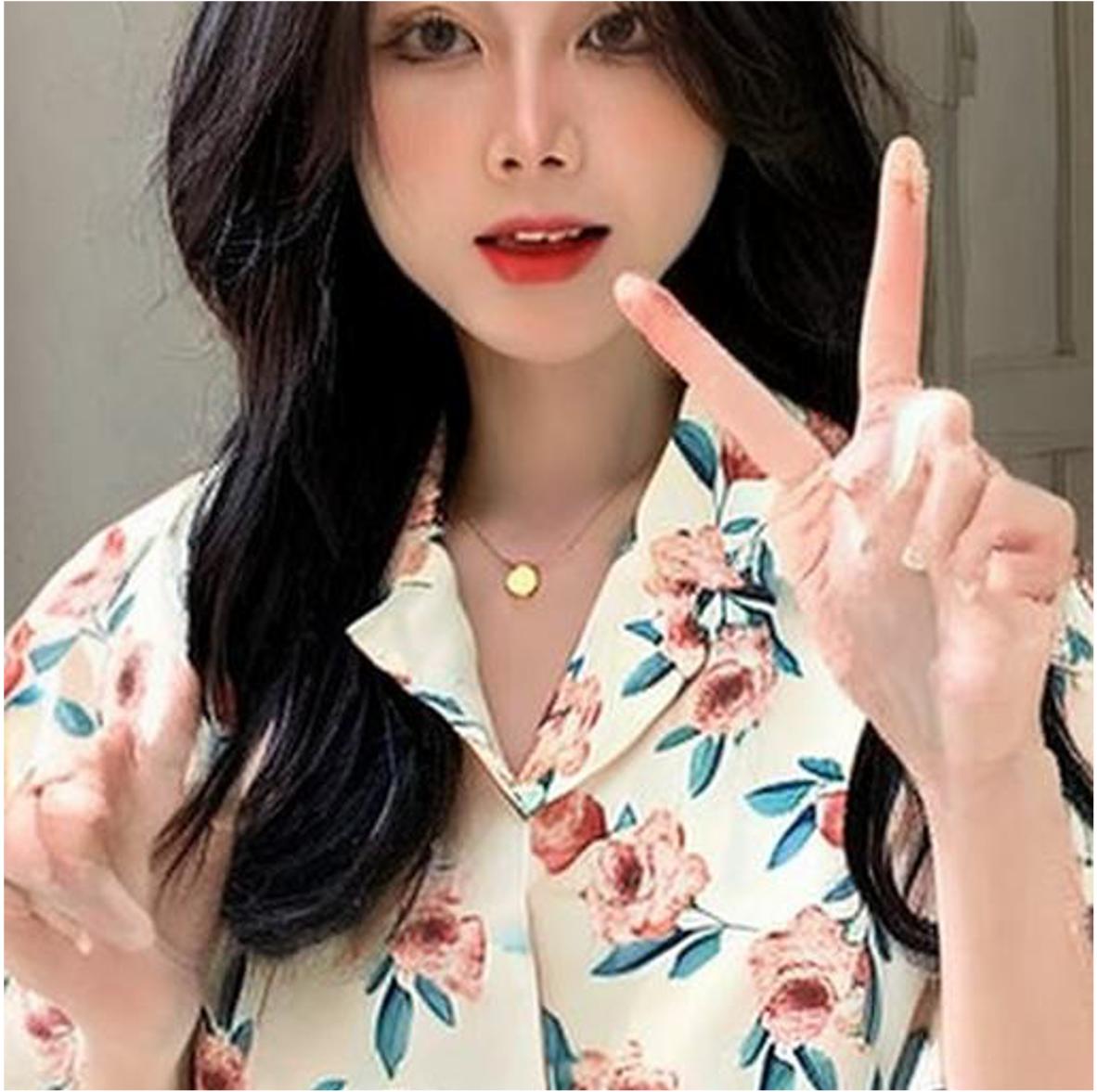} & 
        \includegraphics[width=0.12\linewidth]{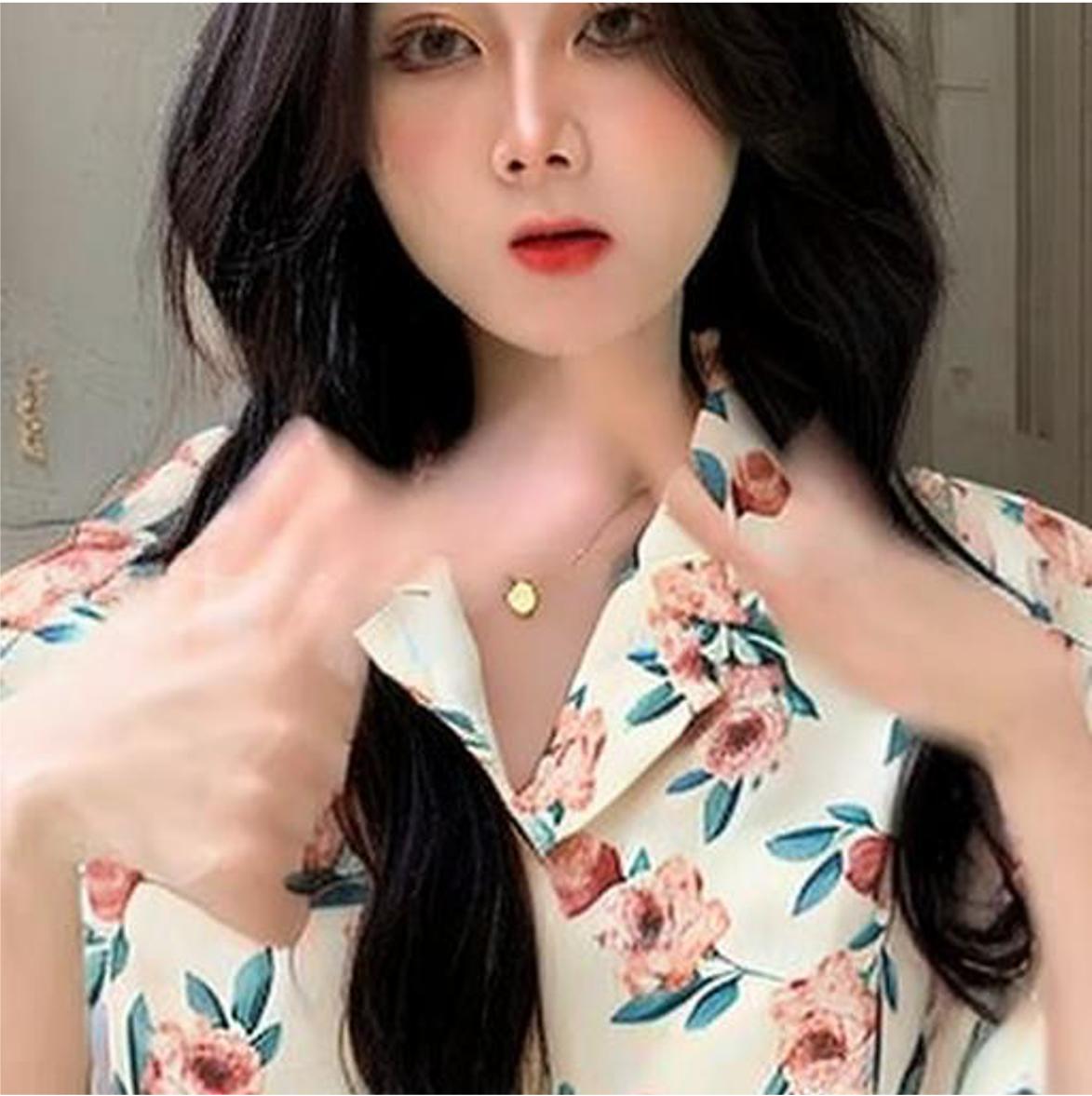} & 
        \includegraphics[width=0.12\linewidth]{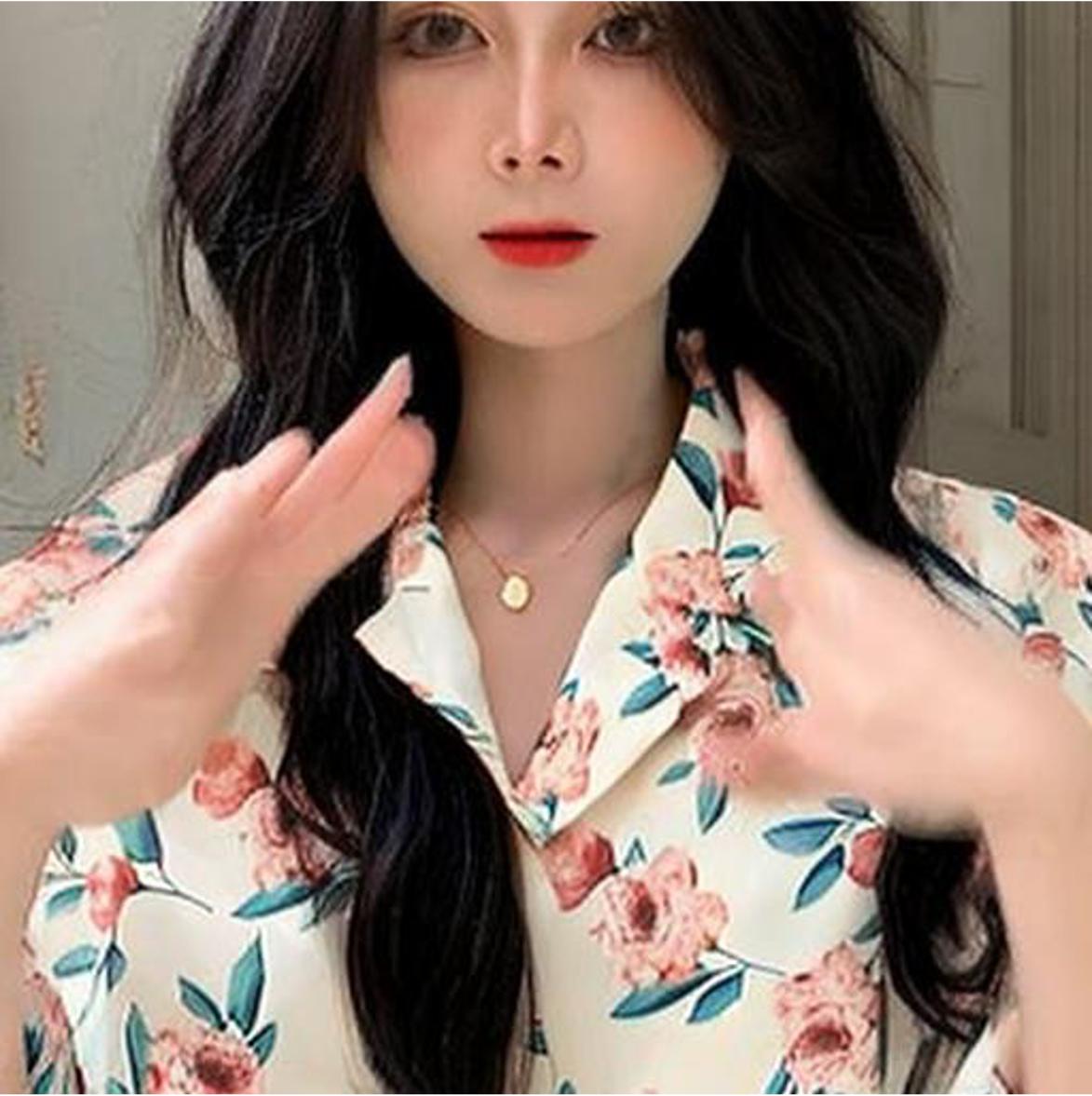} \\
        \rotatebox{90}{~~~~~~{Ours}} &
        \includegraphics[width=0.12\linewidth]{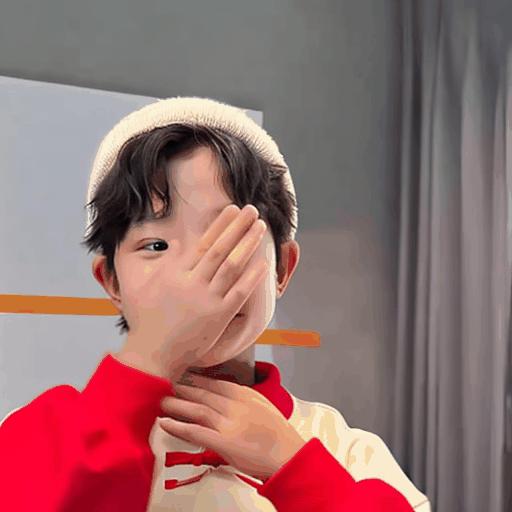} &
        \includegraphics[width=0.12\linewidth]{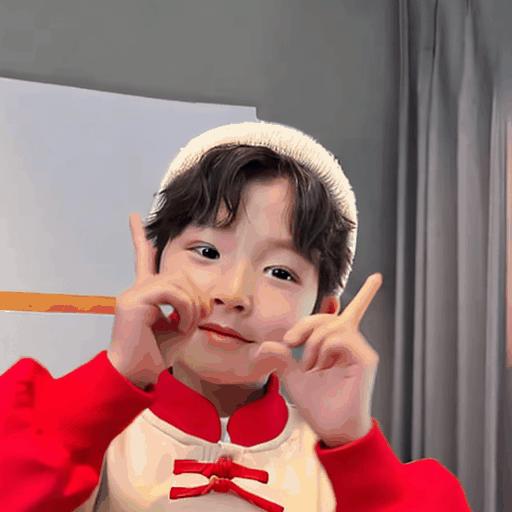} & 
        \includegraphics[width=0.12\linewidth]{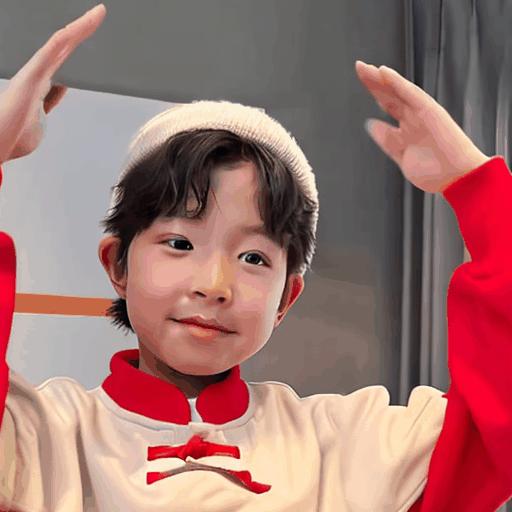} & 
        \includegraphics[width=0.12\linewidth]{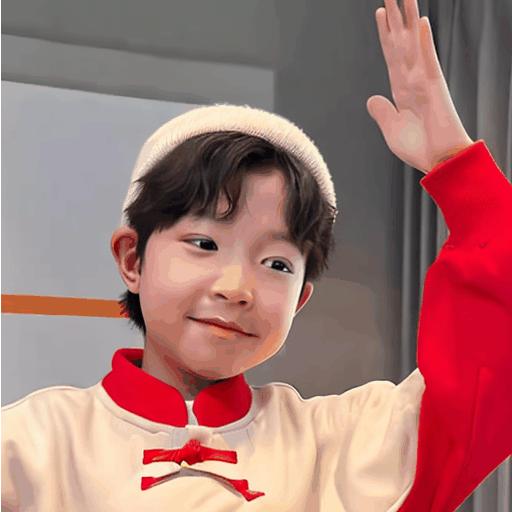} &
        ~&
        \includegraphics[width=0.12\linewidth]{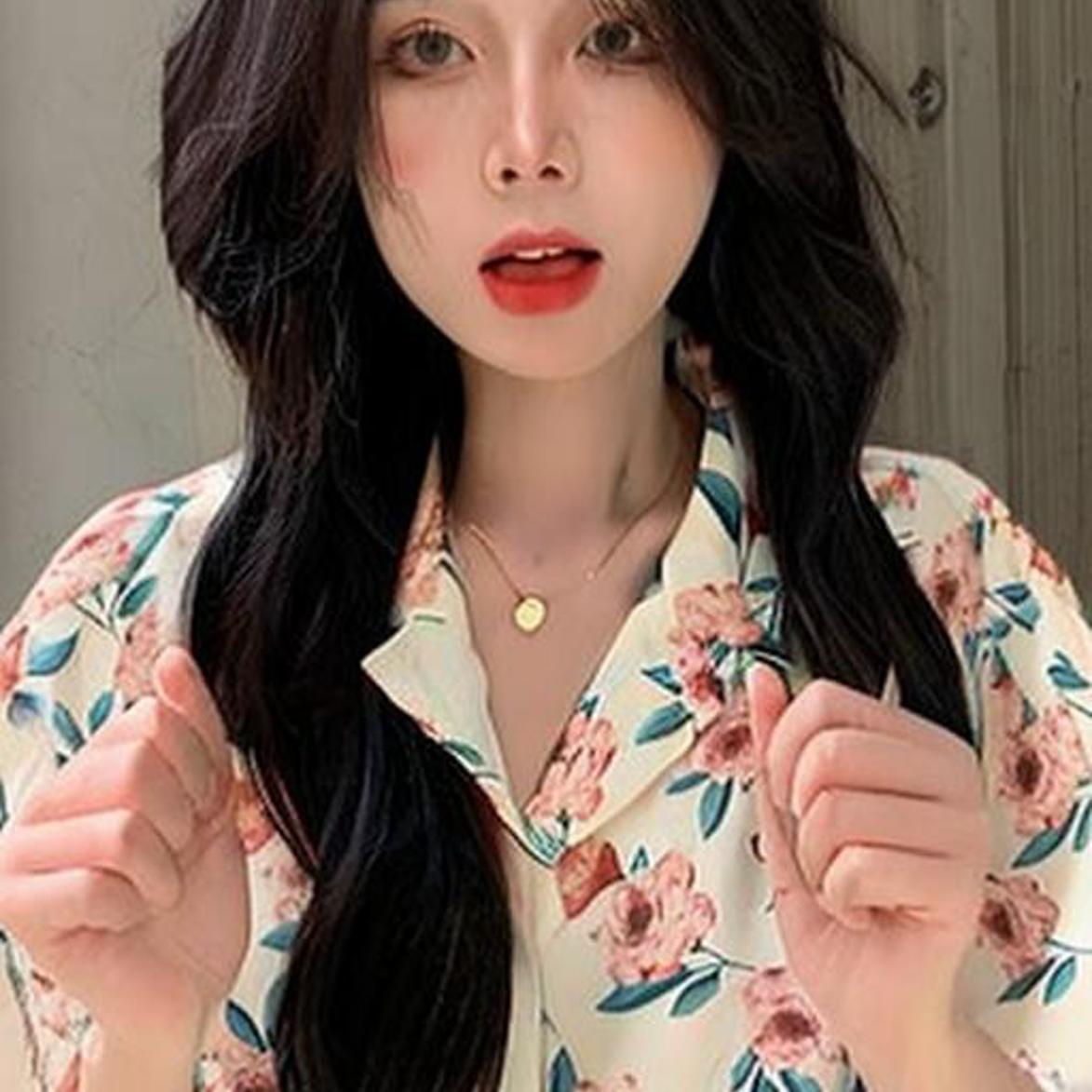} &
        \includegraphics[width=0.12\linewidth]{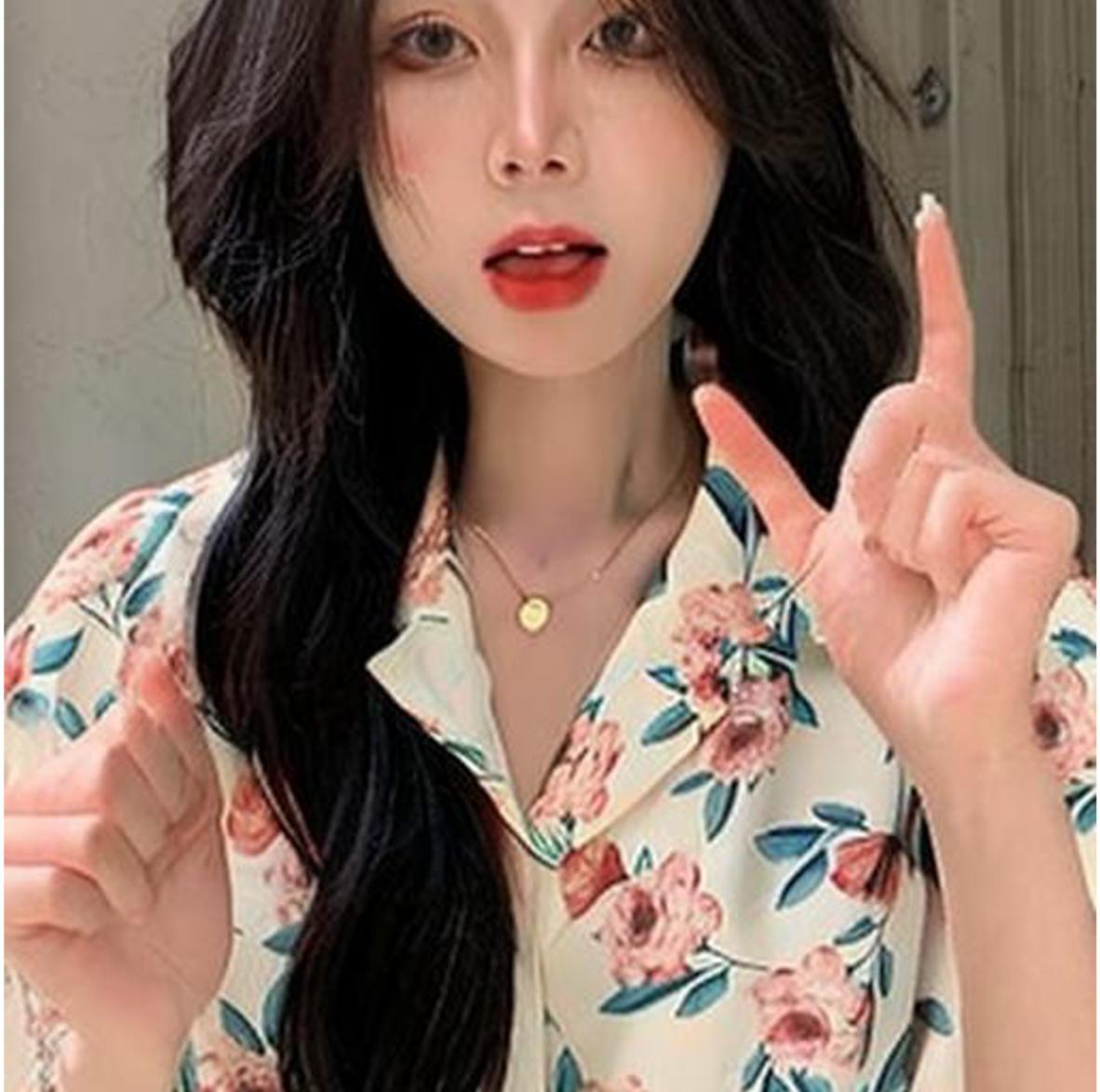} & 
        \includegraphics[width=0.12\linewidth]{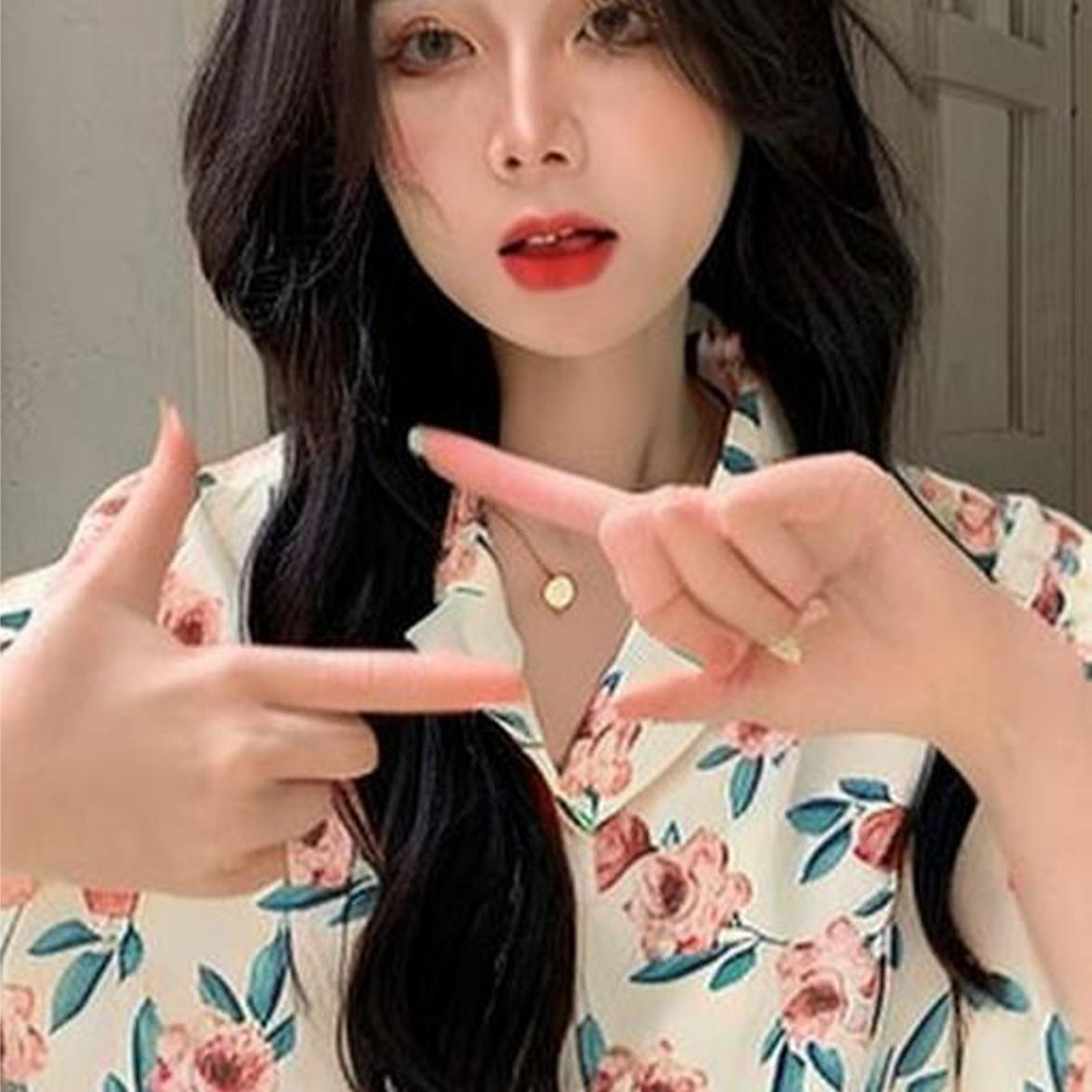} & 
        \includegraphics[width=0.12\linewidth]{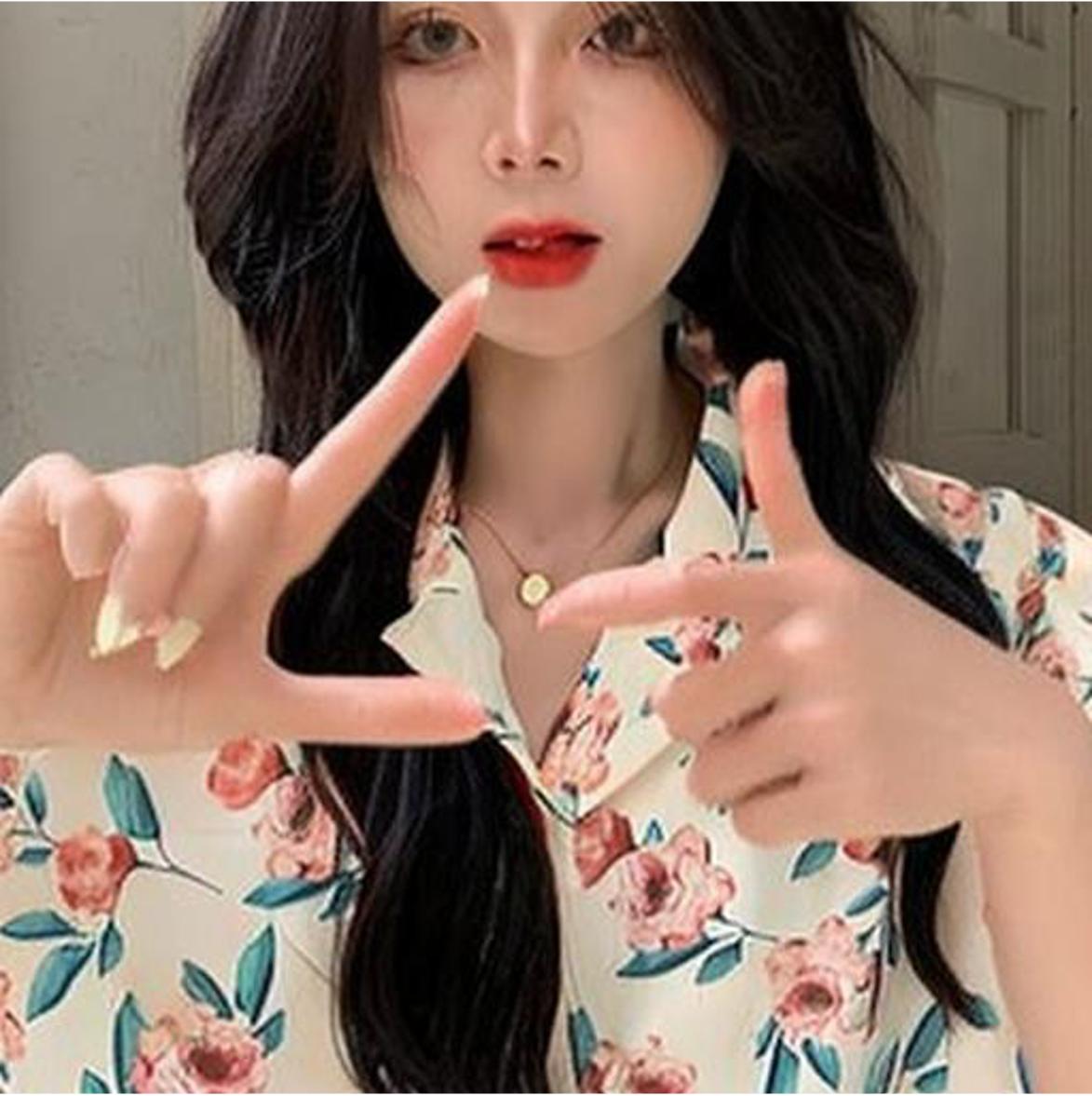} \\
        \rotatebox{90}{~~~~~~~AA} &
        \includegraphics[width=0.12\linewidth]{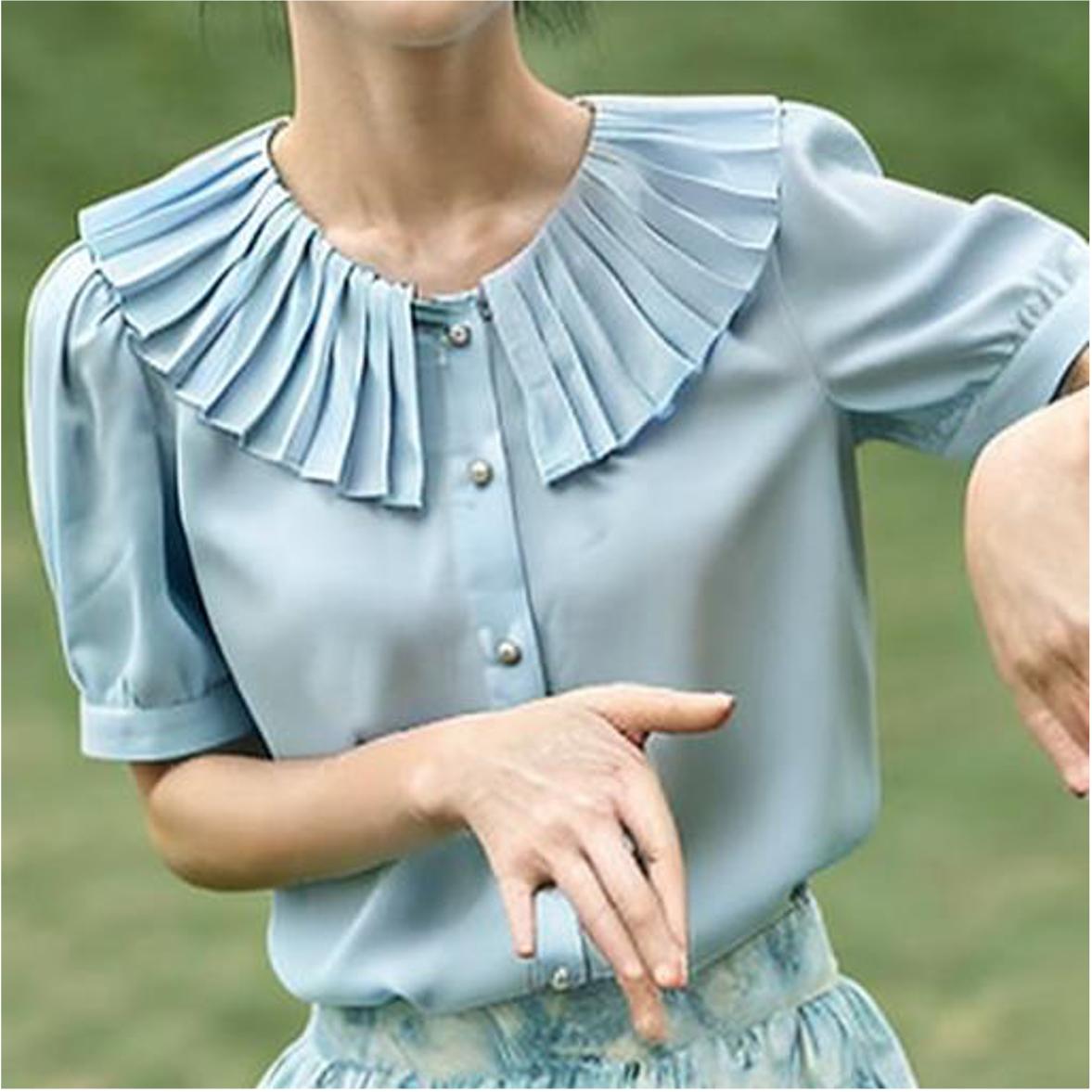} &
        \includegraphics[width=0.12\linewidth]{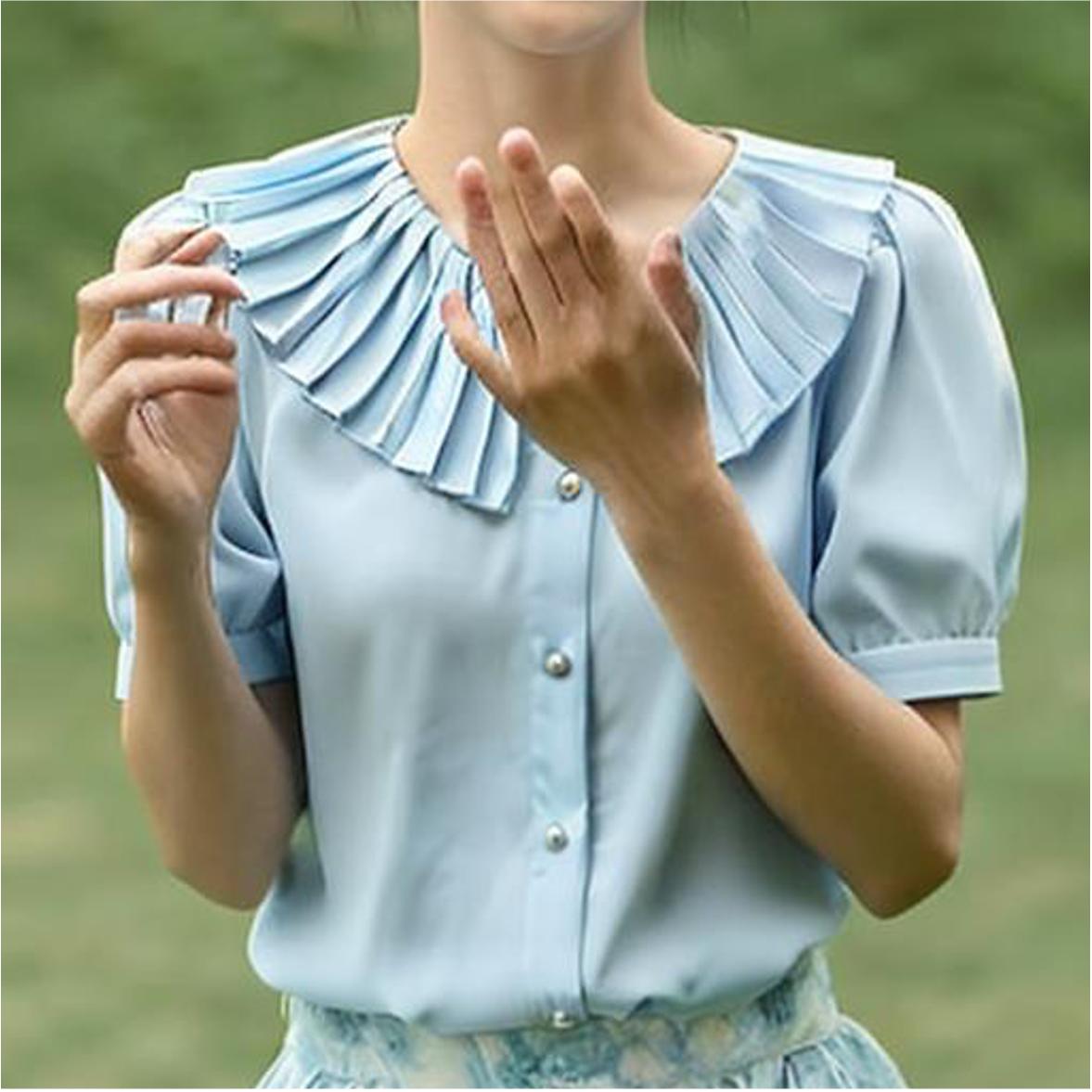} & 
        \includegraphics[width=0.12\linewidth]{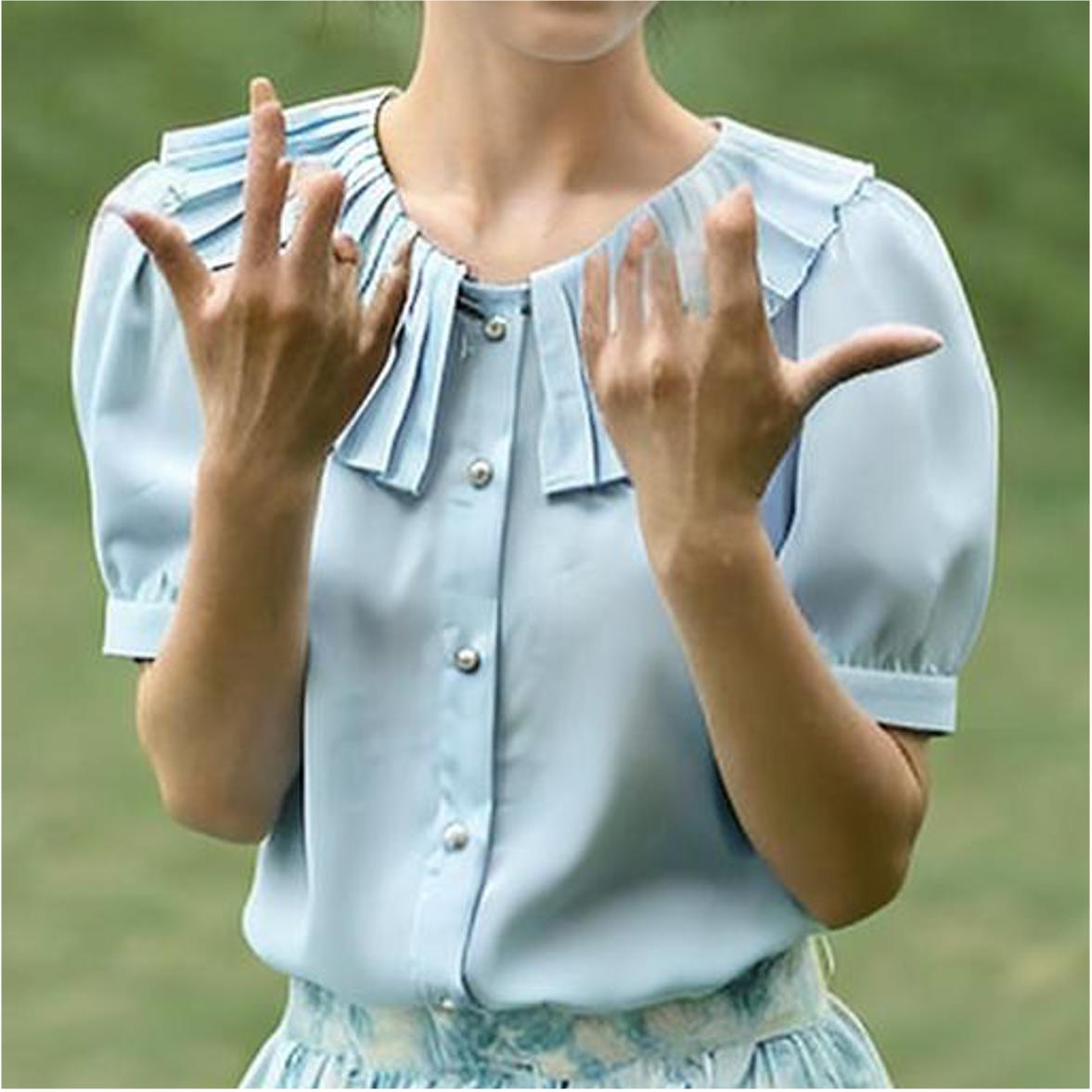} & 
        \includegraphics[width=0.12\linewidth]{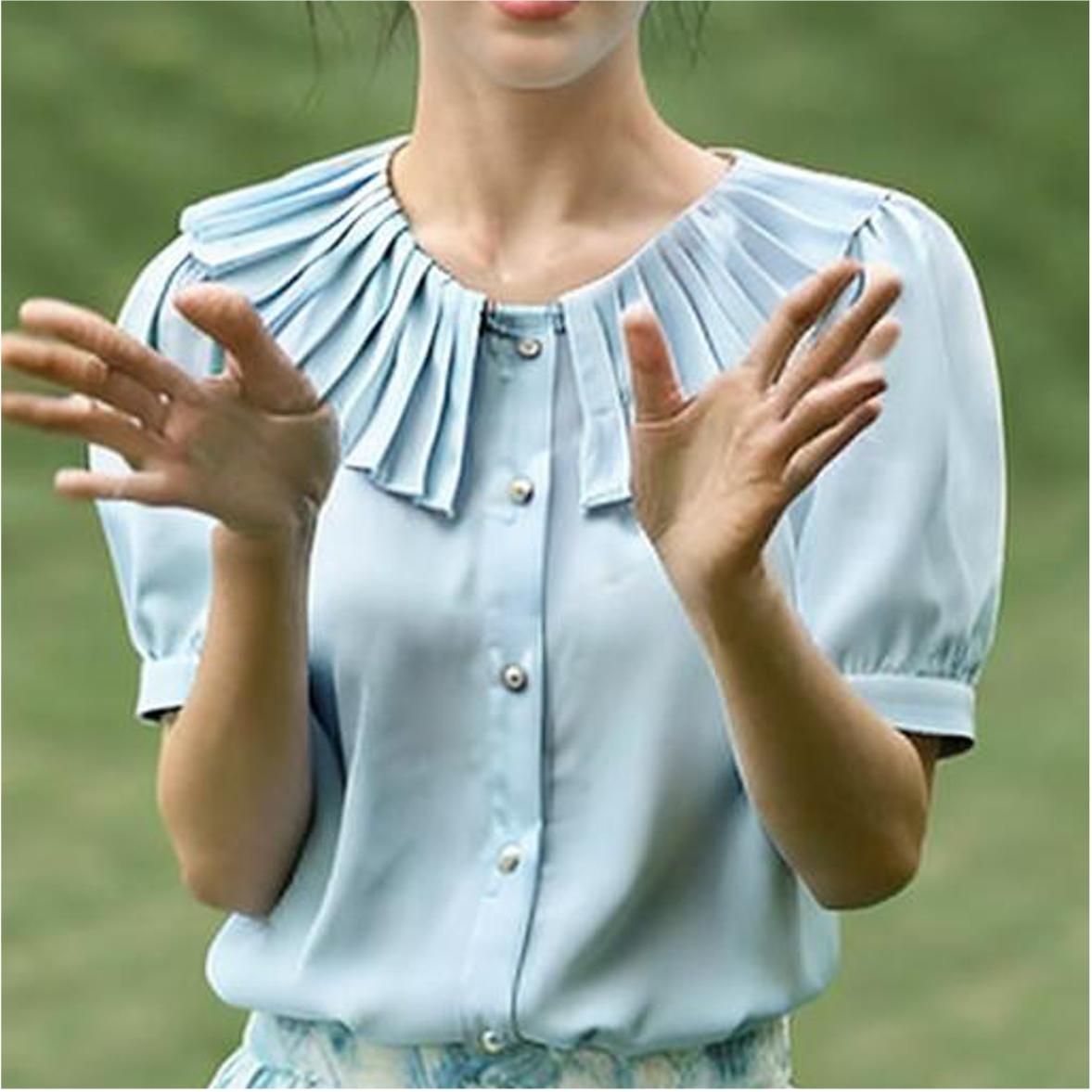} &
        ~&
        \includegraphics[width=0.12\linewidth]{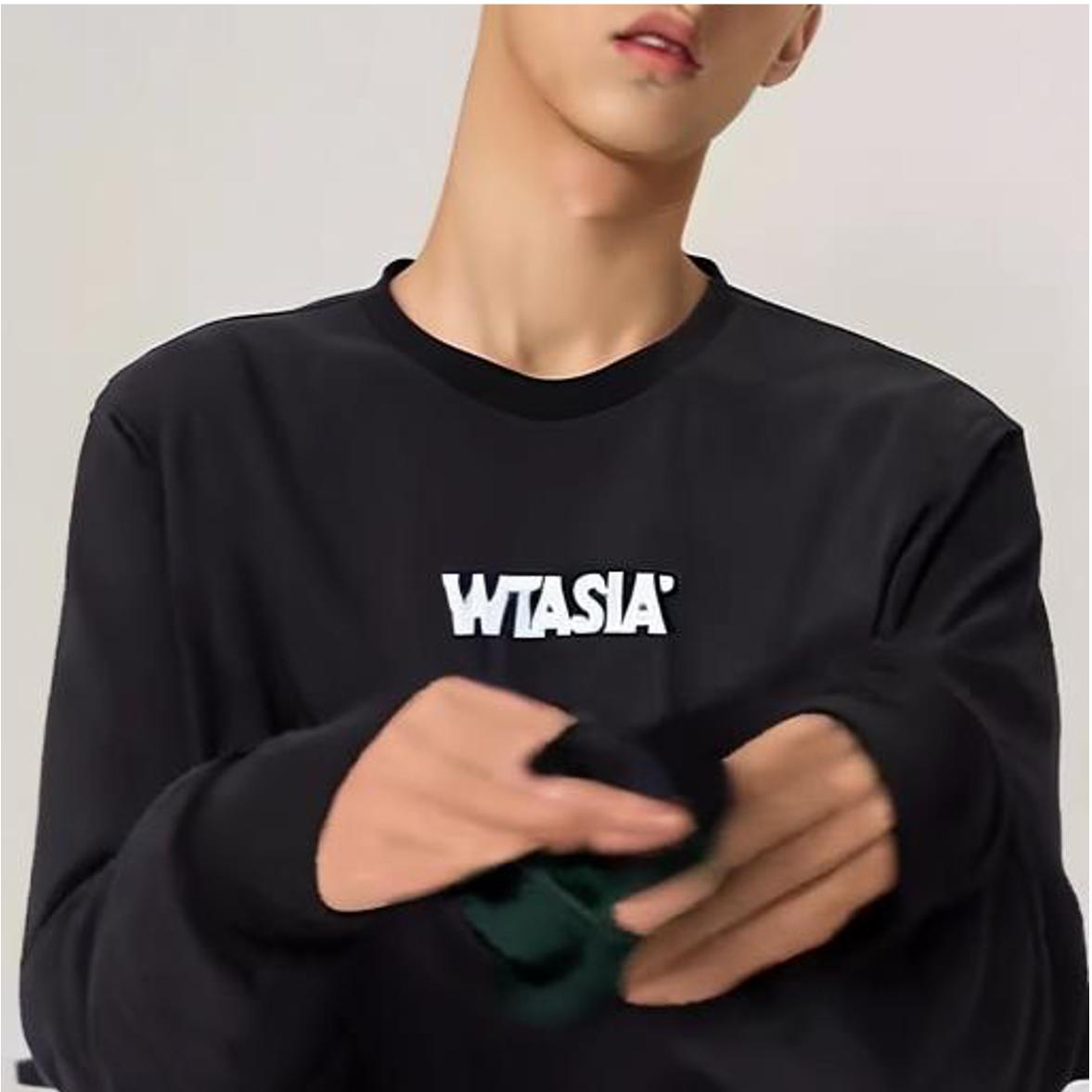} & 
        \includegraphics[width=0.12\linewidth]{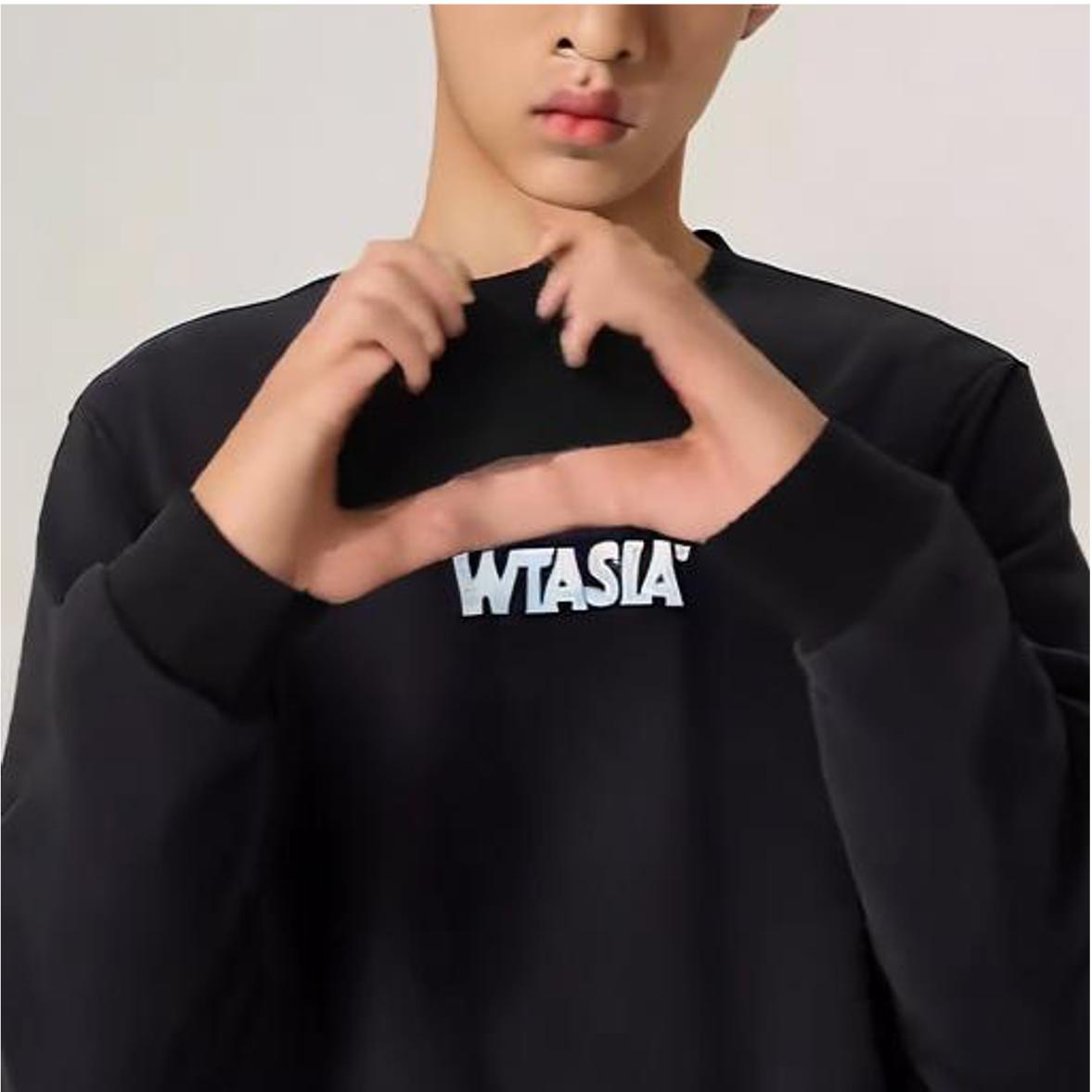} &
        \includegraphics[width=0.12\linewidth]{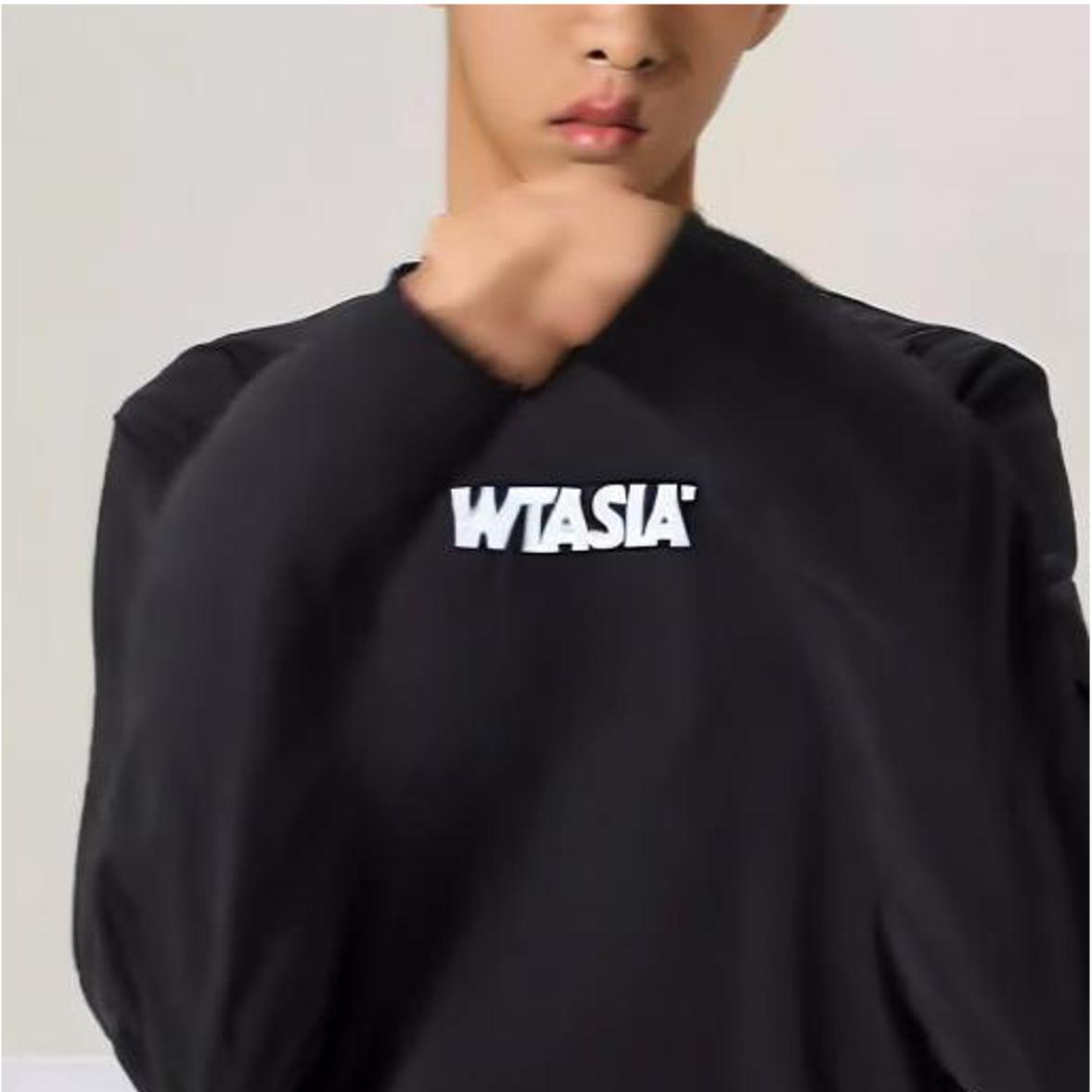} & 
        \includegraphics[width=0.12\linewidth]{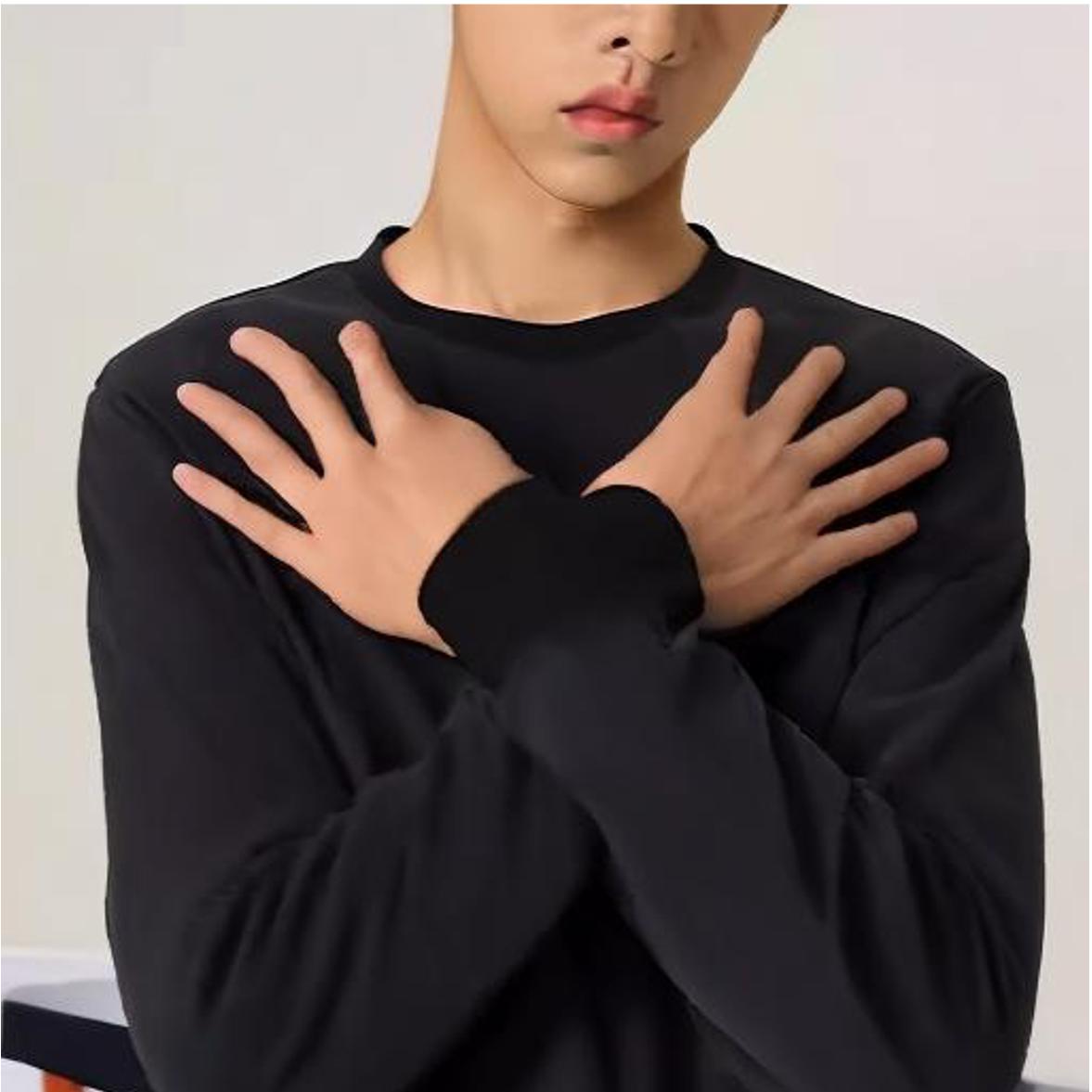} \\
        \rotatebox{90}{~~~~~~{Ours}} &
        \includegraphics[width=0.12\linewidth]{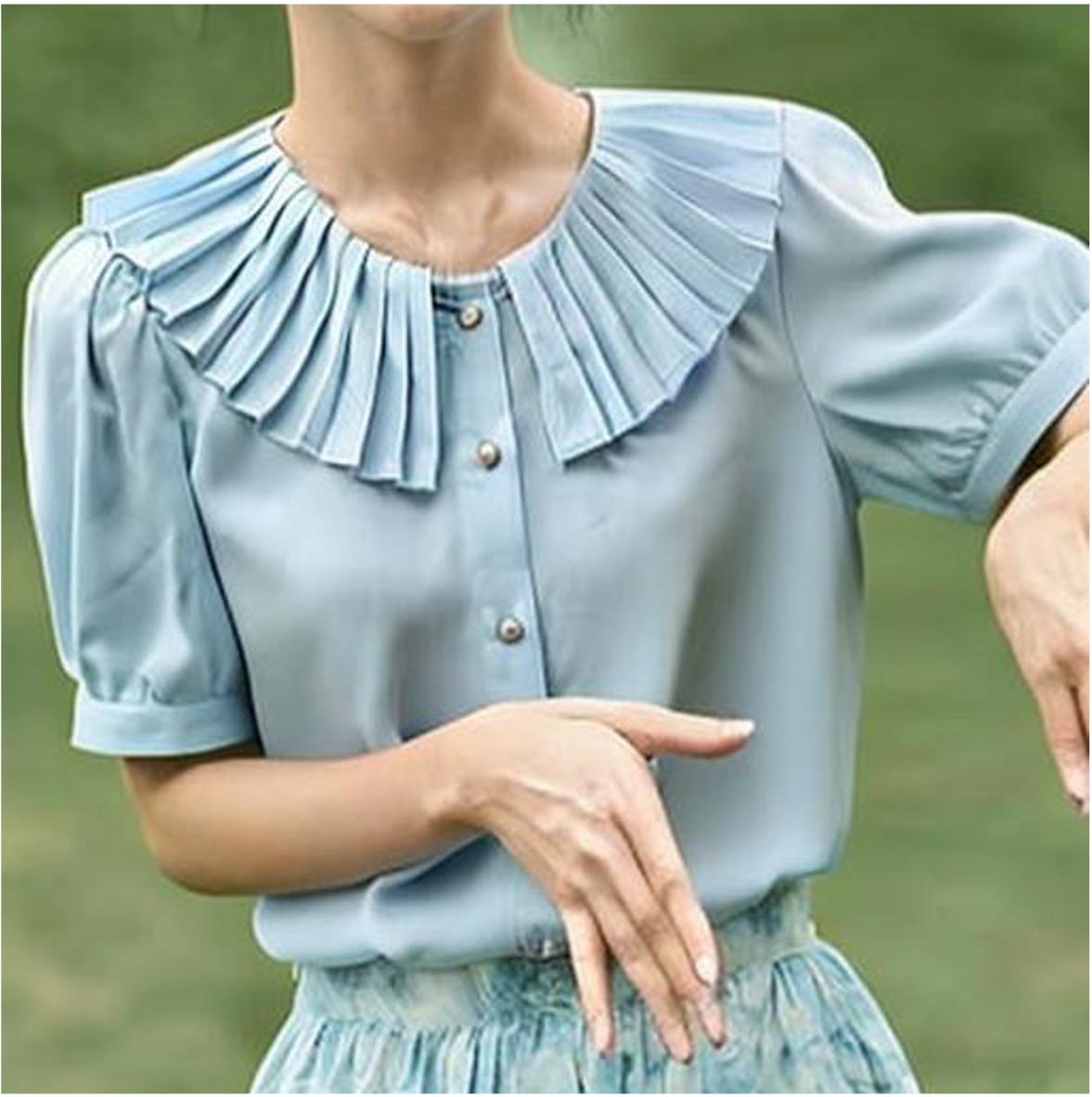} &
        \includegraphics[width=0.12\linewidth]{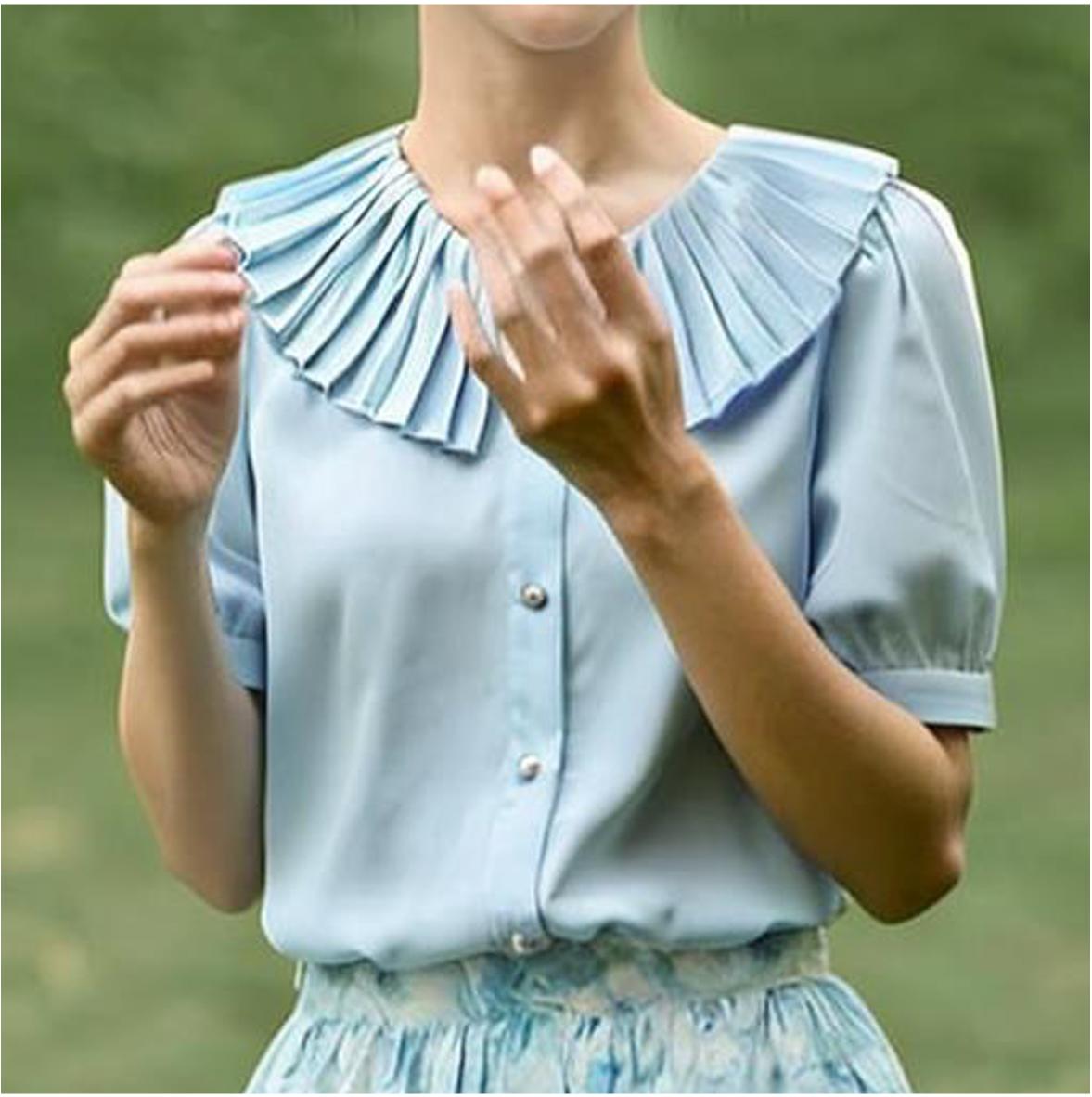} & 
        \includegraphics[width=0.12\linewidth]{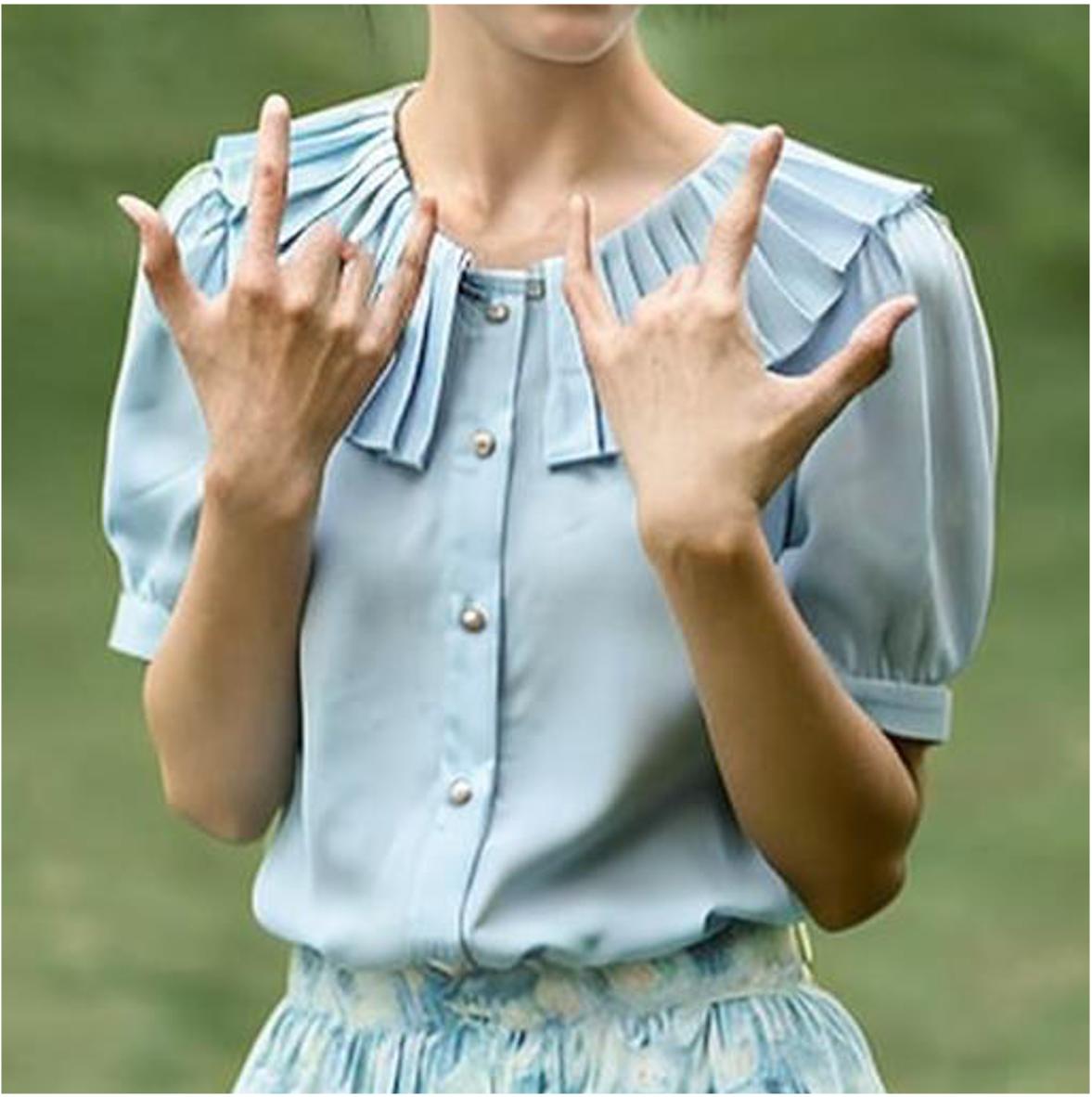} & 
        \includegraphics[width=0.12\linewidth]{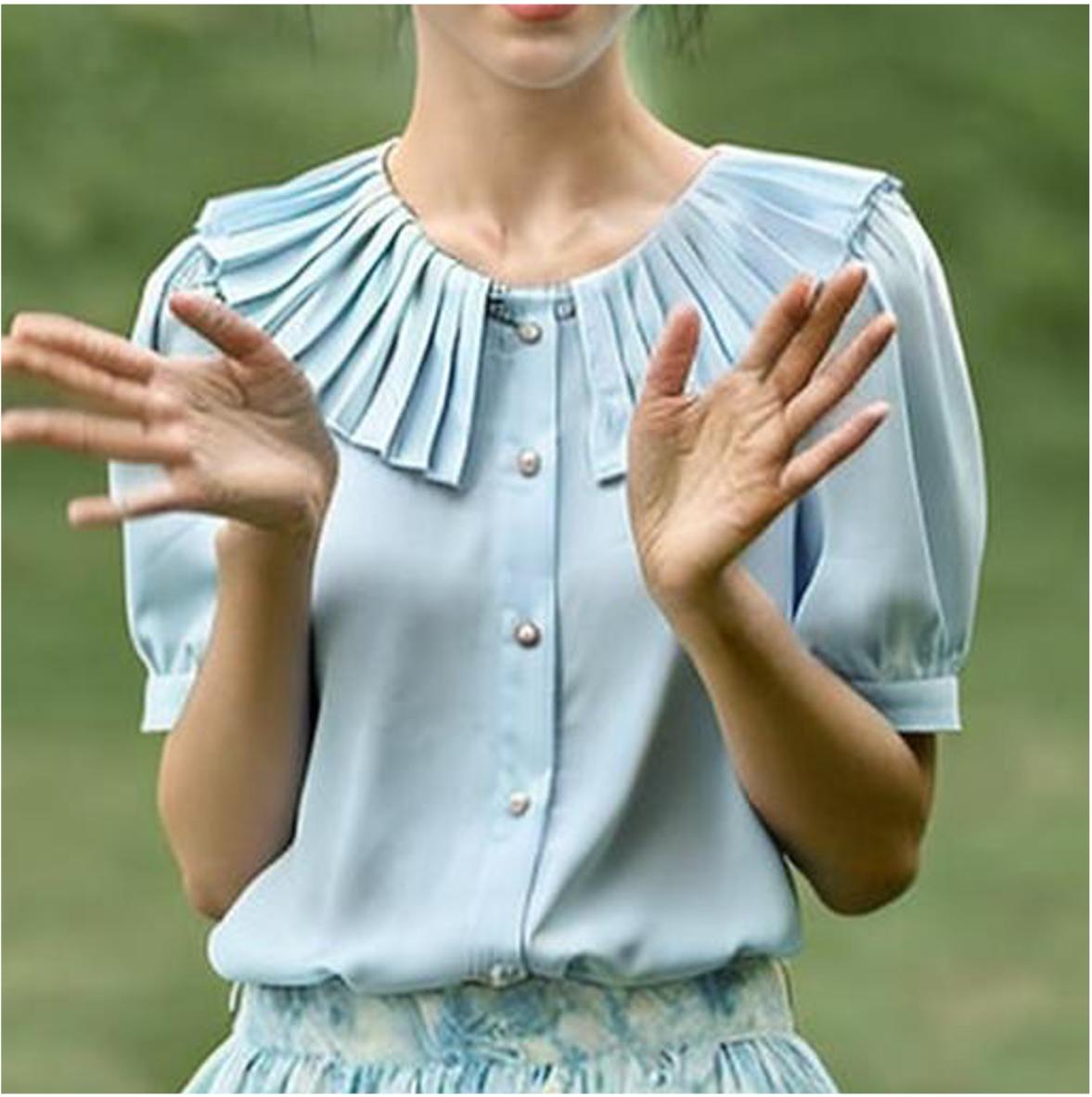} &
        ~&
        \includegraphics[width=0.12\linewidth]{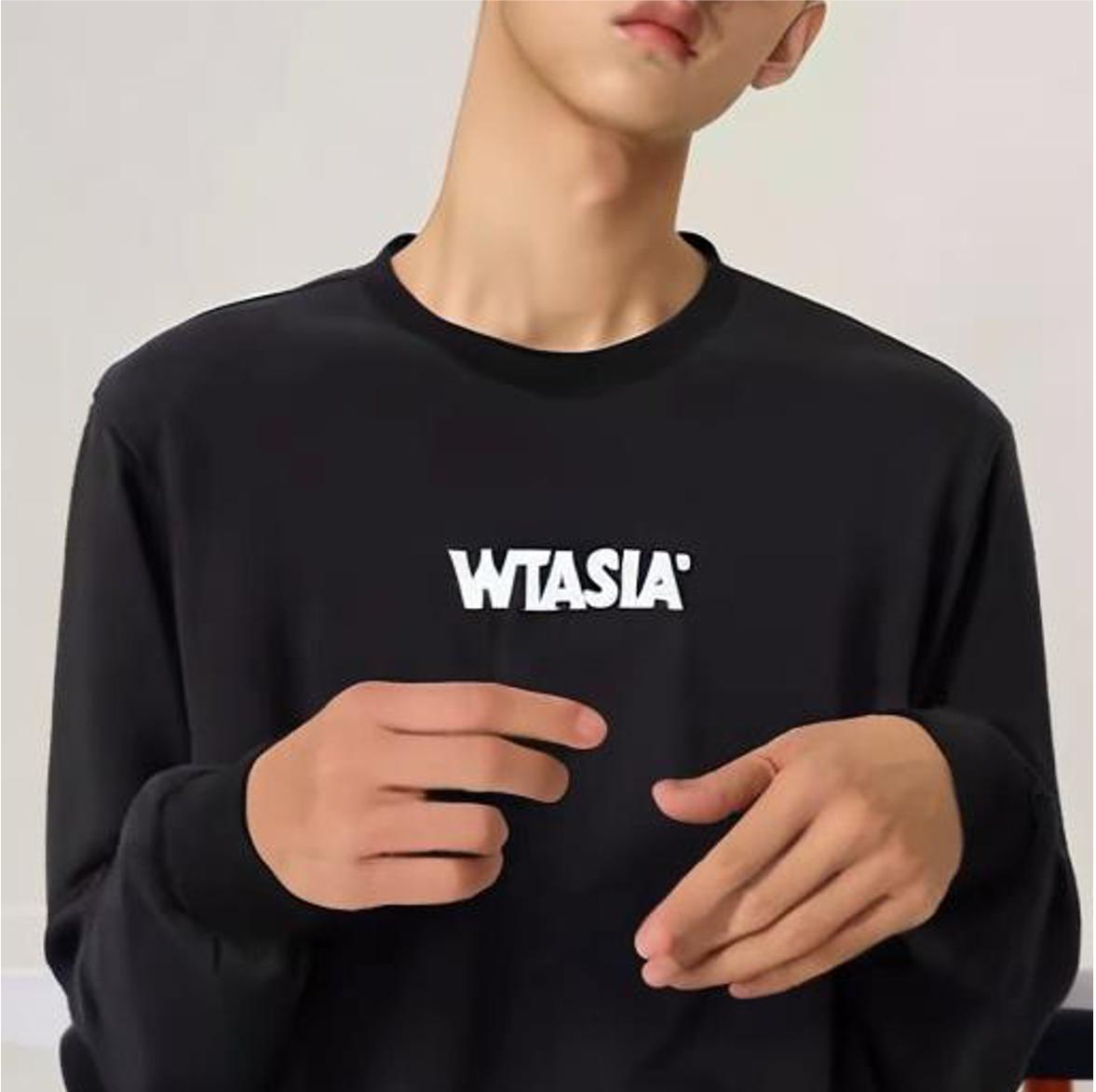} & 
        \includegraphics[width=0.12\linewidth]{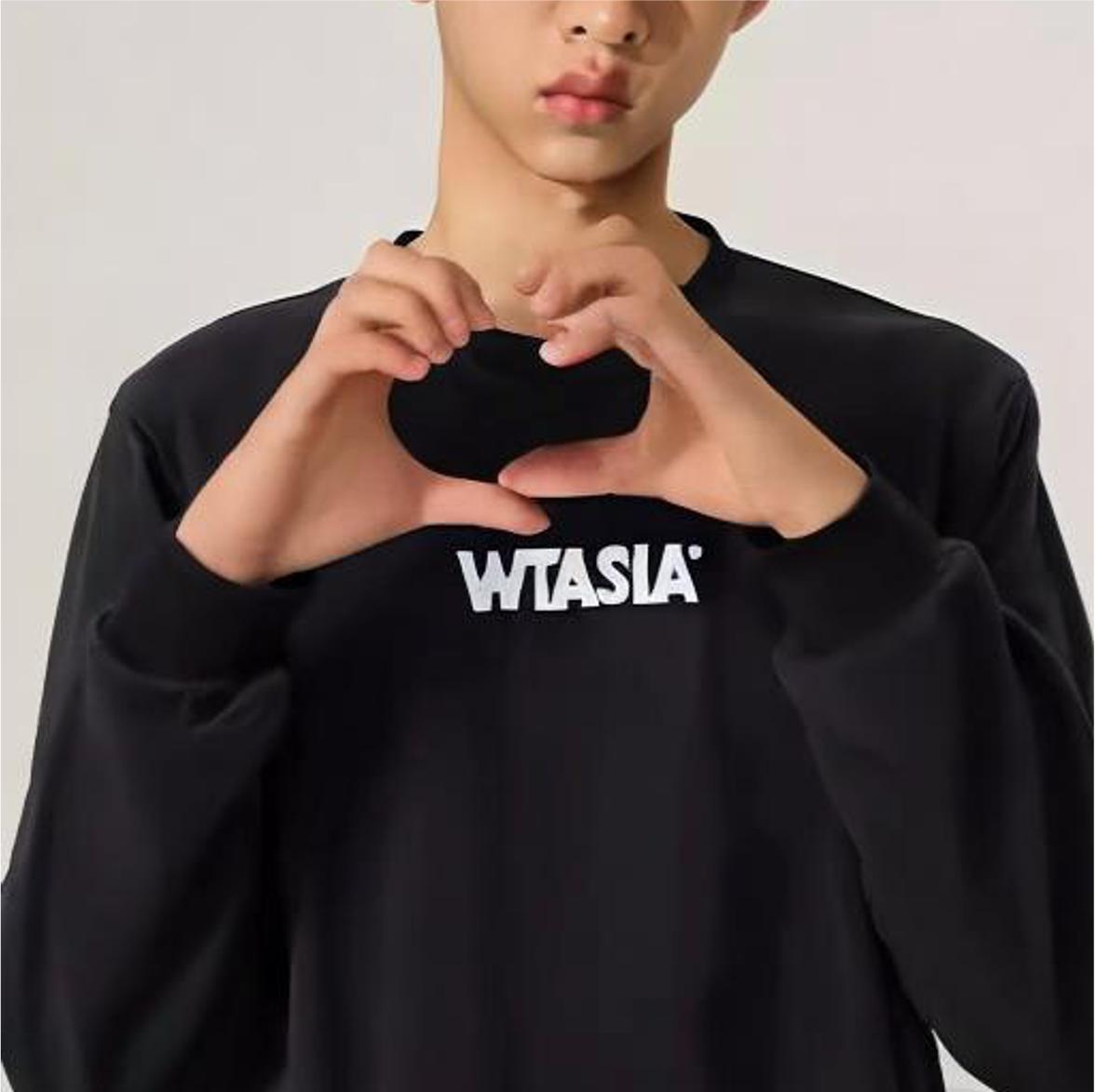} & 
        \includegraphics[width=0.12\linewidth]{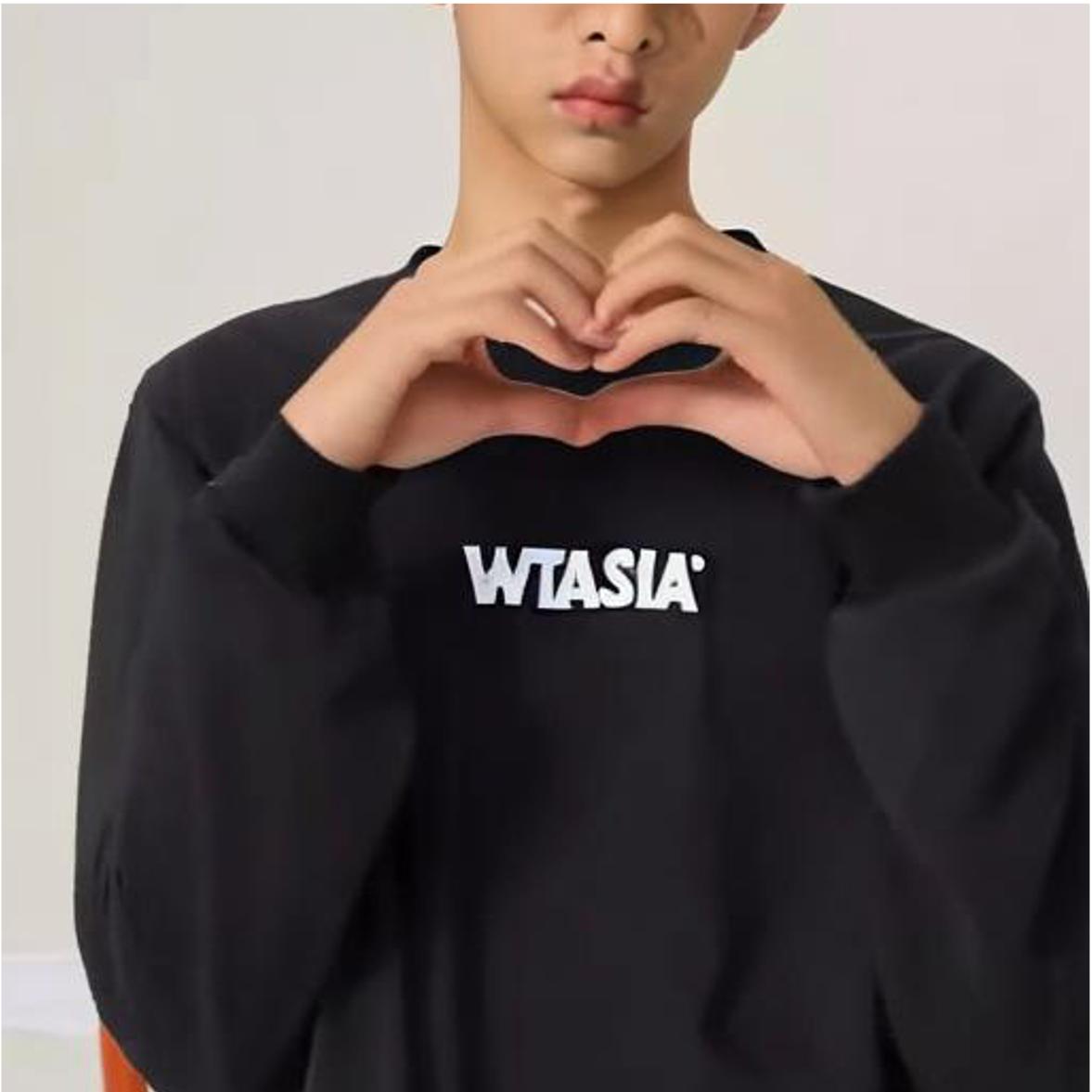} & 
        \includegraphics[width=0.12\linewidth]{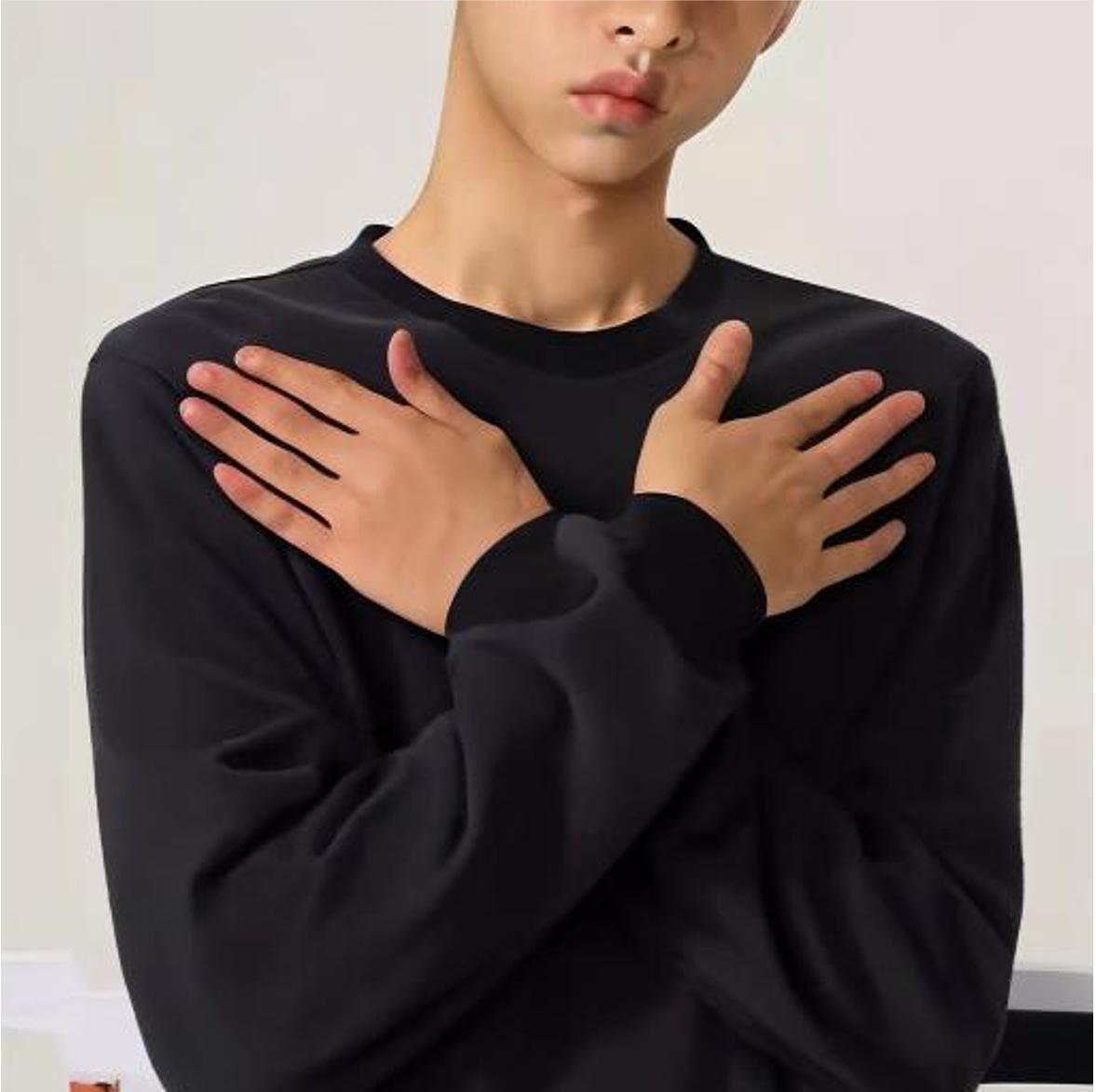} \\
        \rotatebox{90}{~~~~~~~AA} &
        \includegraphics[width=0.12\linewidth]{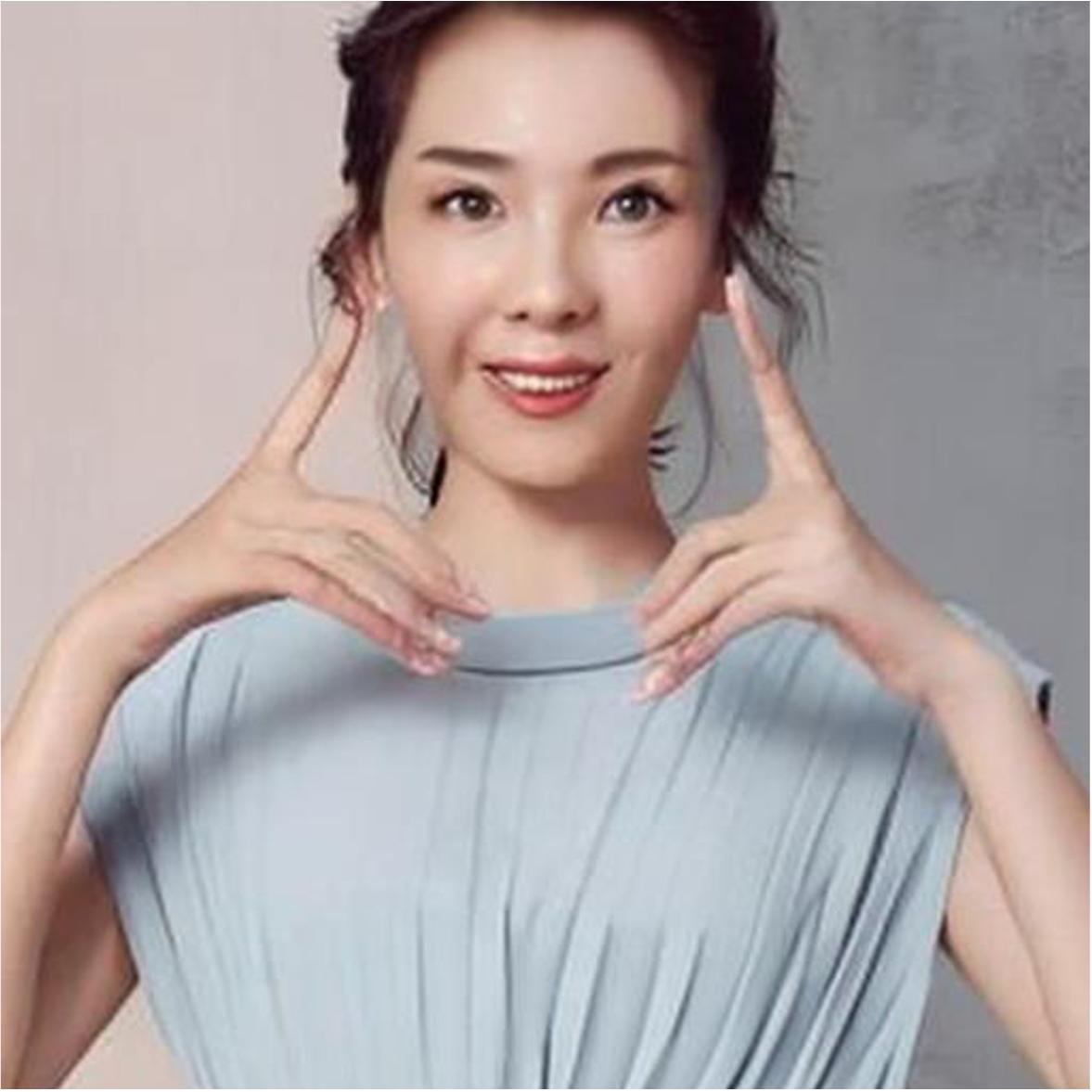} &
        \includegraphics[width=0.12\linewidth]{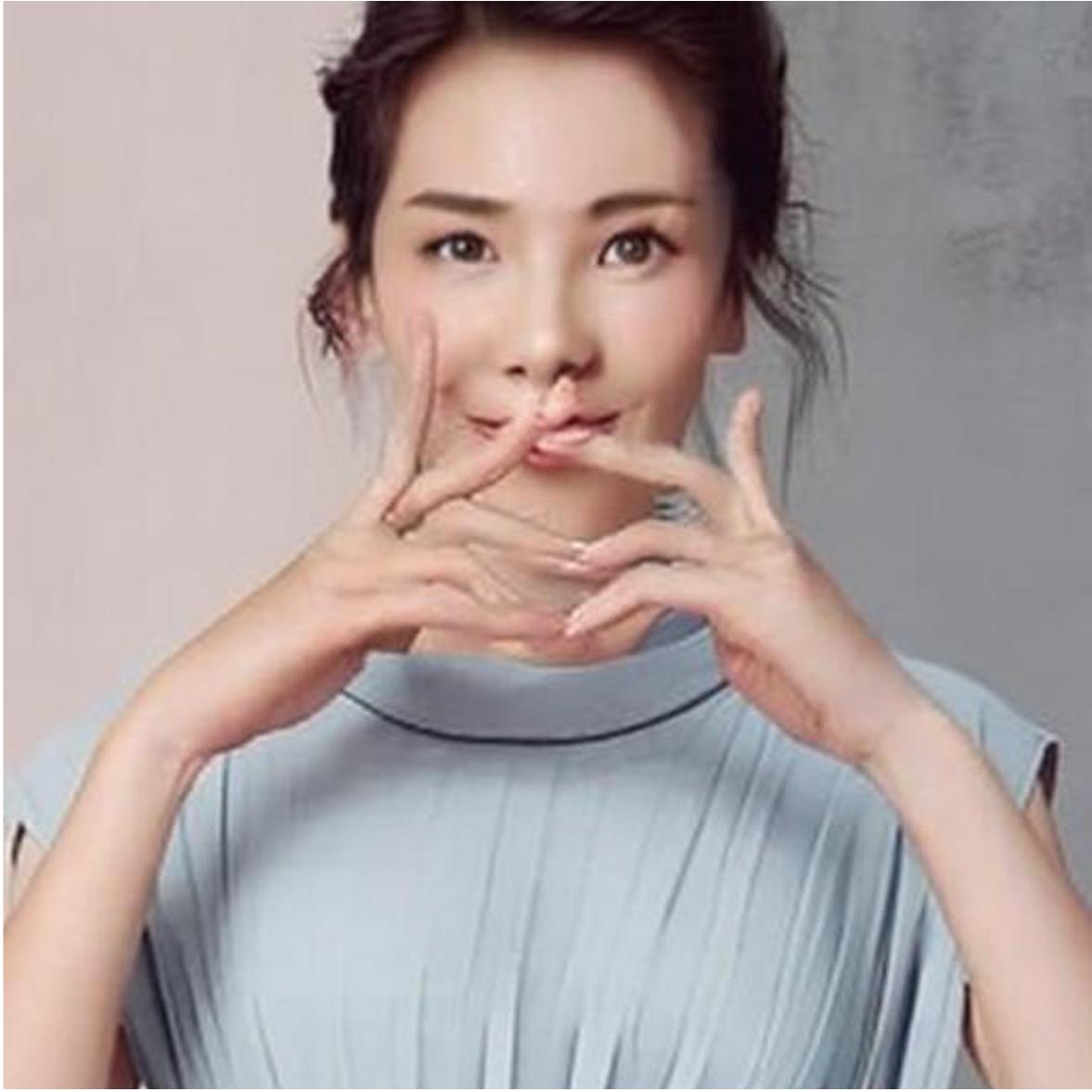} & 
        \includegraphics[width=0.12\linewidth]{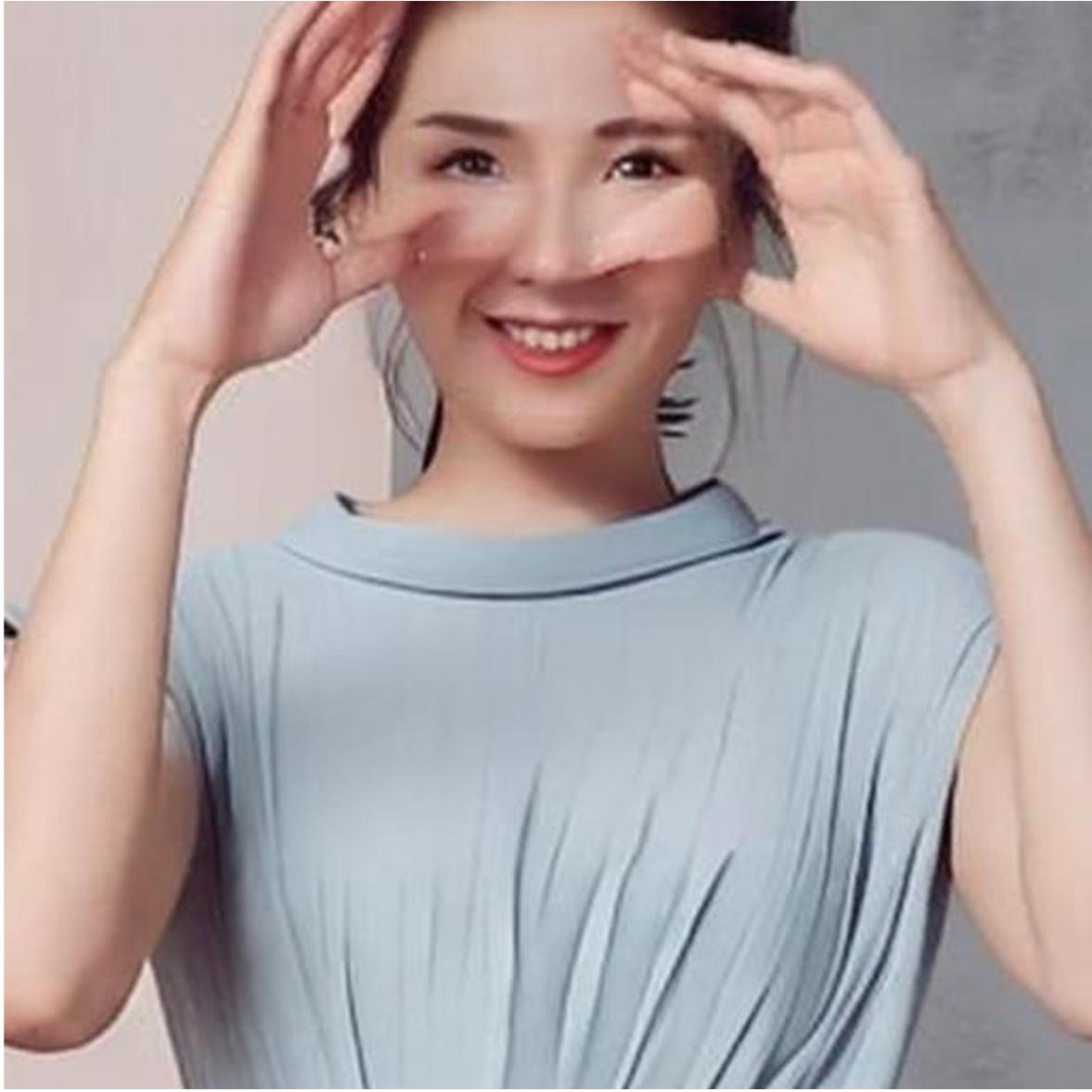} & 
        \includegraphics[width=0.12\linewidth]{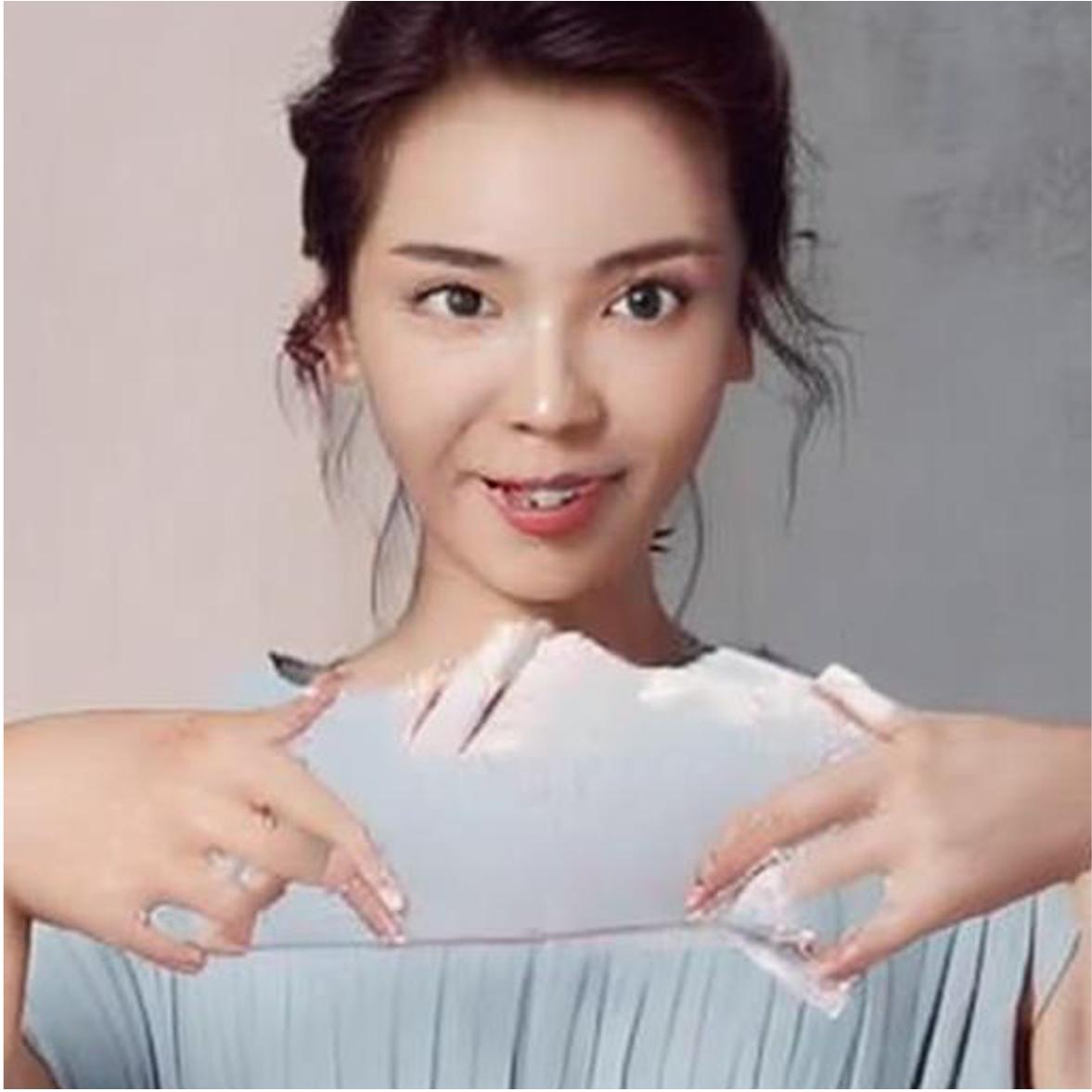} &
        ~&
        \includegraphics[width=0.12\linewidth]{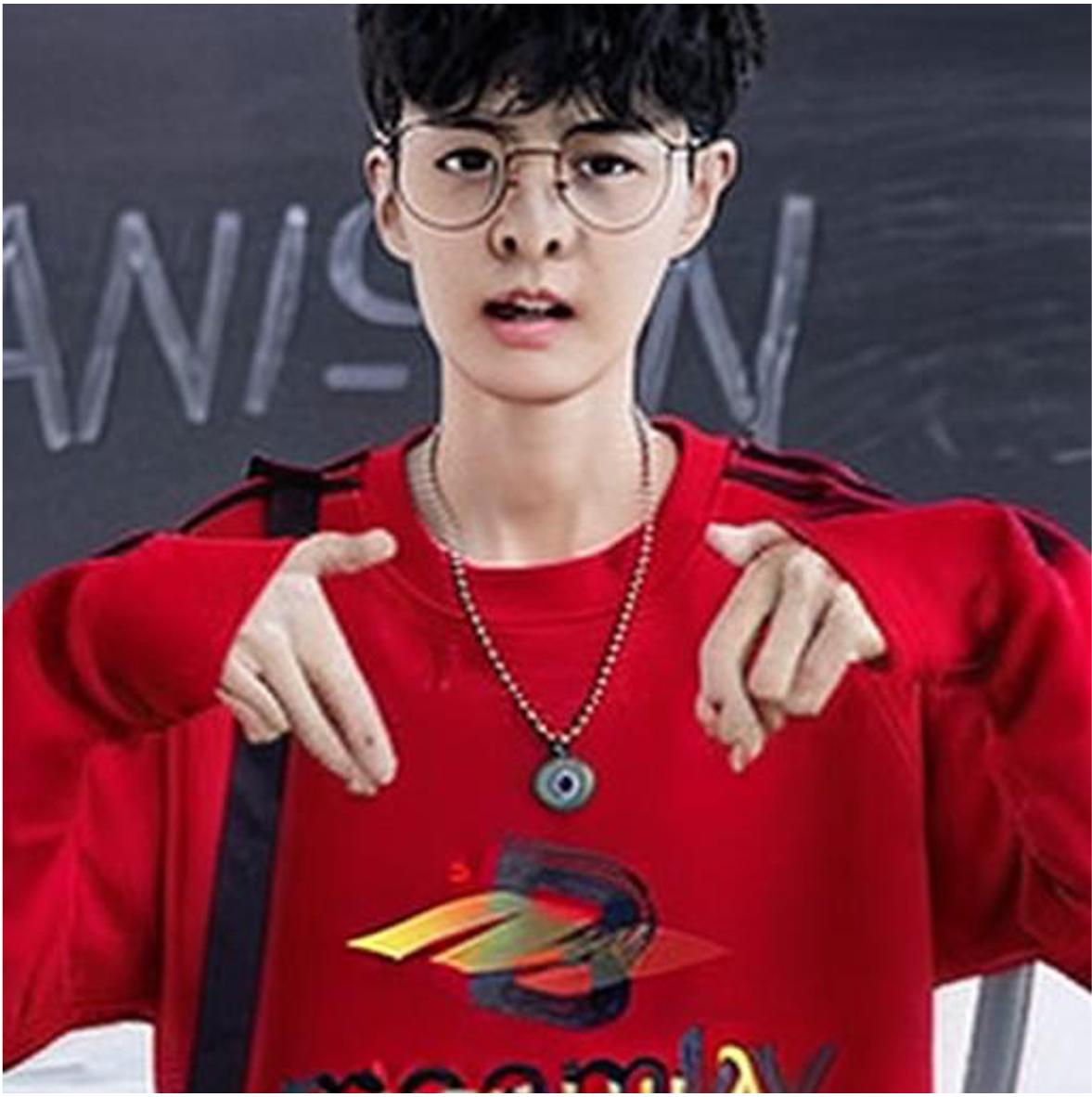} &
        \includegraphics[width=0.12\linewidth]{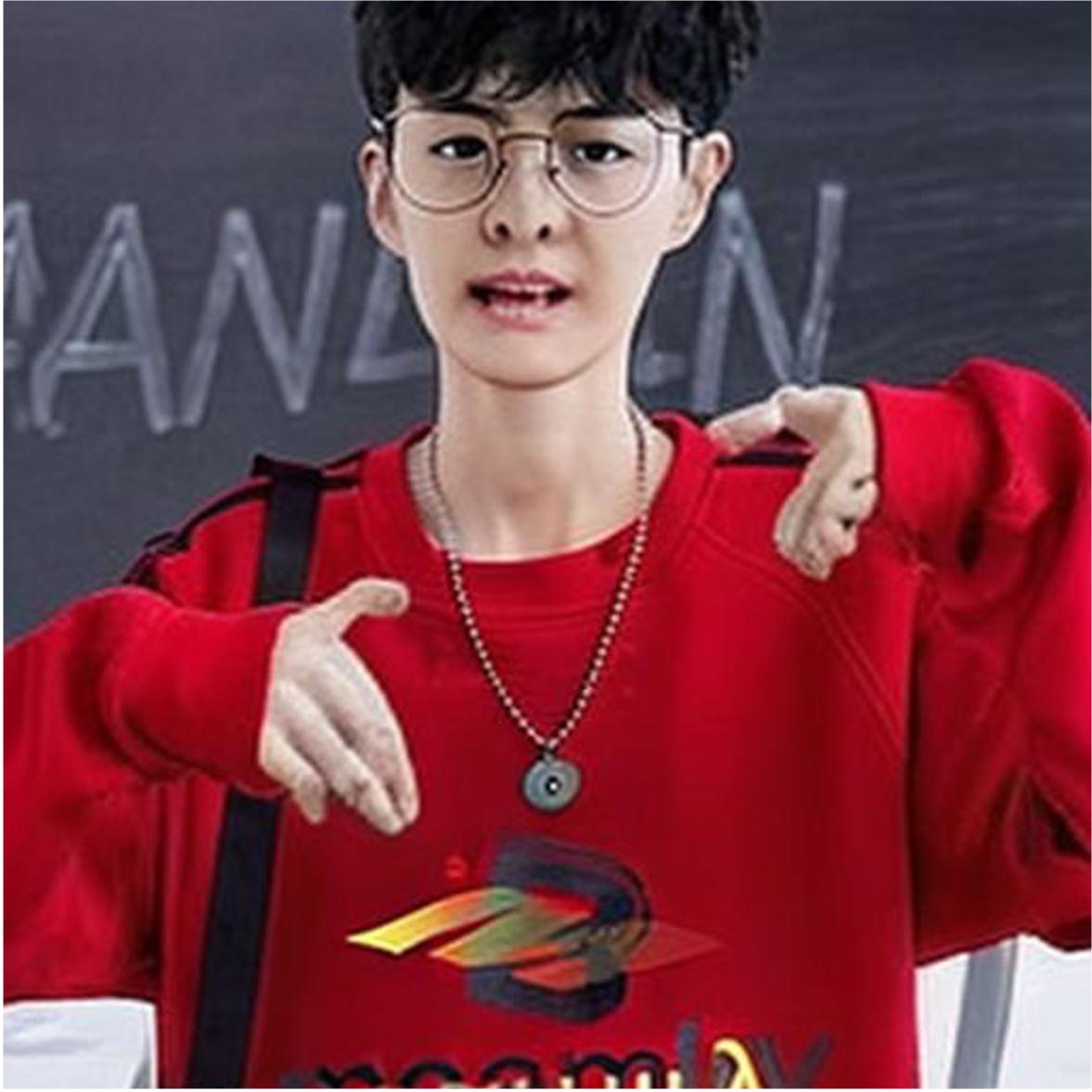} & 
        \includegraphics[width=0.12\linewidth]{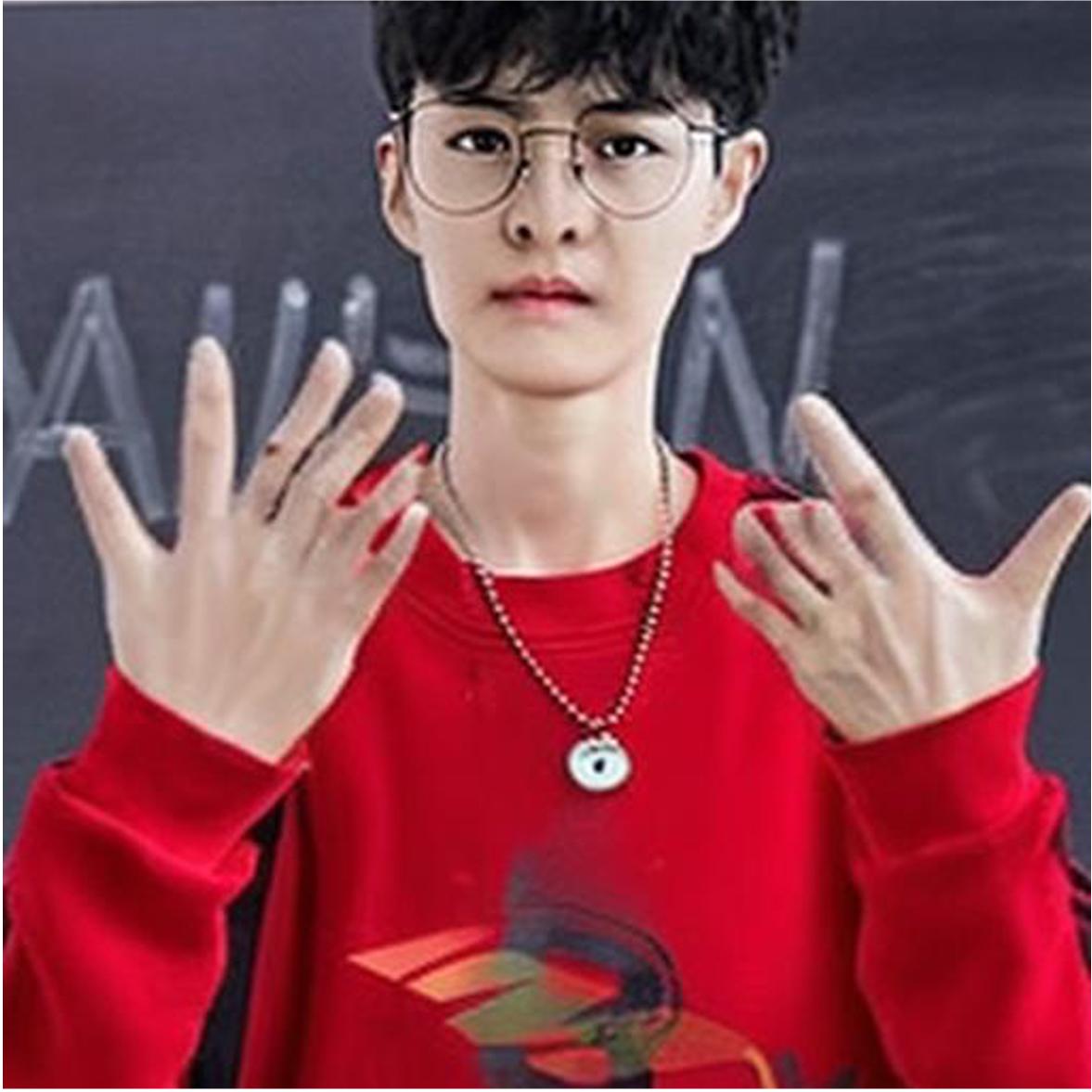} & 
        \includegraphics[width=0.12\linewidth]{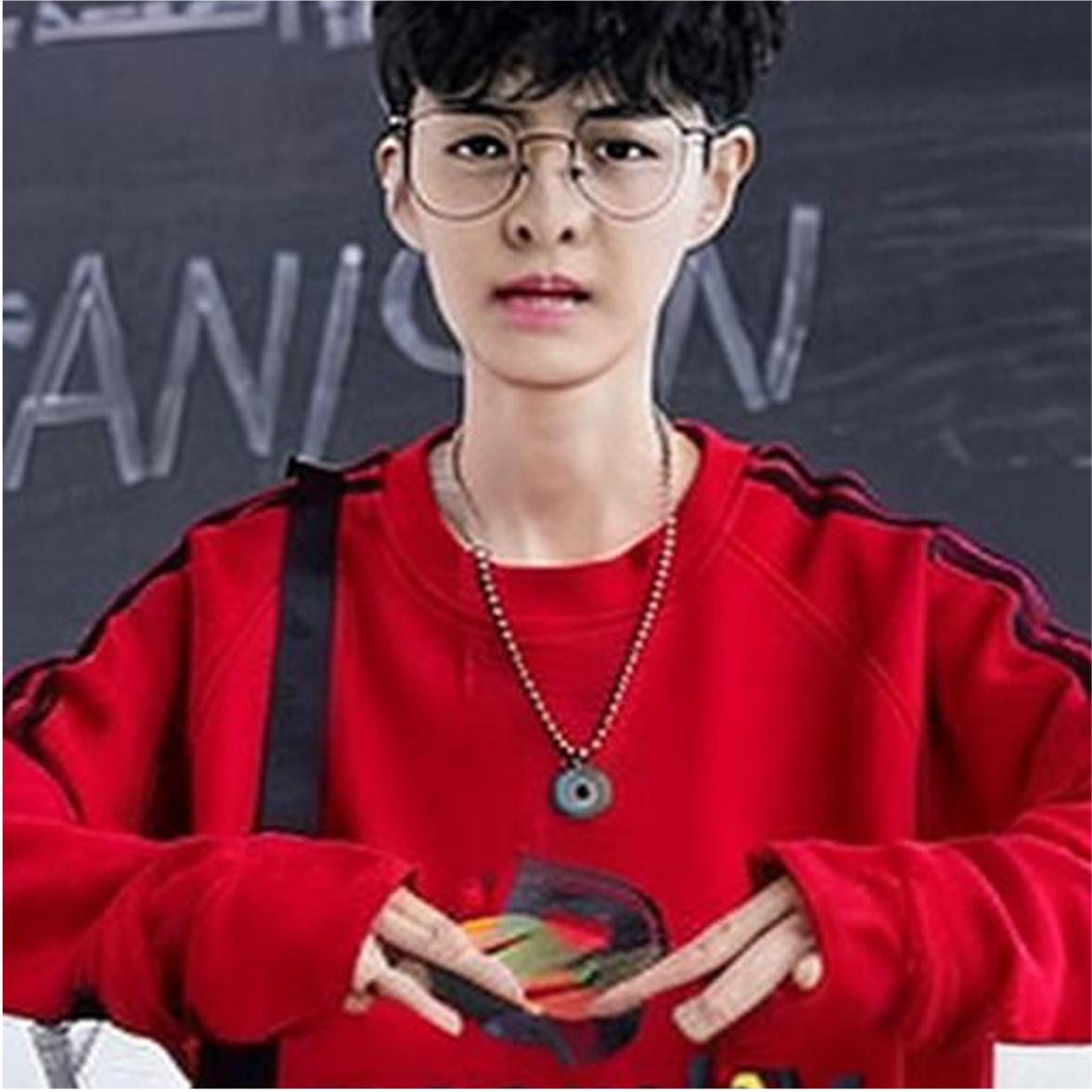} \\
        \rotatebox{90}{~~~~~~{Ours}} &
        \includegraphics[width=0.12\linewidth]{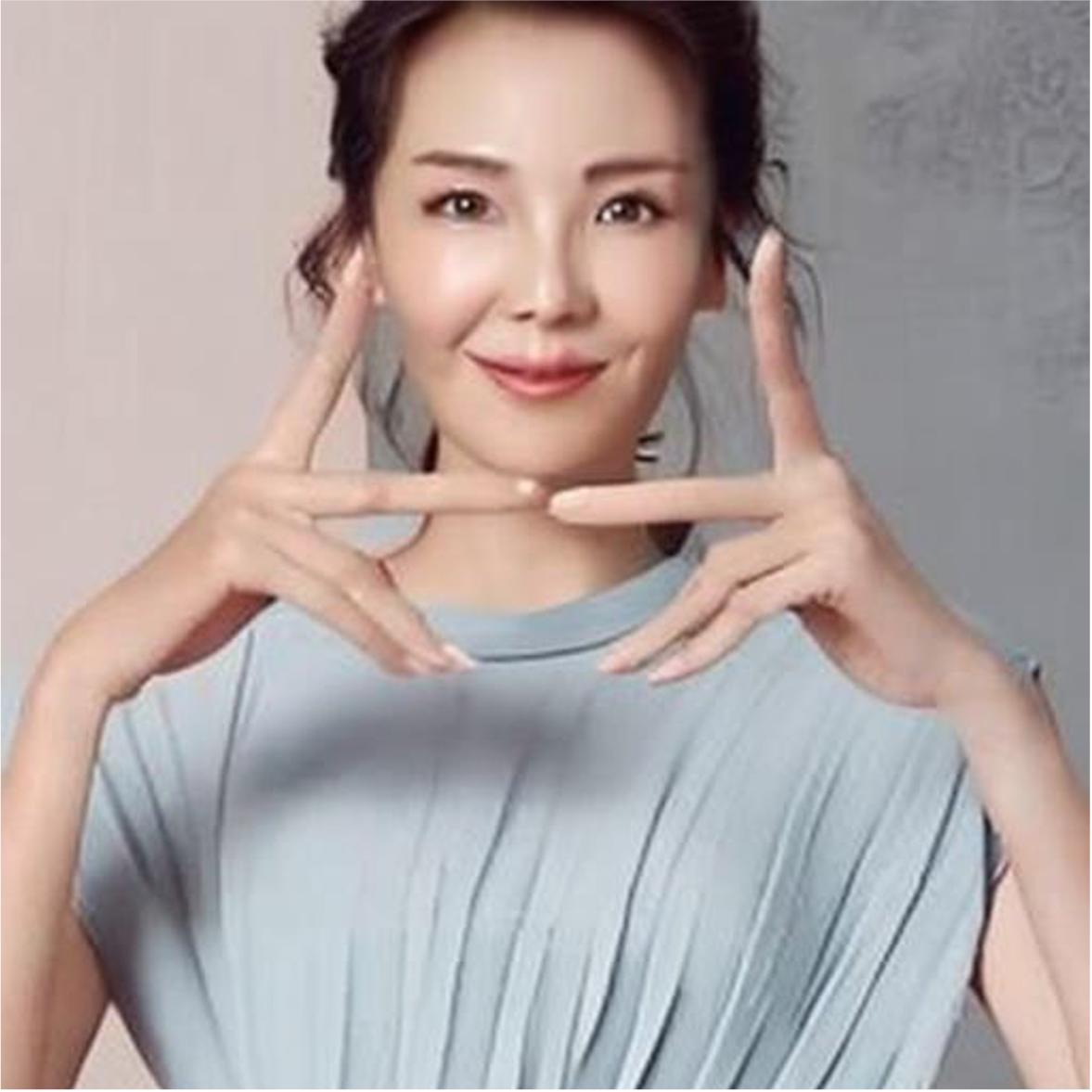} &
        \includegraphics[width=0.12\linewidth]{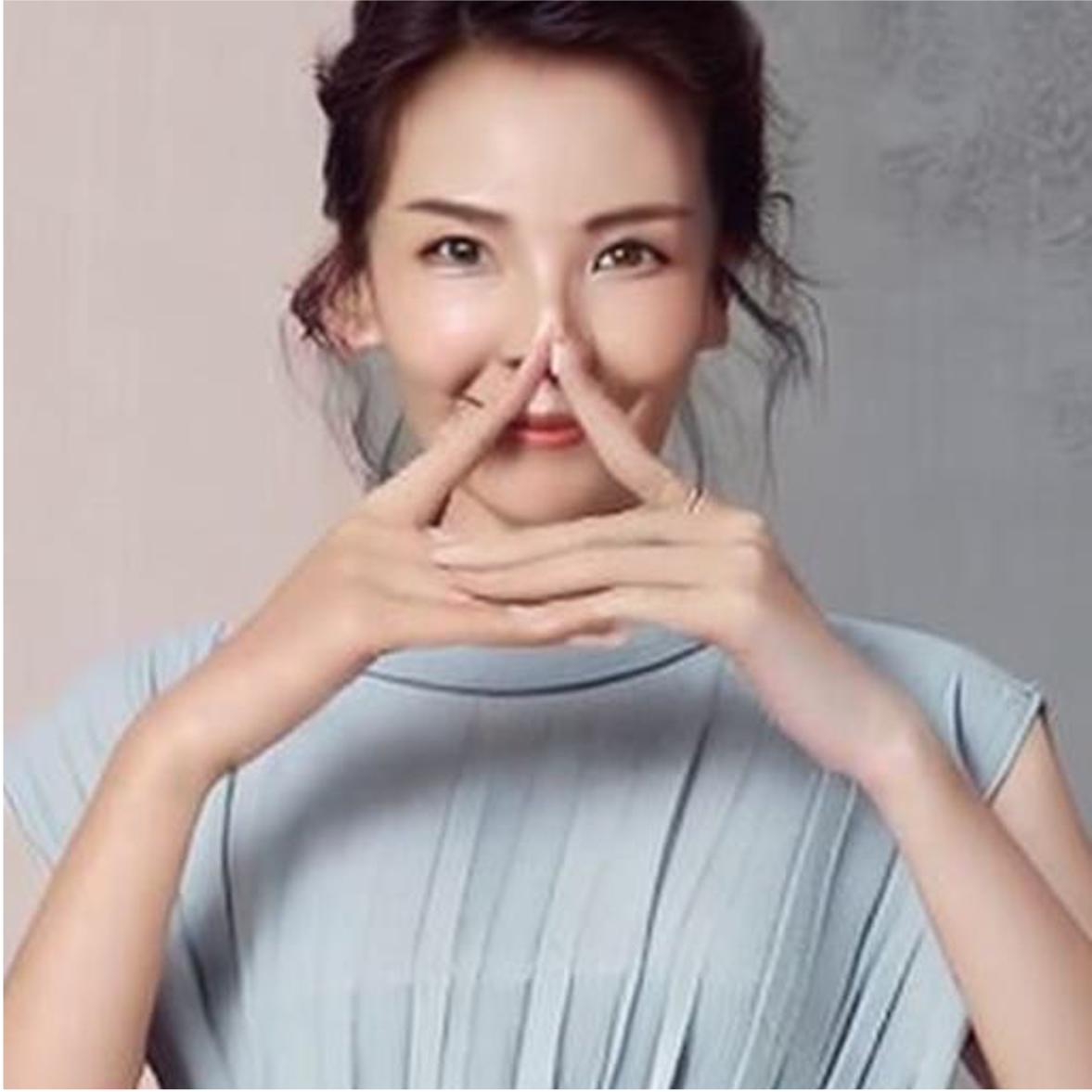} & 
        \includegraphics[width=0.12\linewidth]{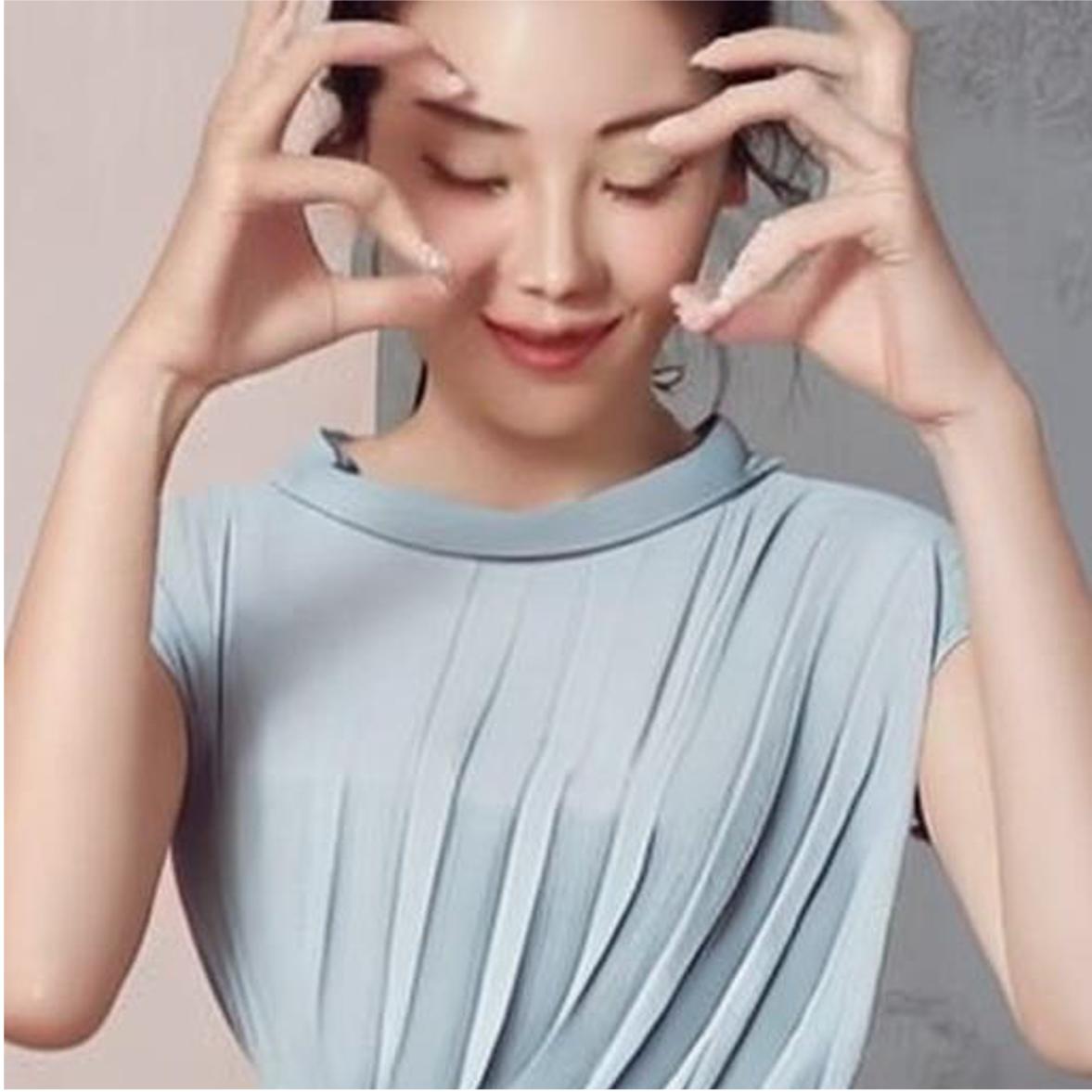} & 
        \includegraphics[width=0.12\linewidth]{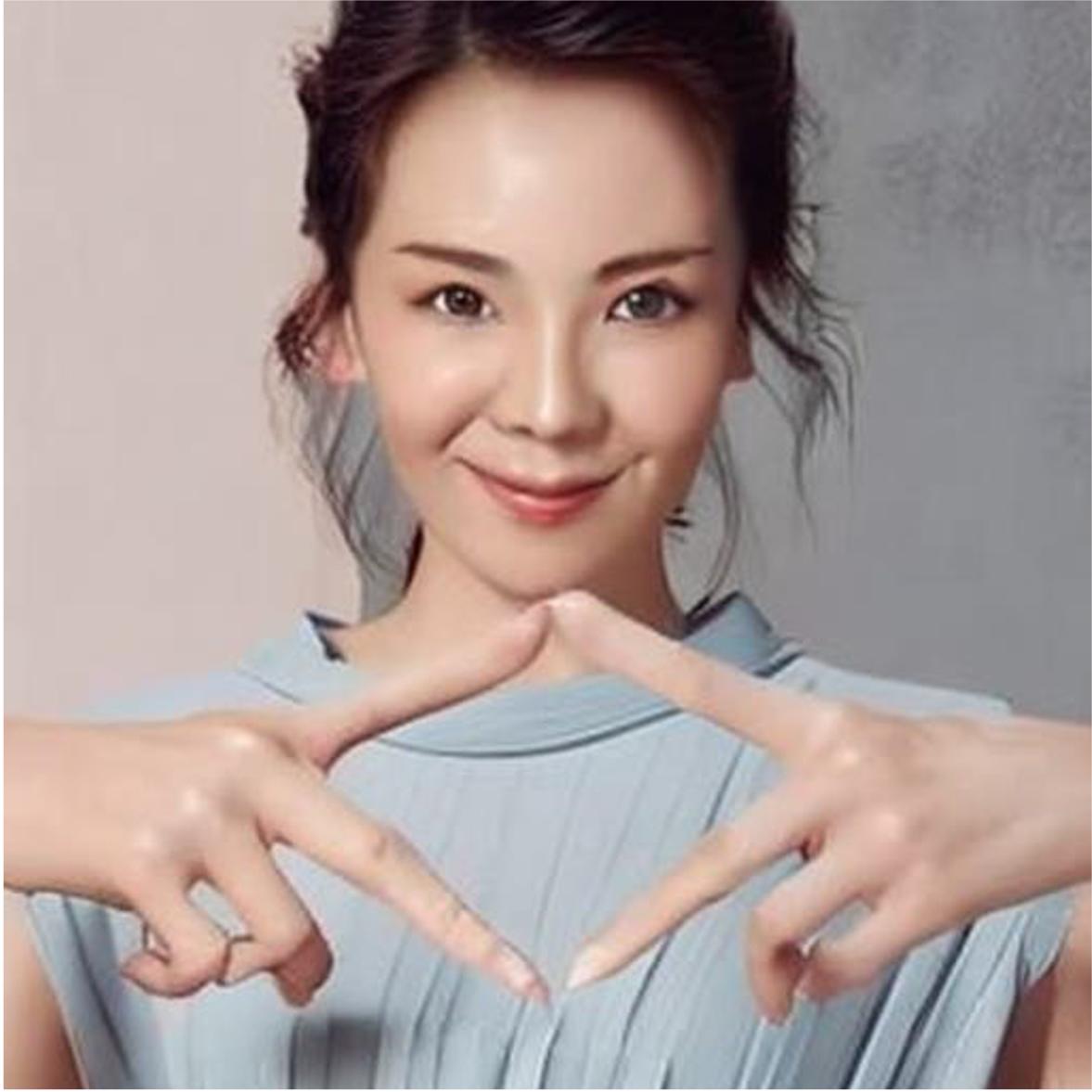} &
        ~&
        \includegraphics[width=0.12\linewidth]{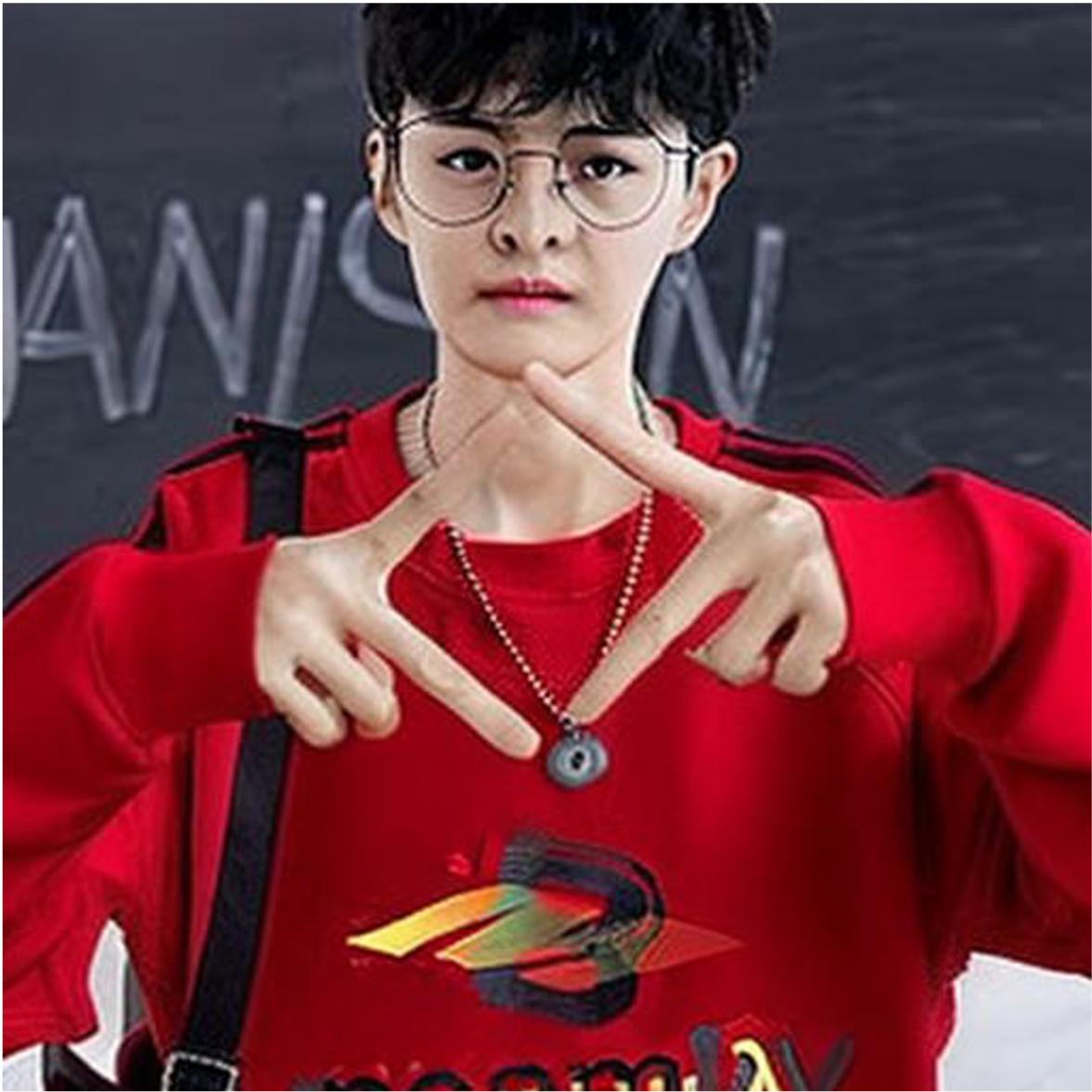} &
        \includegraphics[width=0.12\linewidth]{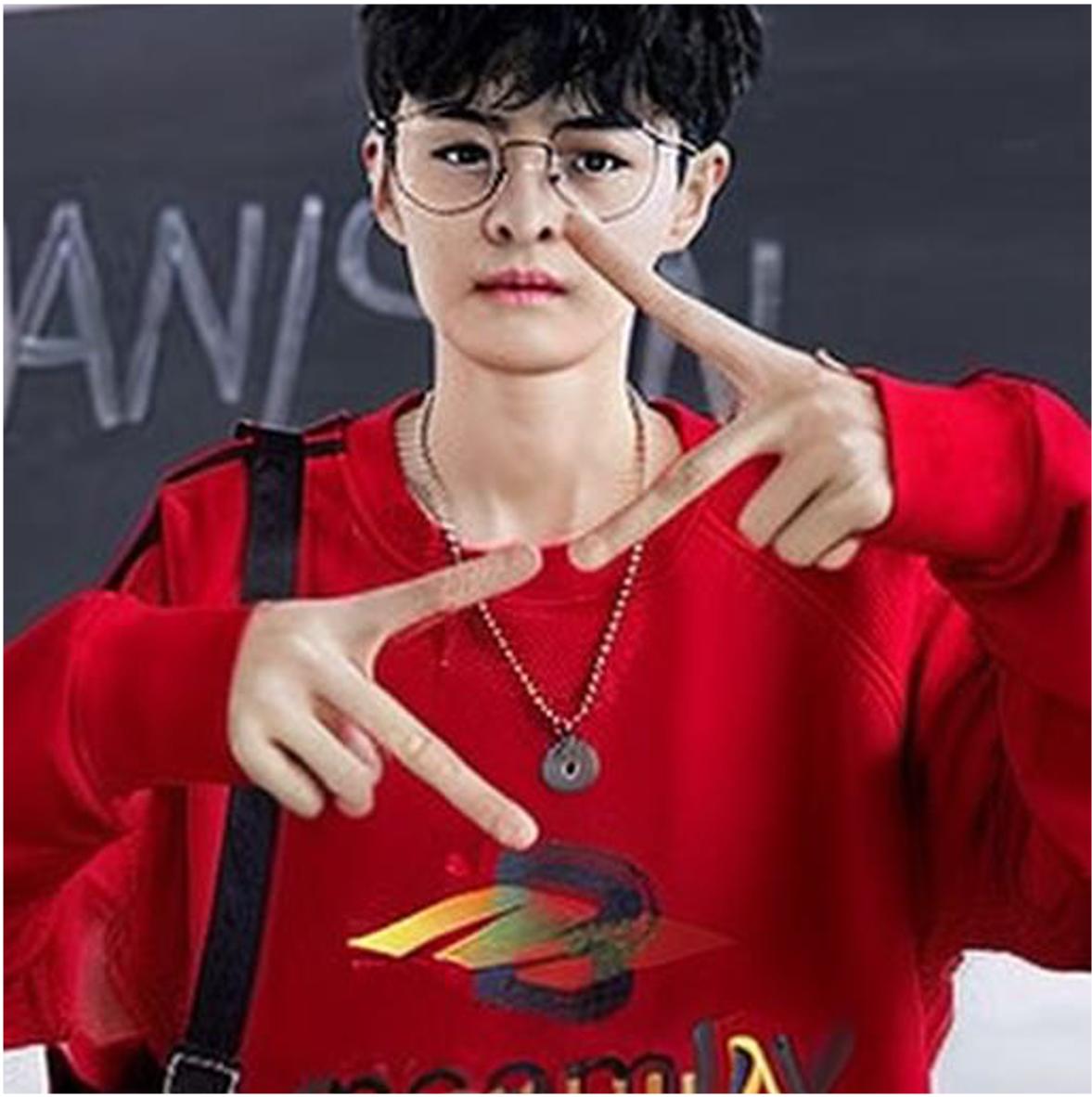} & 
        \includegraphics[width=0.12\linewidth]{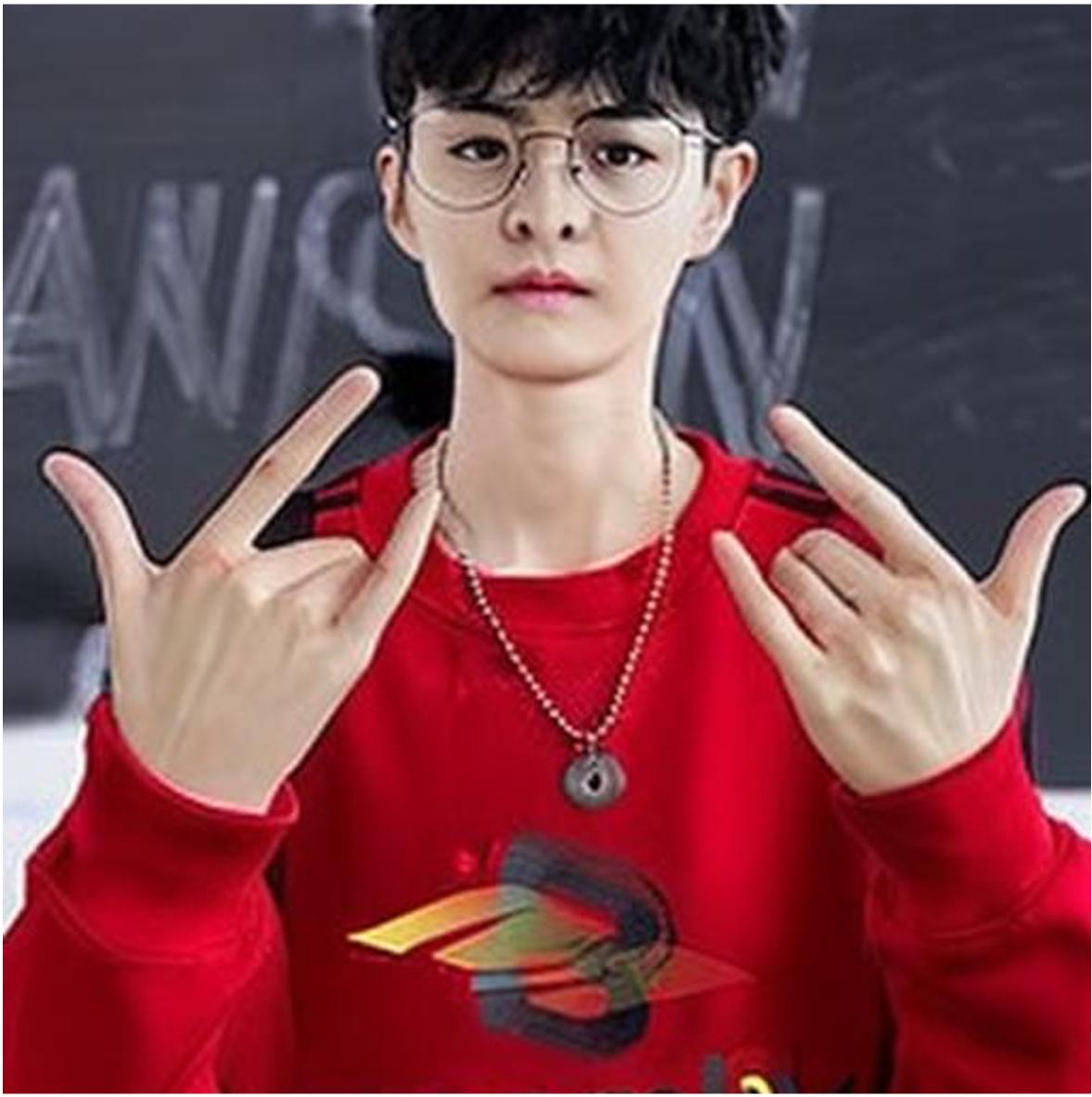} & 
        \includegraphics[width=0.12\linewidth]{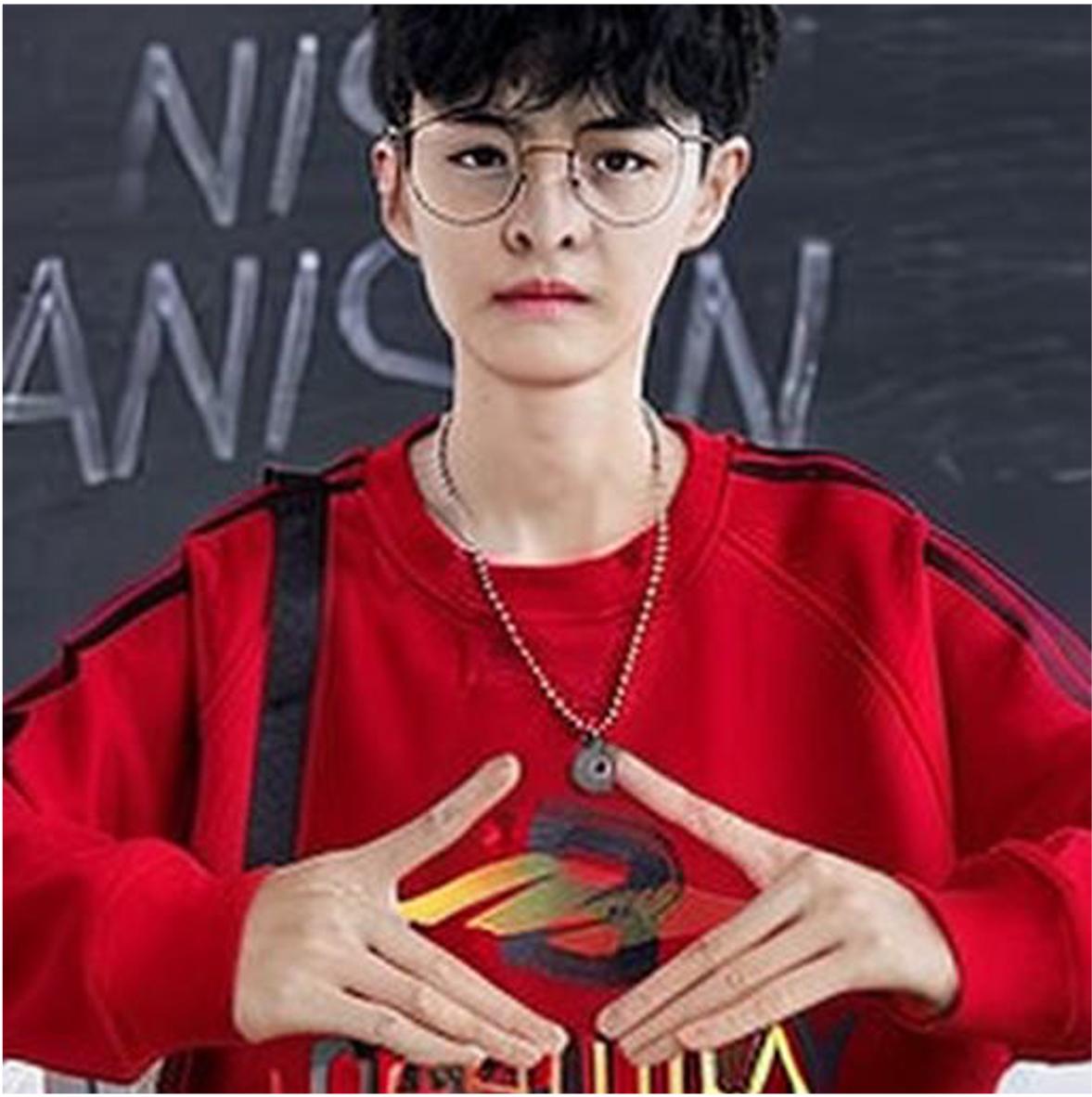} 
    \end{tabular}
    \caption{Comparisons of hand quality between RealisDance and our reproduced Animate Anyone. Thanks to the HaMeR pose sequence, the hand quality of RealisDance surpasses our reproduced Animate Anyone by a large margin.}
    \label{fig:hand_quality}
\end{figure*}

\vspace{1mm}
\noindent\textbf{Hands quality.}
Figure~\ref{fig:hand_quality} compares the hand quality between RealisDance and our reproduced Animate Anyone. Due to inaccurate hand estimation and lack of 3D/depth information, hands generated using DWPose suffer from issues such as the wrong number of fingers (case 1 frame 1), strange finger lengths (case 3 frame 4), incorrect gestures (case 4 frame 2), artifacts (case 7 frame 4), and blurry hands (case 6 frame 3). Thanks to the capability of HaMeR, RealisDance can generate realistic hands even for complex gestures (case 2 frame 3, case 3 frame 2, case 4 frame 2, case 5 frame 3, and case 6 frame 3).

\vspace{1mm}
\noindent\textbf{Video smoothness.} \href{https://damo-vision.com/}{This link} compares the video smoothness between RealisDance and Moore-Animate Anyone~\cite{moore_aa}, as can be seen, the video generated by Moore-Animate Anyone is shaky because the pose sequence is not smooth enough. This also demonstrates our observation that the influence of pose control is so dominant that temporal attention used in the main UNet is insufficient to mitigate video shaking completely. Thanks to the motion module in the pose guidance network, the proposed RealisDance can generate smooth videos even if the pose sequence is shaky.
\section{Conlcusion and Limitation}
In this paper, we focus on improving pose control of existing controllable character animation methods and introduce RealisDance with multi-type poses, the pose gating module, the multi-layer pose guidance network, and the pose shuffle augmentation. We demonstrate the superiority of RealisDance through extensive qualitative comparisons. 

Although RealisDance has made significant improvements, especially in hand quality, it still suffers from two limitations. First, if the pose of the reference image is very distinguished from the pose sequence (for example, one is a close-up half-body pose and the other is a distant full-body pose), the generation quality will be very poor. Second, background stability highly relies on training data. When training data contain a non-static background, the generated background will be shaky.

{
    \small
    \bibliographystyle{ieeenat_fullname}
    \bibliography{main}
}


\end{document}